\documentclass[lettersize,journal]{IEEEtran}
\usepackage{array}
\usepackage{stfloats}
\usepackage{url}
\usepackage{verbatim}

\usepackage{cite}
\usepackage{amsmath,amssymb,amsfonts}
\usepackage{algorithmic}
\usepackage{graphicx}
\usepackage{textcomp}
\usepackage{xcolor}

\usepackage{multirow}
\usepackage{makecell}
\usepackage{bm}
\usepackage{amsmath}
\usepackage[ruled, vlined]{algorithm2e}
\usepackage{subfigure}
\usepackage{diagbox}
\begin{document}

\title{OIPR: Evaluation for Time-series Anomaly Detection Inspired by Operator Interest}

\author{Yuhan~Jing,
        Jingyu~Wang,~\IEEEmembership{Senior Member,~IEEE,}
        Lei~Zhang,
        Haifeng~Sun,
        Bo~He,
        Zirui~Zhuang,
        Chengsen~Wang,
        Qi~Qi,~\IEEEmembership{Senior Member,~IEEE,}
        and Jianxin~Liao,~\IEEEmembership{Senior Member,~IEEE}
\thanks{This work was supported in part by the National Key R\&D Program of China 2024YFE0200800, the National Natural Science Foundation of China under Grants (62471055, U23B2001, 62321001,  62401080, 62101064, 62171057, 62201072, 62071067), the High-Quality Development Project of the MIIT(2440STCZB2584), the Ministry of Education and China Mobile Joint Fund (MCM20200202, MCM20180101), the Project funded by China Postdoctoral Science Foundation (2023TQ0039, 2024M750257, GZC20230320), the Fundamental Research Funds for the Central Universities (2024PTB-004), the 2025 Education and Teaching Reform Project Funding at Beijing University of Posts and Telecommunications (2025YZ005). \textit{(Yuhan Jing and Jingyu Wang contributed equally to this work.)} \textit{(Corresponding authors: Lei Zhang; Bo He.)}}
\thanks{\;\; Yuhan Jing, Jingyu Wang, Haifeng Sun, Bo He, Zirui Zhuang, Chengsen Wang, Qi Qi and Jianxin Liao are with the State Key Laboratory of Networking and Switching Technology, Beijing University of Posts and Telecommunications, Beijing 100876, China (e-mail:$\{$jingyh, wangjingyu, hfsun, hebo, zhuangzirui, cswang, qiqi8266, liaojx$\}$@bupt.edu.cn).}
\thanks{\;\; Lei Zhang is with the China United Network Communications Co., Ltd., Beijing 100033, China (zhangl83@chinaunicom.cn).}
}



\maketitle

\begin{abstract}
With the growing adoption of time-series anomaly detection (TAD) technology, numerous studies have employed deep learning-based detectors to analyze time-series data in the fields of Internet services, industrial systems, and sensors. The selection and optimization of anomaly detectors strongly rely on the availability of an effective evaluation for TAD performance. Since anomalies in time-series data often manifest as a sequence of points, conventional metrics that solely consider the detection of individual points are inadequate. Existing TAD evaluators typically employ point-based or event-based metrics to capture the temporal context. However, point-based evaluators tend to overestimate detectors that excel only in detecting long anomalies, while event-based evaluators are susceptible to being misled by fragmented detection results.
To address these limitations, we propose \textit{OIPR}\footnote{The implementation of the proposed evaluator and the special scenario dataset are available at https://github.com/weatherjyh/OIPR.} (Operator Interest-based Precision and Recall metrics), a novel TAD evaluator with area-based metrics. It models the process of operators receiving detector alarms and handling anomalies, utilizing area under the operator interest curve to evaluate TAD performance. Furthermore, we build a special scenario dataset to compare the characteristics of different evaluators. Through experiments conducted on the special scenario dataset and five real-world datasets, we demonstrate the remarkable performance of \textit{OIPR} in extreme and complex scenarios. It achieves a balance between point and event perspectives, overcoming their primary limitations and offering applicability to broader situations.
\end{abstract}


\section{Introduction}
Time-series anomaly detection (TAD) \cite{AnomalyTransformer, OmniAnomaly, TS-SL} refers to the detection of a series of points with temporal continuity to identify time points or ranges that deviate from normal patterns. TAD is of significant importance in various fields, including fault detection and troubleshooting in industrial systems \cite{Industry3, Industry2}, Internet services \cite{GASF, Diner, EfficientKPI}, sensors \cite{Sensor1, Sensor2, Sensor5, Sensor4}. Detectors used for anomaly detection can be supervised or unsupervised, with their performance typically evaluated using manually annotated labels. Operators compare the detector output with the ground truth labels and calculate one or more metrics to evaluate TAD performance. In the context of binary classification, classical point-wise metrics (\textit{PW}) such as precision, recall, receiver operating characteristic (ROC) curve, and area under the ROC curve (AUROC) are commonly employed for performance evaluation of general anomaly detection tasks\cite{usePW3, usePW4}. In point-wise metrics, precision indicates the proportion of correctly identified positive points among all points predicted as positive, while recall represents the proportion of accurately predicted positive points relative to the total number of actual positive points.

\begin{figure}[t]
	\centering
	\includegraphics[width=3.5in]{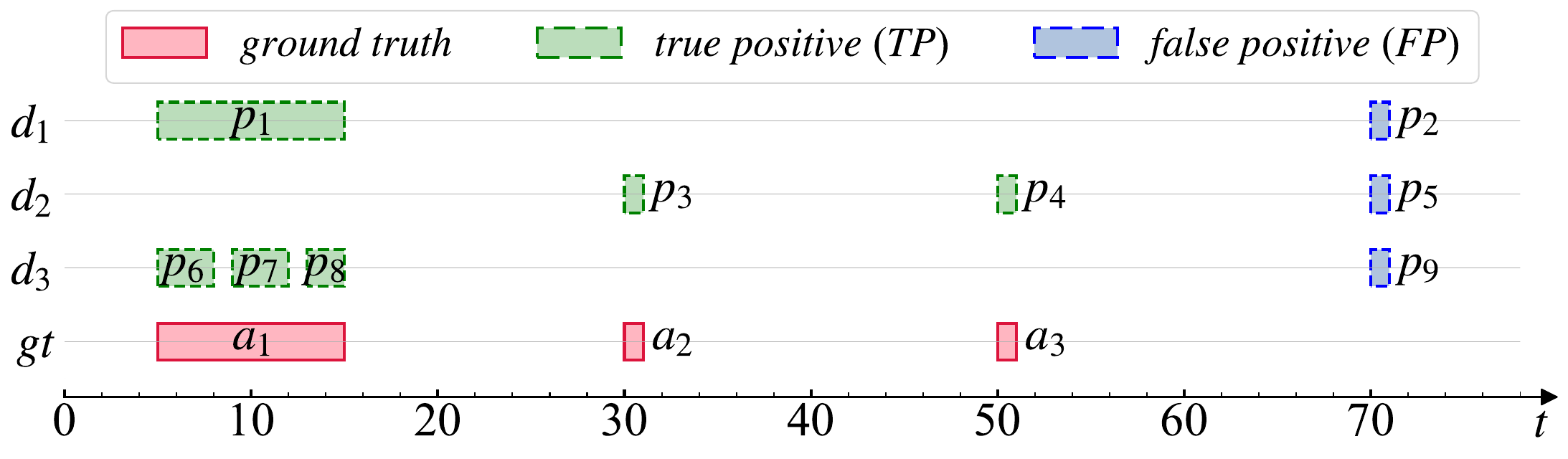}
	\caption{An example highlighting the distinction between point-based and event-based perspectives, comparing three different detectors $d_1$, $d_2$, and $d_3$ for the same ground truth.}
	\label{fig_1}
\end{figure}

\begin{table}[!t]
	\renewcommand{\arraystretch}{1.2}
	\caption{Evaluation results (\textbf{Precision/Recall/F1-score}) of the example in Fig. \ref{fig_1} using \textit{PA}, \textit{TaPR}, and \textit{OIPR}(Ours), respectively.}
	\label{table_1}
	\centering
    \setlength{\tabcolsep}{2pt}
    \scriptsize
	\begin{tabular}{c |ccc}
        \hline
        \renewcommand{\arraystretch}{1.0}
        \diagbox[width=58pt]{\textbf{Detector}}{\textbf{Evaluator}} & \textbf{PA} & \textbf{TaPR} & \textbf{OIPR (Ours)} \\  
        \hline
        \renewcommand{\arraystretch}{1.2}
        $d_1$ & 0.909/0.833/0.87 & 0.5/0.333/0.4 & 0.625/0.518/0.567\\
		% \hline
        $d_2$ & 0.667/0.167/0.267 & 0.667/0.667/0.667  & 0.609/0.485/0.54\\
		% \hline
        $d_3$ & 0.909/0.833/0.87 & 0.75/0.3/0.429 & 0.625/0.518/0.566\\
		\hline
	\end{tabular}
\end{table}

However, due to the continuity of time-series data, large Internet companies commonly employ event-based metrics for data analysis. For instance, within MAXIMO\footnote{https://www.ibm.com/topics/mttr}, the enterprise asset management system of IBM, the metrics of mean time to repair (MTTR) and mean time between failure (MTBF) are essential for assessing the availability and reliability of the system. They focus on capturing the frequency, duration, and recovery of each failure, signifying that ongoing events, rather than isolated time points, are regarded as the fundamental unit in maintenance operations. Therefore, researchers have recognized that the classical \textit{PW} metrics are inadequate for TAD evaluation, as they treat each time point as an individual unit, failing to account for the temporal continuity of anomaly events. In recent years, improved TAD evaluators \cite{PAK, RP/RR, Volume} have been proposed that take into account the continuity of time-series, which can be broadly classified into two categories: point-based and event-based evaluators. Point-based evaluators treat each anomaly point as equal, with specific predicted points adjusted to account for the temporal context. Meanwhile, event-based evaluators evaluate TAD performance based on anomaly events, regardless of their durations. To illustrate the distinctions between the point-based and event-based evaluators, we present an example in Fig. \ref{fig_1}, where three different detectors are applied to the same ground truth $gt$. Detector $d_1$ detects the longest anomaly event $a_1$ using one complete prediction event $p_1$; detector $d_2$ detects two shorter anomaly events, $a_2$ and $a_3$; detector $d_3$ detects $a_1$ using 3 fragmented prediction events ($p_6$, $p_7$, $p_8$). Additionally, each of $d_1$, $d_2$, and $d_3$ contains one false positive event, which are $p_2$, $p_5$, and $p_9$, respectively.

The evaluation results of the example in Fig. \ref{fig_1} using different evaluators are shown in Table \ref{table_1}. In these results, the PA evaluator \cite{PA}, a typical representative of point-based evaluators, considers $d_1$ and $d_3$ much superior to $d_2$. This indicates that point-based evaluators focus only on the number of anomaly points rather than the number of anomaly events.
In contrast, event-based evaluators, such as \textit{TaPR} \cite{TaPR}, treat each anomaly event as equivalent. 
In terms of recall, it assesses that $d_3$ performs slightly worse than $d_1$, due to its incomplete coverage of event $a_1$. $d_2$ is considered superior because it detects more anomaly events. However, in terms of precision, \textit{TaPR} yields counterintuitive results: it regards $d_3$ as having up to 3 correctly detected events, thus giving $d_3$ a significantly higher precision than $d_1$ (0.75 vs. 0.5). The characteristic of precision misleading of event-based evaluators regarding fragmented prediction events proves that merely focusing on event equivalence may lead to distorted evaluation conclusions.

To overcome the above limitations of point-based and event-based evaluators, we propose a novel TAD performance evaluator, named \textit{OIPR} 
(\underline{\textit{O}}perator \underline{\textit{I}}nterest-based \underline{\textit{P}}recision and \underline{\textit{R}}ecall metrics). 
In the example shown in Fig. \ref{fig_1}, \textit{OIPR}’s evaluation results indicate that $d_1$ (which detects more anomaly points) and $d_2$ (which detects more anomaly events) are two competitive detectors, with $d_1$ holding only a slight advantage. Meanwhile, \textit{OIPR} is not affected by the precision misleading caused by fragmented prediction events and therefore regards $d_1$ and $d_3$ to have similar performance. This perspective of balancing duration and quantity enables \textit{OIPR} to be applicable to a wider range of practical scenarios.
Additionally, we provide an artificial dataset containing nine special scenarios to analyze the characteristics of different evaluators under diverse boundary conditions. Experiments are also conducted on five real-world datasets to investigate the efficacy of different evaluators in complex practical scenarios.

The main contributions of this work are as follows:

\begin{itemize}
\item We propose a novel TAD evaluator that models the dynamic changes in operator interest while monitoring the time-series data and responding to detector alarms in real-world scenarios. It addresses the challenges posed by long anomaly events and fragmented detection results, and innovatively calculates precision and recall using the area under the operator interest curve.

\item We establish a special scenario dataset that allows for a comprehensive analysis of characteristics of various TAD evaluators under diverse boundary conditions. This dataset serves as a valuable research material for future investigations of TAD performance evaluation.

\item We conduct experiments on five real-world datasets using both representative and adversary detectors. The results indicate that \textit{OIPR} outperforms the baseline evaluators, exhibiting fewer limitations and greater applicability across a variety of real-world scenarios.

\end{itemize}

\section{Related Work}
\subsection{Time-series Anomaly Detection}
Anomalies in time-series datasets come from various sources, resulting in distinct anomaly characteristics. For instance, external attacks can lead to service interruptions, often shown as abrupt time-series fluctuations \cite{Attack}. Additionally, hardware failures, such as hard drive failures or network device outages, can cause a significant decline in system performance, resulting in sustained deterioration of associated indicators \cite{HardDrive}. Furthermore, the deployment or changes of services can induce variations in the corresponding key performance indicators (KPIs) over time, potentially triggering a series of consecutive or intermittent anomaly points \cite{Changes}. In terms of duration, anomalies can manifest at specific time points or persist across a sequence of consecutive time points \cite{AM}. The latter is referred to as an anomaly event that encompasses multiple anomaly points.

Typical techniques for TAD encompass a range of approaches, including statistical \cite{MPXXX}, machine learning \cite{SVMAD}, and deep learning \cite{PA, BeatGAN} algorithms. Time-series data are generally collected at regular intervals from various agents or sensors, with each time point representing a distinct sample. The detection results generated by the anomaly detector maintain the same discreteness and sampling frequency as the input data. After the process of detection, the discrete results can be systematically organized into prediction events based on temporal continuity, and compared with the ground truth labels to evaluate the performance of TAD. However, establishing a reliable mapping between the ground truth anomaly events and the detection results presents notable challenges. Specifically, temporal factors such as incorrect insertions, deletions, fragmentation, and merging \cite{PMAR} introduce significant ambiguities into the mapping relationships, thereby complicating the evaluation of TAD performance.

\subsection{Existing TAD Evaluators}
Classical \textit{PW} evaluator using point-based precision/recall (P/R) metrics has been employed in the evaluation of conventional TAD tasks\cite{usePW3, usePW4}. However, recent studies have underscored the significance of considering the temporal continuity inherent in time-series data, leading to the development of specialized evaluators. 
The \textit{PA} evaluator was initially introduced in \cite{PA}, which operates on the premise that if at least one point within an anomaly event is detected, the entire anomaly event is deemed successfully identified. Based on this, the evaluator of \textit{PA\%K} \cite{PAK} was proposed, which stipulates that a minimum proportion of points must be detected within a ground truth event to be classified as successfully detected. Both \textit{PA} and \textit{PA\%K} calculate their precision and recall metrics (P/R) based on the number of anomaly points, similar to \textit{PW}.

Different from point-based evaluators, event-based evaluators treat each continuous anomaly interval (i.e., anomaly event) as a single unit. The \textit{RP/RR} evaluator \cite{RP/RR} introduced factors of existence, size, position, and cardinality for the detection of anomaly events, supporting customizable functions or parameters for each factor. 
Meanwhile, \textit{TaPR} \cite{TaPR} addressed the challenge of ambiguous labeling by evaluating each ground truth event or prediction event through a combination of detection scores and portion scores. Both \textit{RP/RR} and \textit{TaPR} regarded each ground truth event as equally significant, irrespective of its duration. This principle is similarly applied to prediction events. Subsequently, the precision and recall metrics are averaged across the ground truth or prediction events.
Another event-based evaluator, the affiliation metrics (AM) \cite{AM}, provided an alternative perspective in which each ground truth event is considered equal, but the predicted results are assigned, measured in points, to the affiliation zone of the nearest ground truth event. The metrics of precision and recall are then averaged across the ground truth events.

\subsection{Summary}
Depending on specific focuses and assumptions, the aforementioned point-based and event-based evaluators will probably generate misleading results in certain extreme scenarios. These scenarios are typically characterized by long anomaly events or fragmented detection results, which restricts the applicability of evaluators \cite{Taxonomy}. In this study, we propose a comprehensive TAD evaluator based on the operator interest to aid operators in selecting more effective detectors for practical applications, and demonstrate its robust performance on the special scenario dataset and five real-world datasets. 

\begin{figure*}[t]
	\centering
	\includegraphics[width=7in]{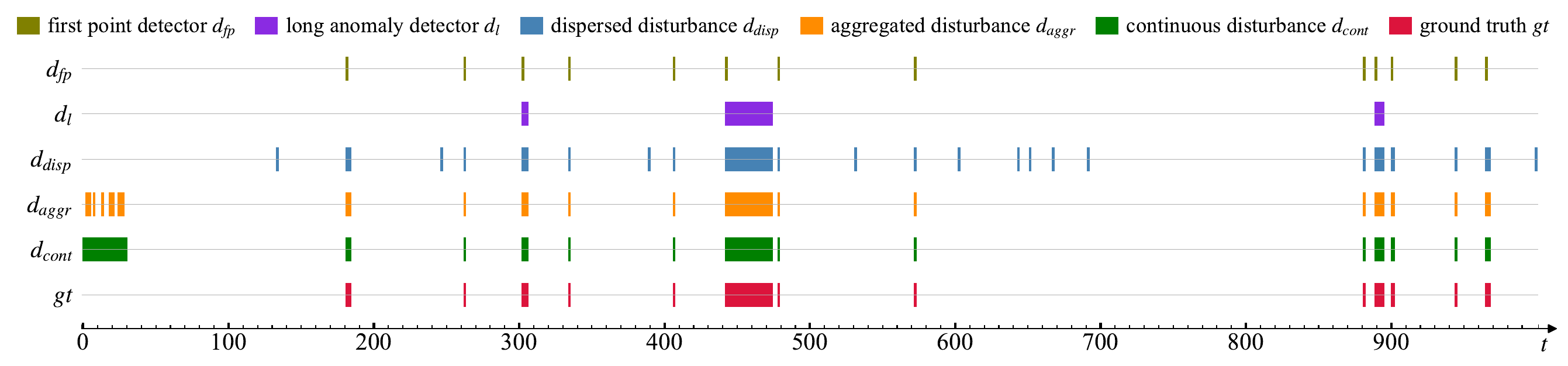}
	\caption{A demonstration scenario which displays 5 adversary detectors, including the first point detector $d_{fp}$, the long anomaly detector $d_l$, the dispersed disturbance detector $d_{disp}$, the aggregated disturbance detector $d_{aggr}$, and the continuous disturbance detector $d_{cont}$.}
	\label{fig_2}
\end{figure*}

\begin{table}[!t]
	\renewcommand{\arraystretch}{1.2}
	\caption{F1-scores of the demonstration in Fig. \ref{fig_2} using different evaluators. \textbf{Bold} text means the highest f1-score in each group.}
	\label{table_2}
    \setlength\tabcolsep{2pt}
    % \fontsize{7.6pt}{9pt}\selectfont
    \scriptsize
	\centering
	\begin{tabular}{c|ccccccc}
        \hline
        \renewcommand{\arraystretch}{1.0}
        \diagbox[width=94pt]{
          \hspace{15pt}\textbf{Detector}
        }{
          \textbf{Evaluator}\hspace{10pt}
        }
        & \bfseries PW & \bfseries PA & \bfseries PA\%K & \bfseries RP/RR & \bfseries TaPR & \bfseries AM & \bfseries OIPR\\
        \hline
        \renewcommand{\arraystretch}{1.2}
        First point detector $d_{fp}$     & 0.371 & \textbf{1.000}   & 0.371 & \textbf{0.926} & \textbf{0.908} & \textbf{0.964} & \textbf{0.896} \\
        Long anomaly detector $d_{l}$ & \textbf{0.848} & 0.848 & \textbf{0.848} & 0.375 & 0.375 & 0.375 & 0.552 \\
        \hline
	      Dispersed disturbance $d_{disp}$   & \textbf{0.919} & \textbf{0.919} & \textbf{0.919} & 0.722 & 0.722 & 0.942 & 0.761 \\
        Aggregated disturbance $d_{aggr}$   & \textbf{0.919} & \textbf{0.919} & \textbf{0.919} & 0.788 & 0.788 & \textbf{0.972} & \textbf{0.912} \\
	    Continuous disturbance $d_{cont}$   & 0.792 & 0.792 & 0.792 & \textbf{0.962} & \textbf{0.962} & 0.966 & 0.903 \\
	\hline
	\end{tabular}
\end{table}

\section{Motivation}\label{sec:3}
\subsection{Problem Formulation}
The problem of TAD and its evaluation can be formally articulated as follows: Given a time-series spanning $T$ time points, denoted as $\bm{x}=\{x_0, x_1, ..., x_{T-1}\}$, the corresponding ground truth labels are represented as $\bm{y}=\{y_0, y_1, ..., y_{T-1}\}$, where $y_t \in \{0, 1\}$ denotes whether the time-series is anomalous (1) or not (0) at time point $t$. For a specific anomaly detector, the detection results are denoted as $\bm{\hat{y}}=\{\hat{y}_0, \hat{y}_1, ..., \hat{y}_{T-1}\}$.

The classical \textit{PW} evaluator evaluates the TAD performance by calculating three primary metrics: precision (P), recall (R), and f1-score (F1). These metrics are defined as follows:
\begin{equation}
  P=\frac{TP}{TP + FP}, \ R=\frac{TP}{TP + FN}, \ 
  F1=\frac{2 \cdot P \cdot R}{P + R}, 
  \label{eq1}
\end{equation}
where TP, FP, and FN represent the number of true positive, false positive, and false negative points, respectively. Other specialized TAD evaluators typically adopt the P/R format, with different solutions for precision and recall metrics. Finally, the f1-score metric is calculated as outlined in (\ref{eq1}).

\subsection{Limitations of Existing Evaluators}\label{sec:3.2}
\subsubsection{Long Anomaly Effect}
As a distinctive binary classification task, TAD prompts researchers to develop specialized evaluators that take into account the temporal continuity of events. In this context, the existence detection of a greater number of anomaly events, rather than detecting more individual anomaly points, has become the primary consideration for operators. However, in point-based evaluators, the significance attributed to anomaly events is proportional to the number of points they encompass. As a result, a limited number of long anomalies can overshadow the influence of a larger quantity of shorter anomalies in the final evaluation outcomes.
To demonstrate the impact of the long anomaly effect, we employ two straightforward adversary detectors, one of which is designated as the first point detector, denoted as $d_{fp}$. It is specifically designed to identify only the initial point of each ground truth event. During periods without any ground truth anomalies, the output of $d_{fp}$ consistently remains at 0. While the first point detector successfully identifies the existence of every anomaly event, it lacks the ability to discern the durations of these anomalies.

Another adversary detector, referred to as the long anomaly detector $d_l(L)$, is designed to identify the ground truth events within a given time-series that have a duration of at least $L$. Specifically, $d_l(L)$ correctly detects all points that fall within the long ground truth events, while producing an output of 0 for all other time points. Although the long anomaly detector is effective in identifying anomaly events with long durations, it fails to detect the other shorter events.

For demonstration purposes, we extract a slice from the SMD dataset \cite{OmniAnomaly}, using both $d_{fp}$ and $d_{l}$ for anomaly detection, as illustrated in Fig. \ref{fig_2}. Among them, $d_{fp}$ accurately reports the occurrence of all thirteen anomaly events, which is significant for operators. In contrast, $d_{l}$ detects only three of these anomaly events. If it is deployed for fault detection in a service, operators will miss the opportunity to recognize these short anomaly events. Within the point-based evaluators, \textit{PW} and \textit{PA\%K} fail to reflect this risk. As demonstrated in Table \ref{table_2}, $d_{l}$ achieves a high f1-score of 0.848 using both evaluators, while $d_{fp}$ attains a significantly low f1-score of 0.371. This discrepancy arises because $d_{l}$ detects a greater number of anomaly points compared to $d_{fp}$. In contrast, \textit{PA} exhibits a distinct behavior of overestimation. It adjusts the detection result of $d_{fp}$ to an ideal detector (whose outputs perfectly match the ground truth), even though it does not actually detect the duration of any event comprising more than one point. 

Event-based evaluators, such as \textit{RP/RR}, \textit{TaPR}, and \textit{AM}, are not impacted by the long anomaly effect. Their evaluation results for the detectors primarily depend on the number of anomaly events they detect, and $d_{fp}$ is evaluated to be much better than $d_{l}$, as demonstrated in Table \ref{table_2}.

\begin{figure}[t]
	\centering
	\includegraphics[width=3.5in]{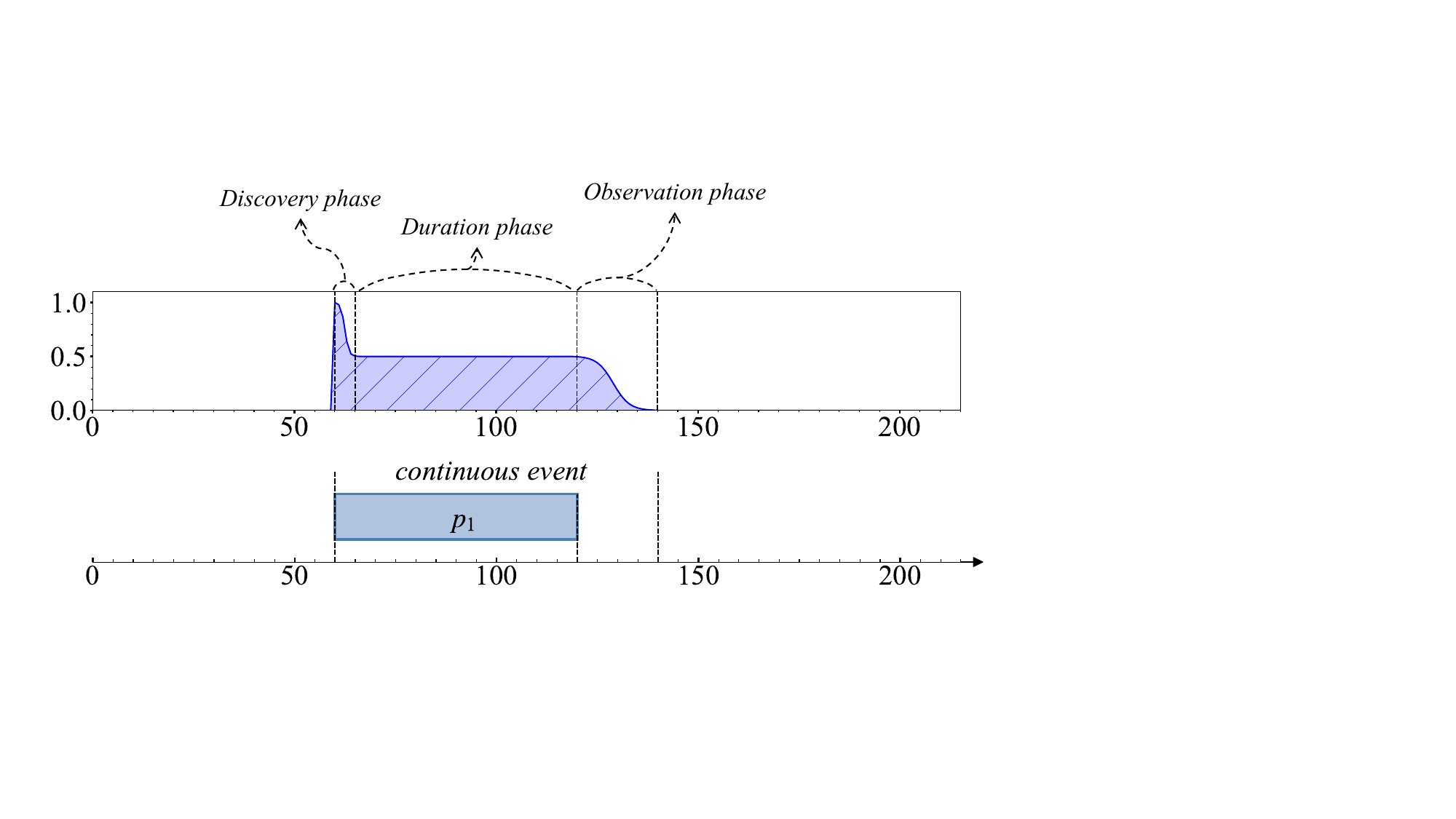}
	\caption{An example of the operator interest curve for an individual continuous anomaly event.}
	\label{fig_3}
\end{figure}

\subsubsection{Fragmentation Effect}
Although using an event-based evaluator can avoid the impact of the long anomaly effect, we have observed another misleading phenomenon, called the fragmentation effect, which stems from the discrete nature of the detection results. In instances where a ground truth event comprises a series of points, the detector is likely to identify only a subset of points that exhibit significant deviations from the normal pattern \cite{PAK}. Consequently, a contiguous event can be fragmented into multiple events in the detection results. The fragmentation effect leads to an increase in the number of true positive events, despite the fact that the successful detection is confined to a single original contiguous event. This phenomenon can also arise when the detector experiences a short-term, one-time disturbance, such as during a service deployment or change. In such cases, the detector is prone to generating a high frequency of false positive points, leading to multiple fragmented false positive events that originate from the same underlying cause. To empirically illustrate the fragmentation effect, we introduce three specific disturbances to the ideal detector, as depicted in Fig. \ref{fig_2}:

\textbf{Dispersed disturbance detector} $d_{disp}$: To simulate the interference induced by random noise, we generate a set of FP points, constituting 1\% of the entire time-series, which are then randomly inserted throughout the detection results.

\textbf{Aggregated disturbance detector} $d_{aggr}$: To simulate the short-term, intermittent disturbances associated with the deployment or change of the service, we randomly introduce a set of FP points, comprising 1\% of the entire time-series, into the initial 3\% of the detection results.

\textbf{Continuous disturbance detector} $d_{cont}$: The initial 3\% of the detection results is configured to a value of 1 to simulate a short-term, continuous disturbance scenario resulting from the deployment or change of service.

In the above three cases, both $d_{aggr}$ and $d_{cont}$ require operators to monitor system status for a period following initialization. During subsequent long-term operation, the detector is highly reliable. In contrast, $d_{disp}$ poses a significant challenge for operators due to its persistent and recurrent generation of false alarms, which ultimately results in resource wastage. As a result, $d_{disp}$ leads to a more significant decrease in the practical utility of the detector.

As illustrated in Table \ref{table_2}, the point-based evaluators, namely \textit{PW}, \textit{PA}, and \textit{PA\%K}, consider $d_{cont}$ as the worst-performing detector due to its highest number of false positive points, without taking into account that these points originate from the same anomaly event. In contrast, the event-based evaluators, \textit{RP/RR} and \textit{TaPR}, suggest that both $d_{aggr}$ and $d_{disp}$ exhibit similar poor performance, even though all false positive points in $d_{aggr}$ are concentrated within a limited period following initialization. Another evaluator that is partially event-based, \textit{AM}, remains unaffected by the fragmentation effect due to its exclusive adoption of an event-based perspective for the ground truth, rather than for the detection results.

\begin{figure}[t]
	\centering
	\includegraphics[width=3.5in]{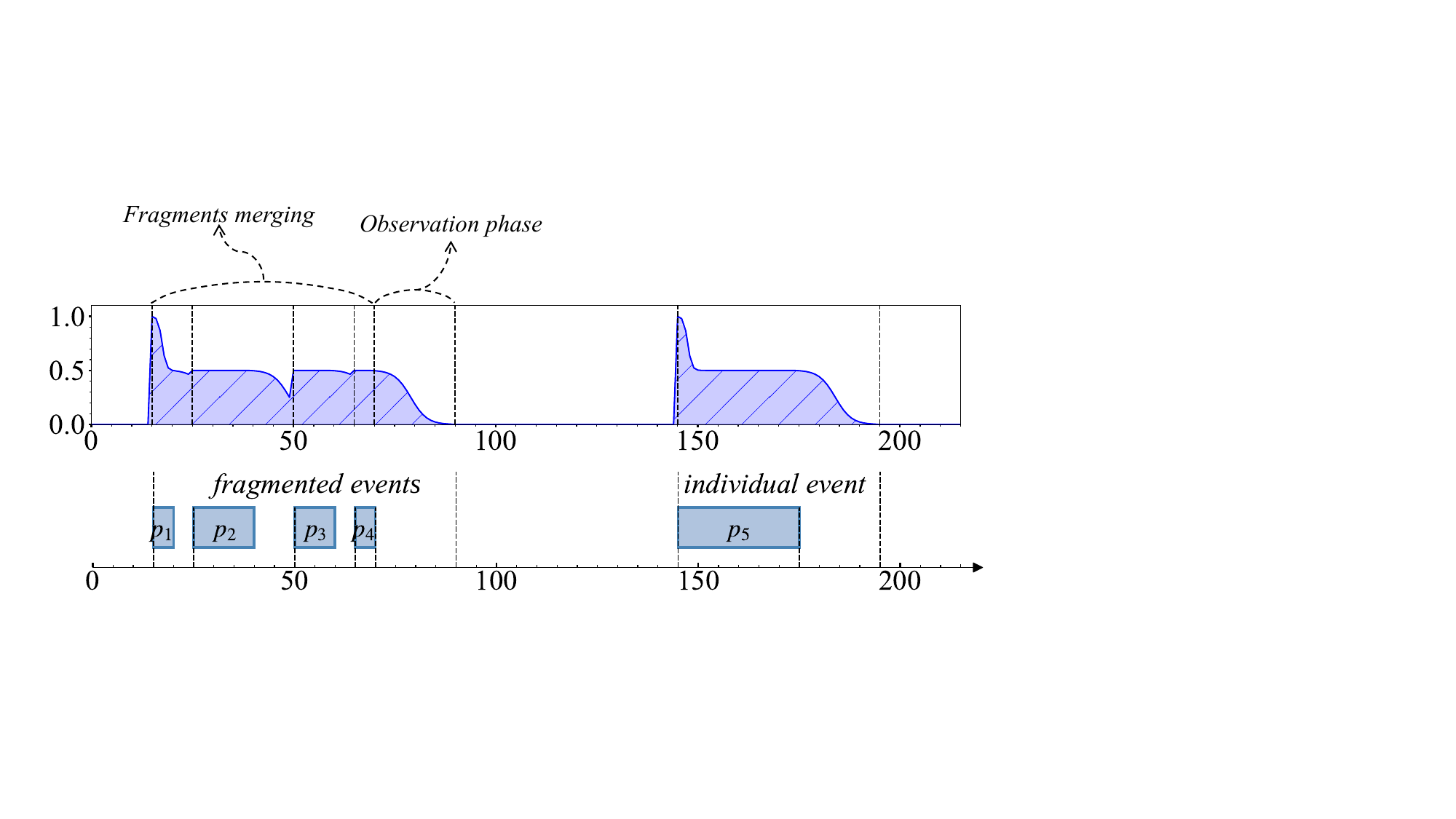}
	\caption{An example of the operator interest curve for fragmented anomaly events.}
	\label{fig_4}
\end{figure}

\begin{figure}[t]
	\centering
	\includegraphics[width=3.5in]{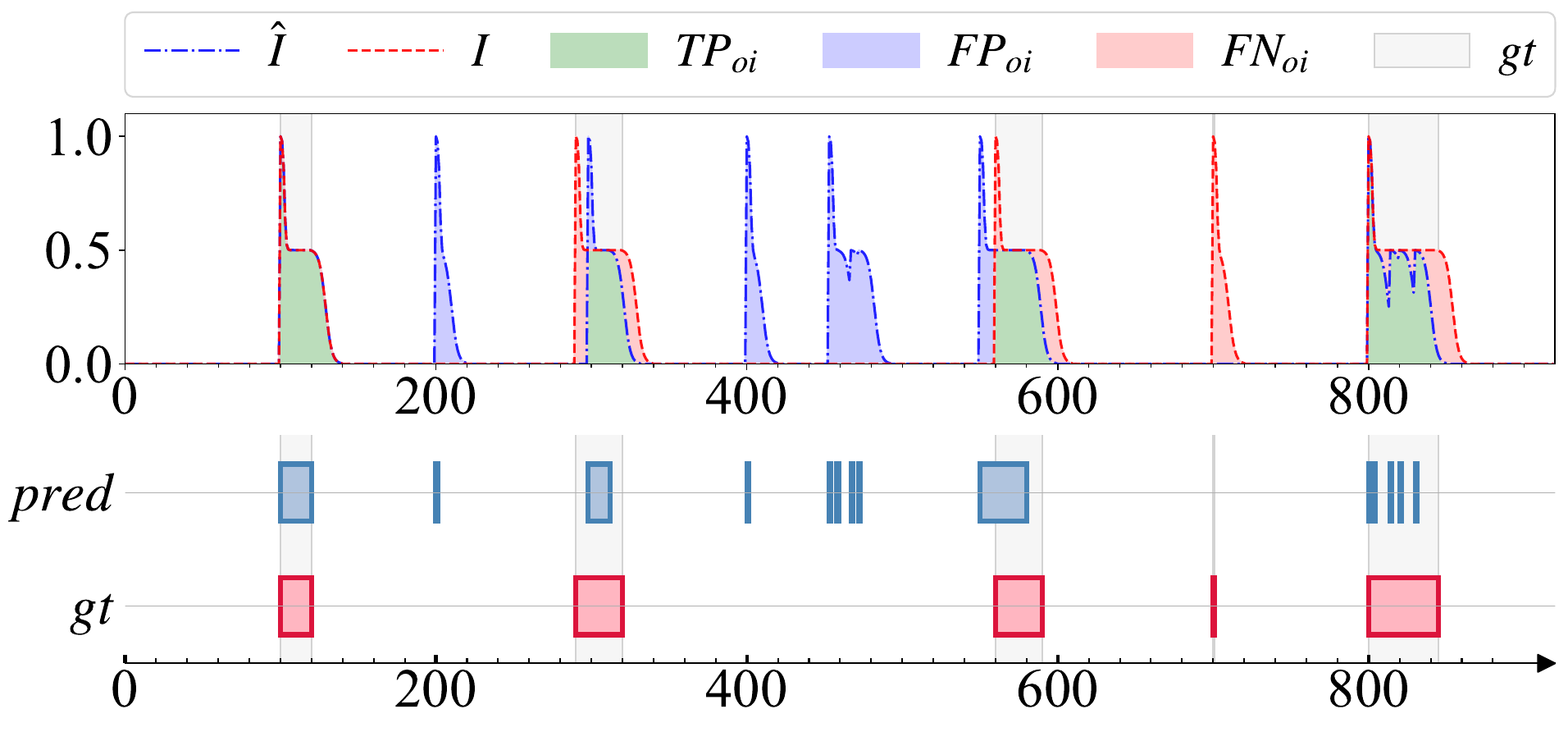}
	\caption{Visualization of the overlapping area of operator curves, corresponding to $TP_{oi}$, $FP_{oi}$, and $FN_{oi}$.}
	\label{fig_5}
\end{figure}

\subsection{Towards Universal TAD Evaluation}\label{sec:3.3}
By means of the adversary detectors and demonstration outlined in the last section, we have elucidated the limitations of point-based and event-based evaluators. In pursuit of developing a more universally applicable TAD evaluator, we have established two primary objectives:

\textbf{Existence detection reward.} The key difference between TAD and conventional binary classification tasks lies in the continuity of events. A universally applicable TAD evaluator should possess the capability to reward the existence detection of ground truth events. Specifically, the first point to detect the existence of an anomaly event should receive a higher reward than using the classical \textit{PW} evaluator.

\textbf{Fragments Merging.} Given that time points serve as the fundamental units for the collection of time-series data, the detection results inherently exhibit a discrete nature. Therefore, a universal TAD evaluator should take into account the potential merging of fragmented events and effectively differentiate between dispersed and aggregated anomaly points.

\renewcommand{\algorithmicrequire}{\textbf{Input:}}
\renewcommand{\algorithmicensure}{\textbf{Output:}}
\begin{algorithm}[t]
\caption{Online calculation process of the operator interest curve for the detection results}
\begin{algorithmic}[1]
\REQUIRE the detection results $\bm{\hat{y}}=\{\hat{y}_0, \hat{y}_1,\ldots,\hat{y}_{T-1}\}$, and \\ the pre-configured parameters $l_{dis}$, $l_{obs}$ and $b_{dur}$.
\ENSURE the operator interest $\bm{\hat{I}}$ for the detection results $\bm{\hat{y}}$.
\STATE $\bm{\hat{I}} \leftarrow \overbrace{\{0,0,\ldots,0\}}^{\scriptsize (T+l_{obs}) \, \text{zeros}}$
\STATE $p_{start} \leftarrow - l_{obs} - 1$
\STATE $p_{end} \leftarrow - l_{obs} - 1$
\STATE $t \leftarrow 0$
\WHILE{$t < T$}
  \IF{$\hat{y}_t = 1$}
    \IF{$t - p_{end} > l_{obs}$}
      \STATE $p_{start} \leftarrow t$
    \ENDIF
    \STATE $\hat{I}_t \leftarrow \omega(t - p_{start})$
    \STATE $p_{end} \leftarrow t$ 
  \ELSE
    \IF{$t - p_{end} \leq l_{obs}$}
      \STATE $\hat{I}_t \leftarrow \omega(t - p_{start}) \cdot \gamma(t - p_{end})$
    \ENDIF
  \ENDIF
  \STATE $t \leftarrow t + 1$
\ENDWHILE
\WHILE{$t < T + l_{obs}$}
  \IF{$t - p_{end} \leq l_{obs}$}
    \STATE $\hat{I}_t \leftarrow \omega(t - p_{start}) \cdot \gamma(t - p_{end})$
  \ENDIF
  \STATE $t \leftarrow t + 1$
\ENDWHILE
\STATE $\bm{\hat{I}} \leftarrow \{\hat{I}_0, \hat{I}_1, \ldots, \hat{I}_{T-1+l_{obs}}\}$
\RETURN $\bm{\hat{I}}$
\end{algorithmic}
\label{al_1}
\end{algorithm}

In addition to the above two objectives, it is also advisable for a universal evaluator to incorporate several other beneficial characteristics, such as addressing ambiguous labeling \cite{TaPR} and providing early detection reward \cite{RP/RR}. Further exploration of evaluator characteristics will be discussed in the context of special scenario dataset experiments in Section \ref{sec:4}.

To develop a more universally applicable TAD evaluator, we draw inspiration from the dynamic attention behavior of operators during the monitoring of anomaly alarms. In the 1960s, American psychologist A. M. Treisman proposed the attention selection theory and the attention attenuation model \cite{Attention}, which posits that unattended information is not completely blocked; instead, its intensity is reduced through an attenuation device.
In computer science, LSTM \cite{LSTM} implements similar attenuation logic via gating mechanisms. Transformer \cite{Transformer} achieves attention attenuation through scaled dot production: computes the input sequence correlation similarity, converts this to attention weights, and forms a data-driven dynamic attenuation process.
In this paper, we propose the operator interest curve to simulate the attention changes of operators when monitoring anomaly alarms, which consists of three phases: (\romannumeral1) Discovery phase: The operator receives alarms and confirms the presence of an anomaly; (\romannumeral2) Duration phase: Alarms persist and the operator takes responsive measures; (\romannumeral3) Observation phase: The anomaly is resolved through maintenance (with the alarms ceased), and the operator continues monitoring to ensure fault recovery. Based on the operator interest curve, we further propose a TAD evaluator, which offers an existence detection reward and enables the merging of potentially fragmented events.

\section{Methodology}\label{sec:4}
\subsection{Operator Interest Curve}\label{sec:4.1}
In this section, we outline the methodology employed to derive the operator interest curve for TAD, which captures the dynamics of operators in discovering and handling anomalies. To construct the operator interest curve, we propose a set of functions, denoted as $\Phi(i)$, which characterize the phases of discovery, duration, and observation associated with an individual anomaly event:
\begin{equation}
    \Phi(i)=\left \{
    \begin{aligned}
       &\omega(i), 0 \leq i < l_{dis} + l_{dur}\\
       &\omega(l_{dis} + l_{dur}) \times \gamma(i - l_{dis} - l_{dur} + 1), \\
       &\ \ \ \ l_{dis} + l_{dur} \leq i < l_{dis} + l_{dur} + l_{obs}\\
       &0, i \geq l_{dis} + l_{dur} + l_{obs}
    \end{aligned}
    \right.
    ,
\label{eq2}
\end{equation}
where $i$ represents the distance from the current point to the initial point of the anomaly event. $l_{dis}$, $l_{dur}$ and $l_{obs}$ represent the lengths of the discovery, duration, and observation phases, respectively. The discovery phase and the duration phase of the anomaly event are characterized by a shared continuous interest function $\omega(\cdot)$, while $\gamma(\cdot)$ represents the operator interest for the observation phase.
In Fig. \ref{fig_3}, we present an example of the operator interest curve for a continuous event: starting from the first anomaly point reported by the detector, $\Phi(i)$ jumps from 0 to 1. This indicates that the operator devotes the highest level of interest to the newly detected anomaly event; then, the level of interest decreases within the length of the discovery phase until transitioning to the duration phase, signifying that the operator shifts from receiving alarm information to conducting anomaly troubleshooting.
During the duration phase, the operator maintains interest in the anomaly event and conducts troubleshooting. Thus, the value of $\Phi(i)$ remains nearly unchanged with only a slight decrease. Finally, when the anomaly event is resolved and the detector stops reporting alarms, the monitoring enters the observation phase. At this stage, $\Phi(i)$ decreases continuously from its value at the end of the duration phase until it reaches 0, indicating that the operator continues to monitor the system for a period after the alarm ceases to ensure the recovery of the anomaly event.

As a default approach, we employ a reversed and scaled $\text{sigmoid}$ function for $\omega(\cdot)$:
\begin{equation}
    \omega(i) \! =  \! \left \{
    \begin{aligned}
       & 1, i = 0\\
       & b_{dur} \! + \! (1 \! - \! b_{dur}) \! \times \! \frac{1 \! -  \! \text{sigmoid}(\frac{10}{l_{dis}}i \! - \! 5)}{1 - \text{sigmoid}(-5)}, i  \! >  \! 0
    \end{aligned}
    \right.
    \!,
\label{eq3}
\end{equation}
where $\text{sigmoid}(x)=\frac{1}{1+e^(-x)}$. The operator interest decreases from 1 to near $b_{dur}$ throughout the discovery phase and remains nearly constant at $b_{dur}$ during the duration phase. Here, $b_{dur}$ denotes the lower boundary of operator interest for the discovery and duration phases. The rate of decrease is determined by the configuration of $l_{dis}$, which is set by default to 1/4 of the average length of ground truth events, rounded up to the nearest integer.

\begin{table}[!t]
	\renewcommand{\arraystretch}{1.2}
	\caption{Parameters employed in the experiments. $\overline{L}_a$ means the average length of the ground truth events within the dataset.}
	\label{table_3}
	\centering
	\begin{tabular}{p{24pt}<{\centering}  p{89pt}<{\centering}  p{101pt}<{\centering}}
		\hline
        {\bfseries Evaluator} & {\bfseries Parameter} & {\bfseries Value}\\
        \hline
        \bfseries PA\%K & Percentage threshold $K$ & 50\\
        \hline

	\multirow{6}{*}{\bfseries \makecell[c]{RP/RR}} 
        & Relative weight of existence reward $\alpha$ & \multirow{2}{*}{0.5} \\
	\cline{2-3} 
        & Overlap cardinality function $\gamma()$ & \multirow{2}{*}{$1/x$} \\
	\cline{2-3} 
        & Positional bias function $\delta()$ & \multirow{2}{*}{\makecell[c]{front-end bias for RR,\\ flat bias for RP}} \\
	\cline{2-3}
        & Overlap size function $\omega()$ & default function in \cite{RP/RR} \\  
	\hline
 
        \multirow{6}{*}{\bfseries TaPR} 
        & Relative weight of detection score $\alpha$ & \multirow{2}{*}{0.5} \\
        \cline{2-3} 
        & Number of Ambiguous Points $\delta$ & 5 for SMD/Special Scenarios, $\overline{L}_a$ for other datasets \\
        \cline{2-3} 
        & Detection possibilities threshold $\theta$ & \multirow{2}{*}{0} \\
	\hline
 
        \multirow{6}{*}{\bfseries \makecell[c]{OIPR \\(Ours)}} 
        & Length of discovery phase $l_{dis}$ & 5 for SMD/Special Scenarios, $\overline{L}_a/4$ for other datasets \\
        \cline{2-3} 
        & Length of observation phase $l_{obs}$ & 20 for SMD/Special Scenarios, $\overline{L}_a$ for other datasets \\
        \cline{2-3} 
        & Lower bound of interest in duration phase $b_{dur}$ & \multirow{2}{*}{0.5} \\
	\hline
	\end{tabular}
\end{table}

The interest function for the observation phase, $\gamma(i)$, is calculated according to (\ref{eq4}):
\begin{equation}
    \gamma(i)=\left \{
    \begin{aligned}
       &1, i = 0\\
       &\frac{1 - sigmoid(\frac{10}{l_{obs}}i - 5)}{1 - sigmoid(-5)}, 0 < i \leq l_{obs}\\
       &0, i > l_{obs}
    \end{aligned}
    \right.
    .
\label{eq4}
\end{equation}

The rate of decline is determined by the parameter $l_{obs}$, where a shorter $l_{obs}$ leads to a more rapid reduction during the observation phase. Typically, the default value of $l_{obs}$ is set to the average length of ground truth events.

\begin{table}[!t]
	\renewcommand{\arraystretch}{1.2}
        \setlength\tabcolsep{4pt}
	\caption{Descriptions of real-world datasets.}
	\label{table_4}
	\centering
	\begin{tabular}{p{40pt}<{\centering} p{195pt}<{\raggedright}}
        \hline
        \textbf{Dataset} & \makecell[c]{\textbf{Description}} \\
        \hline
        \multirow{2}{*}{\makecell[c]{\textbf{MSL\cite{Hundman}}}} & Spacecraft telemetry data from the Mars Science Laboratory (MSL) rover, Curiosity. \\
        \hline
        \multirow{2}{*}{\makecell[c]{\textbf{SMAP\cite{Hundman}}}} & Spacecraft telemetry data from the Soil Moisture Active Passive (SMAP) satellite.\\
        \hline
        \multirow{3}{*}{\makecell[c]{\textbf{PSM\cite{PSM}}}} & A benchmark from eBay’s multiple application server nodes encompasses 26 features describing server metrics, like CPU utilization and memory usage. \\
        \hline
        \multirow{2}{*}{\makecell[c]{\textbf{SMD \cite{OmniAnomaly}}}} & A 5-week dataset from a large Internet company contains 3 entity groups across 28 machines. \\
        \hline
        \multirow{2}{*}{\makecell[c]{\textbf{SWaT \cite{SWaT}}}} & Secure Water Treatment (SWaT) dataset collected from a physical testbed simulating a modern water-treatment plant. \\
        \hline
        \multirow{3}{*}{\makecell[c]{\textbf{Sensor-} \\ \textbf{Scope \cite{SensorScope}}}} & Environmental data including temperature, humidity, and solar radiation, collected from multiple sensors of a typical tiered sensor measurement system. \\
        \hline
        \multirow{3}{*}{\makecell[c]{\textbf{NAB\_} \\ \textbf{Traffic\cite{NAB}}}} & Real-time traffic data from Minnesota’s Twin Cities Metro area, including data from different sensors collecting occupancy, speed, and travel time. \\
        \hline
        \multirow{2}{*}{\makecell[c]{\textbf{IOPS \cite{TSB-UAD}}}} & Performance indicators that reflect the scale, quality of web services, and health status of a machine.\\
        \hline
        \multirow{2}{*}{\makecell[c]{\textbf{Yahoo \cite{Yahoo}}}} & Real production traffic data from the Yahoo production systems. \\
	    \hline
	\end{tabular}
\end{table}

\subsection{Merging Fragmented Anomaly Events}
In the previous section, we introduced the operator interest curve for an individual continuous anomaly event. However, it is essential to recognize that the prediction events can be fragmented. To address this issue, we propose a forward process for the online calculation of the operator interest curve, which is detailed in Algorithm \ref{al_1}.

Fig. \ref{fig_4} presents an example of the operator interest curve for fragmented anomaly events. Temporary interruptions in reporting anomaly points result in a decline in the operator interest, though it does not drop to 0 immediately. If new anomaly points are reported within $l_{obs}$, the fragmented anomaly events are merged. Conversely, if a new anomaly point is reported after the end of the last observation phase, it is classified as the initiation of a distinct individual anomaly event.

Using the same methodology as outlined in Algorithm \ref{al_1}, we derive the operator interest curve of the ground truth labels, denoted as $\bm{I}$. While the ground truth labels are not affected by the fragmentation effect, its operator interest curve represents the anticipated troubleshooting process of the ideal detector and serves as a benchmark for evaluation.

\vspace{-2mm}
\subsection{Precision and Recall metrics in OIPR}
The precision and recall metrics in \textit{OIPR} are derived using the operator interest curves. In this work, conventional number of true positive points is replaced with the true positive area $TP_{oi}$, which measures the overlapping area between the operator interest curve of the ground truth, $\bm{I}$, and that of the detection results, $\bm{\hat{I}}$. Likewise, the number of predicted positive points is replaced as the area under $\bm{\hat{I}}$. Therefore, the values for true positive area $TP_{oi}$ and false positive area $FP_{oi}$ in \textit{OIPR} are defined as:

\begin{equation}
  TP_{oi} = \text{AUC}( \text{min}( \bm{I}, \bm{\hat{I}} ) ),
\label{eq5}
\end{equation}
\begin{equation}
  FP_{oi} = \text{AUC}( \bm{\hat{I}} ) - TP_{oi},
\label{eq6}
\end{equation}
where $\text{AUC}(\cdot)$ refers to the computation of the area under a specific curve, while $\text{min}(\cdot)$ indicates the selection of the smaller value between two time-series at each time point to generate a new sequence. Subsequently, the precision metric in \textit{OIPR}, denoted as $P_{oi}$, is defined as:
\begin{equation}
  P_{oi} = \frac{TP_{oi}}{TP_{oi} + FP_{oi}} = \frac{\text{AUC}(\text{min}(\bm{I},\bm{\hat{I}}))}{\text{AUC}(\bm{\hat{I}})}.
\label{eq7}
\end{equation}

Using a similar methodology, we transform the ground truth positive points to the area under $\bm{I}$. Thus, the false negative area $FN_{oi}$ is calculated as follows:
\begin{equation}
  FN_{oi} = \text{AUC}( \bm{I} ) - TP_{oi}.
\label{eq8}
\end{equation}

The recall metric in \textit{OIPR}, denoted as $R_{oi}$, is defined as:
\begin{equation}
  R_{oi} = \frac{TP_{oi}}{TP_{oi} + FN_{oi}} = \frac{\text{AUC}(\text{min}(\bm{I},\bm{\hat{I}}))}{\text{AUC}(\bm{I})}.
\label{eq9}
\end{equation}

In addition, we present a visual illustration in Fig. \ref{fig_5} to demonstrate the AUC calculation range in \textit{OIPR}, with areas for $TP_{oi}$, $FP_{oi}$, and $FN_{oi}$ indicated separately.

\begin{table*}[!t]
	\renewcommand{\arraystretch}{1.2}
	\caption{Qualitative conclusions derived from experiments on the special scenario dataset. Beneficial characteristics of the evaluator are marked with \checkmark (present) and  $\times$ (absent). Misleading characteristics are indicated by $\circ$ (present) and - (absent).}
	\label{table_5}
	\centering
	\begin{tabular}{p{145pt}<{\centering} | p{25pt}<{\centering} p{25pt}<{\centering} p{40pt}<{\centering} p{25pt}<{\centering} p{40pt}<{\centering} p{25pt}<{\centering} p{50pt}<{\centering} }
    \hline
    \renewcommand{\arraystretch}{1.0}
    \diagbox[width=150pt]{
      \hspace{10pt}\textbf{Characteristics}
    }{
      \textbf{Evaluator}\hspace{10pt}
    }
    & \textbf{PW} & \textbf{PA} & \textbf{PA\%K} & \textbf{RP/RR} & \textbf{TaPR} & \textbf{AM} & \textbf{OIPR (Ours)}\\
    \hline
    \renewcommand{\arraystretch}{1.2}
   \textbf{ Existence detection reward} & $\times$ & \checkmark & *Piecewise & \checkmark & \checkmark & \checkmark & \checkmark \\
    \textbf{Overlapping proportion awareness} & \checkmark & $\times$ & *Piecewise & \checkmark & \checkmark & \checkmark & \checkmark \\
    \textbf{Fragments Merging} & $\times$ & $\times$ & $\times$ & $\times$ & $\times$ & $\times$ & \checkmark \\
	  \textbf{Addressing ambiguous labels} & $\times$ & $\times$ & $\times$ & $\times$ & *Only delay & \checkmark & \checkmark \\
	\textbf{Early detection reward} & $\times$ & $\times$ & $\times$ & \checkmark & $\times$ & $\times$ & \checkmark \\ 
    \hline
	  \textbf{Fragmentation misleading in precision} & - & - & - & $\circ$ & $\circ$ & - & - \\  
	  \textbf{Long anomaly misleading} & $\circ$ & $\circ$ & $\circ$ & - & - & - & *Mitigated \\     
	  \textbf{Sparse anomaly misleading} & - & - & - & - & - & $\circ$ & - \\
    \hline
	  \textbf{Number of custom parameters/functions} & 0 & 0 & 1 & 4 & 4 & 0 & 3 \\       
	\hline
    \multicolumn{8}{l}{\multirow{2}{*}{\makecell[l]{*Piecewise: According to the custom configuration of parameter $K$, \textit{PA\%K} has overlapping proportion awareness only in the $\leq K$ percentage \\ segment of events, and the existence detection reward only in the $>K$ percentage segment.}}}\\
    \multicolumn{8}{l}{\makecell[l]{}}\\ 
    \multicolumn{8}{l}{\makecell[l]{*Only delay: The ambiguous points after the end of the events are considered, while those before the start of the event are not.}}\\ 
    \multicolumn{8}{l}{\makecell[l]{*Mitigated: The influence of long anomaly misleading is mitigated rather than eliminated.}}\\ 
    \hline
    \end{tabular}
\end{table*}

\section{Experiments}
\subsection{Experimental Setup}
\textbf{Baseline TAD evaluators}. We conducted a comparative analysis against six baseline evaluators and \textit{\textbf{OIPR}} in the context of TAD evaluation. The baselines include three point-based evaluators: the classical point-wise evaluator \textit{\textbf{PW}}, the point adjustment evaluator \textit{\textbf{PA}} \cite{PA}, and the top-k point adjustment evaluator \textit{\textbf{PA\%K}} \cite{PAK}. In addition, three event-based evaluators are also included: the range-based TAD evaluator \textit{\textbf{RP/RR}} \cite{RP/RR}, the time-series aware evaluator \textit{\textbf{TaPR}} \cite{TaPR}, and the affiliation evaluator \textit{\textbf{AM}} \cite{AM}. The parameters employed for experiments are summarized in Table \ref{table_3}.

\textbf{Datasets}. The experiments are conducted on a special scenario dataset and five real-world datasets.

\begin{itemize}
\item Special scenario dataset. We have constructed an artificial dataset consisting of 24 evaluator-sensitive cases specifically designed for the evaluation of TAD. These cases are categorized into 9 distinct scenarios, each crafted to emphasize one or two evaluator characteristics. By leveraging the special scenario dataset, we can effectively demonstrate the limitations of various evaluators in a clear and concise manner.

\item Real-world datasets. The real-world datasets we used for experiments consists of data among various fields, such as spacecraft telemetry, web services, water treatment, environmental sensors, city traffic, and real production data. Descriptions for each dataset are listed in Table \ref{table_4}.
\end{itemize}

\begin{table*}[htbp]
\renewcommand{\arraystretch}{1.1}
\caption{Experimental results of advanced detectors Autoformer, DLinear and Timesnet on real-world datasets, evaluated by baseline evaluators and \textit{OIPR}. Evaluation metrics are presented in the \textbf{Precision/Recall/F1-score} format. \textbf{Bold} text indicates the highest f1-score and \underline{underlined} text represents the second-highest f1-score.}
\setlength{\tabcolsep}{1pt}
% \fontsize{7.5pt}{8.9pt}\selectfont
\scriptsize
\centering
\begin{tabular}{p{24pt}<{\centering} | p{50pt}<{\centering} | lllllll}
\hline
\renewcommand{\arraystretch}{1.0}
\textbf{Dataset} & 
\diagbox[width=53pt]{
  \textbf{Detector}
}{
  \textbf{Evaluator}\hspace{-3pt}
}
& \makecell[c]{\textbf{PW}} & \makecell[c]{\textbf{PA}} & \makecell[c]{\textbf{PA\%K}} & \bfseries \makecell[c]{\textbf{RP/RR}} & \bfseries \makecell[c]{\textbf{TaPR}} & \makecell[c]{\textbf{AM}} & \makecell[c]{\textbf{OIPR (Ours)}}  \\
\hline
\renewcommand{\arraystretch}{1.1}
\multirow{3}{*}{\makecell[c]{\textbf{MSL}}}
& Autoformer & \underline{0.81/0.721/0.763} & \underline{0.842/0.902/0.871} & \underline{0.831/0.835/0.833} & 0.041/0.608/0.077 & 0.144/0.739/0.241 & \textbf{0.756/0.956/0.845}$^{\triangledown}$ & 0.2/0.83/0.322 \\
& DLinear    & 0.963/0.604/0.742 & 0.968/0.71/0.819  & 0.968/0.701/0.813 & \textbf{0.128/0.423/0.196} & \textbf{0.197/0.449/0.274} & 0.736/0.626/0.677 & \textbf{0.481/0.571/0.522} \\
& Timesnet   & \textbf{0.88/0.723/0.794}$^{\circ}$ & \textbf{0.897/0.854/0.875}$^{\circ}$ & \textbf{0.895/0.835/0.864}$^{\circ}$ & \underline{0.05/0.597/0.093} & \underline{0.148/0.7/0.244} & \underline{0.765/0.909/0.831} & \underline{0.244/0.779/0.372} \\
\hline
\multirow{3}{*}{\makecell[c]{\textbf{SMAP}}}
& Autoformer & 0.541/0.614/0.575 & 0.6/0.781/0.678   & 0.556/0.653/0.6   & 0.01/0.483/0.02   & 0.097/0.617/0.167 & 0.513/0.731/0.603 & 0.211/0.614/0.314 \\
& DLinear    & \underline{0.983/0.524/0.684} & \underline{0.984/0.57/0.722} & \underline{0.984/0.56/0.714} & \textbf{0.179/0.456/0.257} & \textbf{0.286/0.529/0.371} & \underline{0.739/0.661/0.698} & \textbf{0.289/0.529/0.374} \\
& Timesnet   & \textbf{0.807/0.605/0.692}$^{\circ}$ & \textbf{0.844/0.783/0.812}$^{\circ}$ & \textbf{0.816/0.642/0.719}$^{\circ}$ & \underline{0.019/0.618/0.037} & \underline{0.118/0.695/0.202} & \textbf{0.635/0.869/0.734}$^{\triangledown}$ & \underline{0.223/0.678/0.336} \\
\hline
\multirow{3}{*}{\makecell[c]{\textbf{PSM}}}
& Autoformer & 1.0/0.789/0.882   & 1.0/0.822/0.902   & 1.0/0.81/0.895    & \underline{0.833/0.395/0.536} & \underline{0.889/0.399/0.551} & 0.965/0.43/0.595  & \underline{0.963/0.693/0.806} \\
& DLinear    & \textbf{0.998/0.932/0.964} & \textbf{0.998/0.981/0.99} & \textbf{0.998/0.973/0.986} & \textbf{0.653/0.685/0.668} & \textbf{0.77/0.719/0.744} & \underline{0.938/0.796/0.861} & \textbf{0.896/0.903/0.899} \\
& Timesnet   & \underline{0.932/0.95/0.941}$^{\circ}$  & \underline{0.935/0.999/0.966}$^{\circ}$ & \underline{0.935/0.995/0.964}$^{\circ}$ & 0.107/0.787/0.188 & 0.288/0.915/0.438 & \textbf{0.819/0.983/0.894}$^{\triangledown}$ & 0.537/0.935/0.682 \\
\hline
\multirow{3}{*}{\makecell[c]{\textbf{SMD}}}
& Autoformer & 0.77/0.659/0.71   & 0.77/0.659/0.71   & 0.77/0.659/0.71   & 0.818/0.534/0.646 & 0.818/0.547/0.655 & 0.941/0.543/0.689 & 0.828/0.58/0.682 \\
& DLinear    & \textbf{0.901/0.819/0.858} & \textbf{0.901/0.819/0.858} & \textbf{0.901/0.819/0.858} & \textbf{0.765/0.737/0.751} & \textbf{0.781/0.758/0.769} & \underline{0.955/0.749/0.84} & \textbf{0.84/0.786/0.812}\\
& Timesnet   & \underline{0.855/0.826/0.84 } & \underline{0.855/0.826/0.84 } & \underline{0.855/0.826/0.84 } & \underline{0.691/0.754/0.721} & \underline{0.732/0.784/0.757} & \textbf{0.946/0.766/0.847}$^{\triangledown}$ & \underline{0.787/0.797/0.792} \\
\hline
\multirow{3}{*}{\makecell[c]{\textbf{SWaT}}}
& Autoformer & 1.0/0.656/0.792   & 1.0/0.657/0.793   & 1.0/0.657/0.793   & 0.75/0.019/0.037 & 0.999/0.029/0.056 & 1.0/0.029/0.056   & \underline{0.997/0.455/0.625} \\
& DLinear    & \textbf{1.0/0.926/0.961} & \textbf{1.0/0.958/0.979} & \textbf{1.0/0.958/0.979} & \textbf{1.0/0.806/0.893}& \textbf{1.0/0.828/0.906} & \textbf{1.0/0.825/0.904} & \textbf{0.989/0.905/0.945} \\
& Timesnet   & \underline{0.747/0.931/0.829}$^{\circ}$ & \underline{0.752/0.958/0.843}$^{\circ}$ & \underline{0.752/0.958/0.843}$^{\circ}$ & \underline{0.02/0.679/0.039}$^{\circleddash}$ & \underline{0.096/0.887/0.173}$^{\circleddash}$ & \underline{0.765/0.976/0.858}$^{\circleddash}$ & 0.198/0.93/0.327 \\
\hline
\multirow{3}{*}{\makecell[c]{\textbf{Sensor-} \\ \textbf{Scope}}}
& Autoformer & 0.411/0.02/0.038  & 0.895/0.344/0.475 & 0.424/0.022/0.041 & 0.453/0.175/0.233 & 0.554/0.207/0.28  & 0.749/0.371/0.464 & 0.537/0.196/0.266 \\
& DLinear    & \underline{0.545/0.026/0.049} & \underline{0.951/0.502/0.646} & \underline{0.551/0.027/0.052} & \underline{0.444/0.253/0.299} & \underline{0.55/0.323/0.39} & \underline{0.664/0.581/0.607} & \underline{0.418/0.293/0.338} \\
& Timesnet   & \textbf{0.402/0.06/0.104} & \textbf{0.903/0.923/0.907} & \textbf{0.41/0.064/0.109} & \textbf{0.363/0.46/0.378} & \textbf{0.487/0.522/0.485} & \textbf{0.624/0.892/0.728} & \textbf{0.397/0.699/0.493} \\
\hline
\multirow{3}{*}{\makecell[c]{\textbf{NAB\_} \\ \textbf{Traffic}}}
& Autoformer & \underline{0.512/0.064/0.113} & \underline{0.93/0.916/0.917 } & \textbf{0.596/0.199/0.235}$^{\circledcirc}$ & \underline{0.404/0.474/0.396} & \underline{0.452/0.492/0.44 } & \underline{0.754/0.893/0.804} & \underline{0.357/0.401/0.334} \\
& DLinear    & 0.438/0.059/0.104 & \textbf{0.932/0.999/0.964} & \underline{0.53/0.197/0.234}  & \textbf{0.464/0.509/0.448} & \textbf{0.494/0.53/0.484} & \textbf{0.725/0.943/0.809} & \textbf{0.318/0.44/0.335 }\\
& Timesnet   & \textbf{0.284/0.091/0.132}$^{\circledcirc}$ & 0.811/0.999/0.892 & 0.357/0.22/0.23   & 0.221/0.515/0.29  & 0.257/0.551/0.335 & 0.645/0.955/0.769 & 0.2/0.623/0.276 \\
\hline
\multirow{3}{*}{\makecell[c]{\textbf{IOPS}}}
& Autoformer & \textbf{0.457/0.263/0.194}$^{\circledcirc}$ & \underline{0.578/0.617/0.441} & \textbf{0.475/0.326/0.23}$^{\circledcirc}$  & \underline{0.414/0.401/0.255} & \underline{0.435/0.455/0.287} & 0.783/0.643/0.643 & \underline{0.387/0.422/0.262} \\
& DLinear    & \underline{0.489/0.197/0.161} & \textbf{0.689/0.668/0.527} & \underline{0.505/0.229/0.181} & \textbf{0.474/0.494/0.374} & \textbf{0.534/0.569/0.431} & \textbf{0.823/0.788/0.764} & \textbf{0.441/0.468/0.329} \\
& Timesnet   & 0.118/0.271/0.121 & 0.316/0.867/0.41  & 0.13/0.313/0.137  & 0.087/0.613/0.135 & 0.112/0.708/0.173 & \underline{0.621/0.946/0.745}$^{\triangledown}$ & 0.091/0.582/0.147 \\
\hline
\multirow{3}{*}{\makecell[c]{\textbf{Yahoo}}}
& Autoformer & 0.117/0.388/0.138 & 0.124/0.426/0.162 & 0.117/0.399/0.146 & 0.117/0.4/0.149   & 0.125/0.418/0.161 & 0.487/0.678/0.527 & 0.111/0.395/0.146 \\
& DLinear    & \textbf{0.436/0.626/0.424} & \textbf{0.443/0.656/0.444} & \textbf{0.439/0.64/0.433} & \textbf{0.431/0.644/0.427} & \textbf{0.449/0.66/0.447} & \underline{0.765/0.757/0.725} & \textbf{0.438/0.643/0.435} \\
& Timesnet   & \underline{0.298/0.867/0.369} & \underline{0.301/0.886/0.378} & \underline{0.299/0.876/0.373} & \underline{0.313/0.88/0.388} & \underline{0.333/0.899/0.411} & \textbf{0.716/0.954/0.803}$^{\triangledown}$ & \underline{0.307/0.872/0.383} \\
\hline
\multicolumn{9}{l}{Lack of beneficial characteristics: $^{\circledcirc}$lack of existence detection reward, $^{\circleddash}$lack of fragments merging.} \\
\multicolumn{9}{l}{Misleading characteristics: $^{\circ}$long anomaly misleading, $^{\triangledown}$sparse anomaly misleading.} \\
\hline
\end{tabular}
\label{table_6}
\end{table*}

\begin{figure*}[htbp]
    \setlength{\subfigcapskip}{-2.9pt}  % 局部生效，仅当前图的子图间距
	\centering
	\subfigure[PSM.] {\includegraphics[width=3.5in]{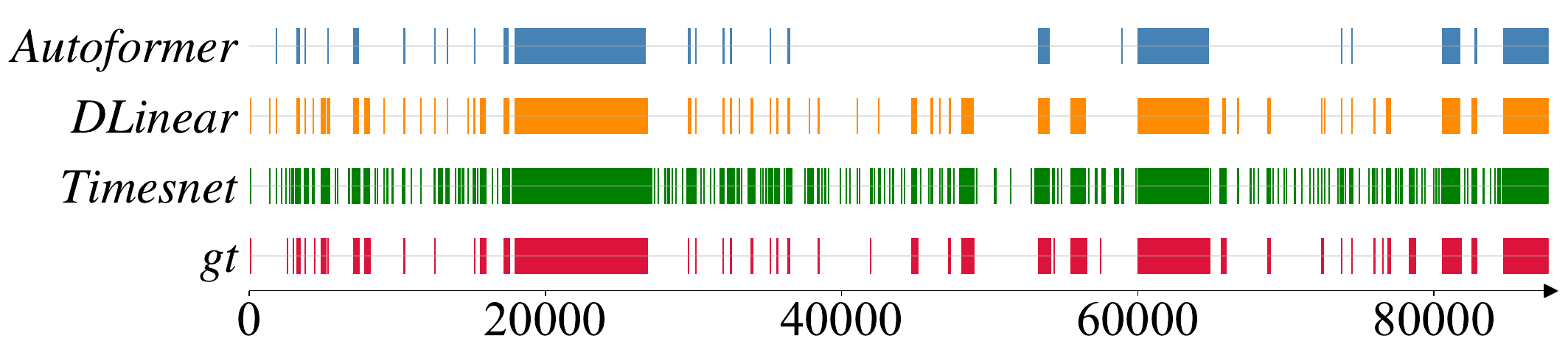}\label{fig_6a}}
	\subfigure[IOPS.] {\includegraphics[width=3.5in]{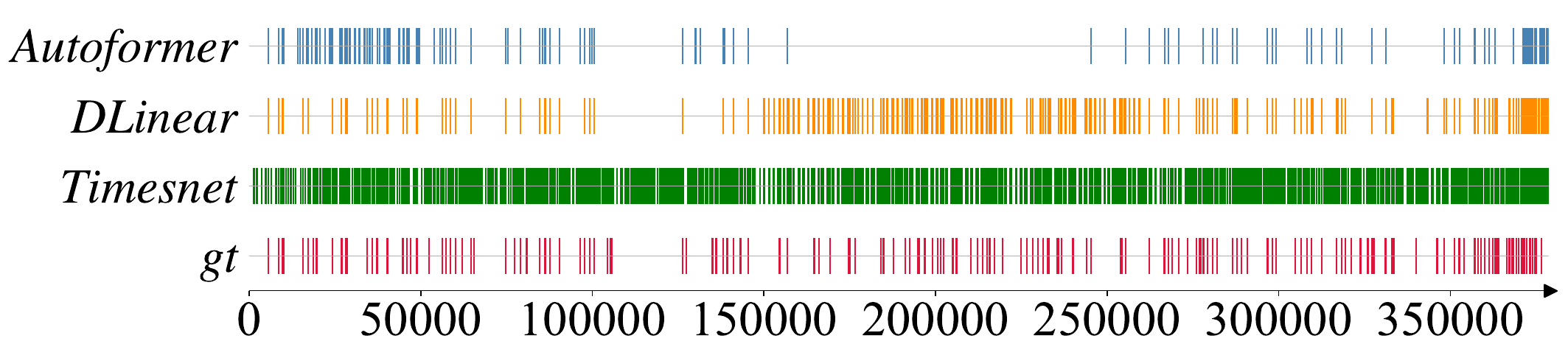}\label{fig_6b}}\\
    \vspace{-5pt}
	\subfigure[SWaT.] {\includegraphics[width=3.5in]{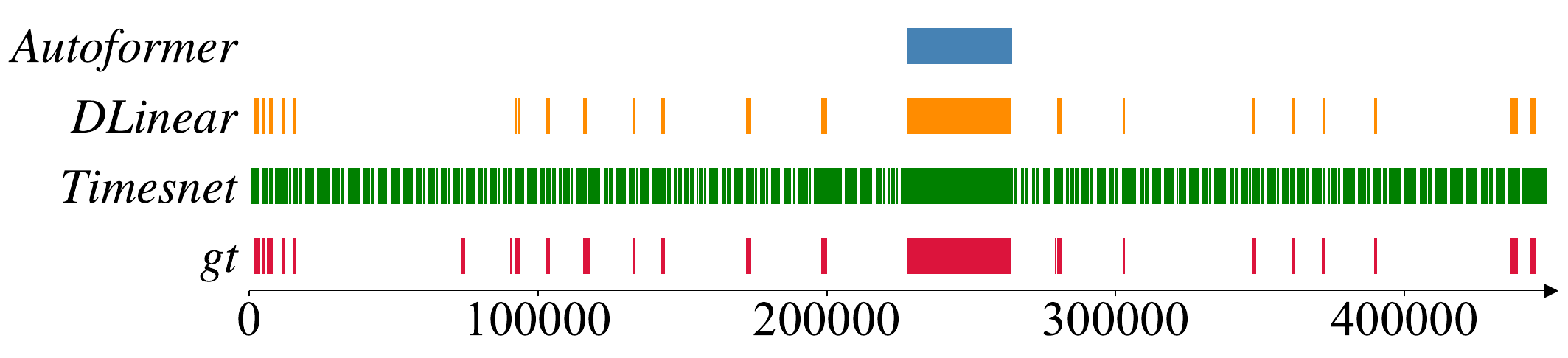}\label{fig_6c}}
    \subfigure[Yahoo.] {\includegraphics[width=3.5in]{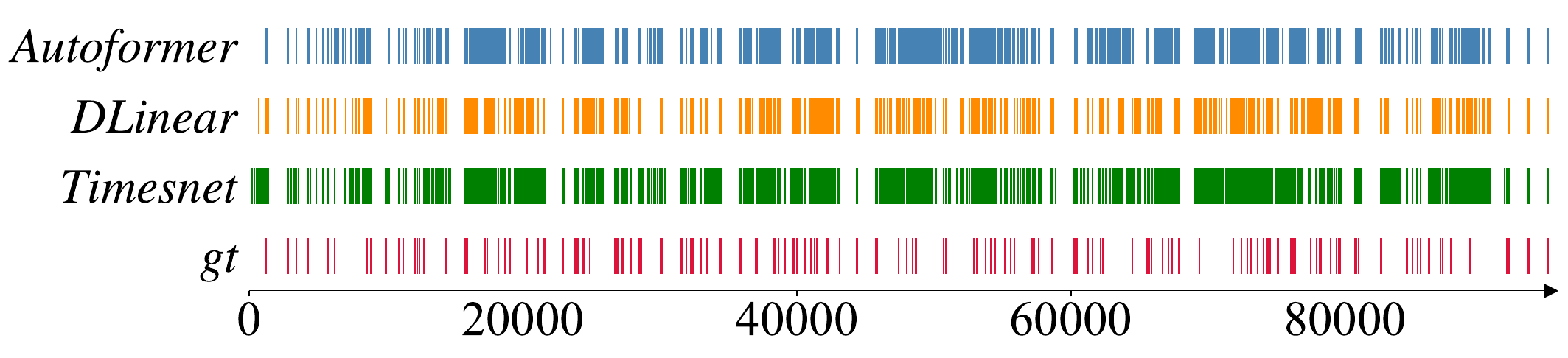}\label{fig_6d}}
	\caption{Demonstrations of TAD results on several real-world datasets.}
    \vspace{-5pt}
	\label{fig_6}
\end{figure*}

\vspace{-2mm}
\subsection{Experimental Results}
\subsubsection{Experimental Results on the Special Scenario Dataset}
In this section, we present several significant qualitative conclusions derived from experiments on the special scenario dataset, with the results shown in Table \ref{table_5}. The evaluator characteristics involved in these qualitative conclusions include the existence detection reward and fragments merging mentioned in Section \ref{sec:3.3}, as well as several additional ones:

\begin{itemize}
    \item \textbf{Overlapping proportion awareness}: For a specific ground truth event, a detector that identifies a greater proportion of anomaly points within it should be awarded a higher recall reward. This characteristic encourages the detector to identify as many points as possible within a single ground truth event, thereby improving the accuracy of event duration reporting.
    
    \item \textbf{Addressing ambiguous labels}: The manual labeling process introduces ambiguity in defining anomaly event boundaries, resulting in anomalies affecting points before or after the ground truth event. Detecting these ambiguous points indicates a partial success in identifying the anomaly. Consequently, evaluators that address ambiguous labels should reward recall for these points.
    
    \item \textbf{Early detection reward}: It encourages detectors to identify anomalies early in the occurrence of the ground truth events, thereby improving the detection timeliness.

    \item \textbf{Fragmentation misleading in precision}: This misleading characteristic is present in several event-based evaluators. Detectors identifying a ground truth event through multiple fragmented events can achieve higher precision scores than those detect using a complete event. This discrepancy primarily arises from the misleading increase in the count of true positive events.
    
    \item \textbf{Long anomaly misleading}: This misleading characteristic is present in most point-based evaluators, where the significance of long anomalies far surpasses that of short anomalies. Consequently, detectors that identify more short anomalies may be outperformed by those that focus solely on long anomalies, hindering the detection of a more diverse range of anomalies.
    
    \item \textbf{Sparse anomaly misleading}: A pitfall observed in the \textit{AM} evaluator. The f1-score is overestimated due to the mapping from long absolute distance to limited relative precision and recall distances.
\end{itemize}

A comprehensive introduction and detailed experimental results pertaining to the special scenario dataset are presented in Appendix A. Among the above characteristics, \textit{OIPR} effectively incorporates all beneficial factors while minimizing potential misleading influences. Compared to baseline point-based and event-based evaluators, it offers enhanced universality and reduces deficiencies in extreme scenarios.

\begin{table*}[htbp]
\renewcommand{\arraystretch}{1.2}
\caption{Experimental results of the first point detector $d_{fp}$ and the long anomaly detector $d_l$, evaluated by baseline evaluators and \textit{OIPR}. Evaluation metrics are presented in the \textbf{Precision/Recall/F1-score} format. \textbf{Bold} text indicates the highest f1-score and \underline{underlined} text represents the second-highest f1-score.}
\setlength{\tabcolsep}{2pt}
\scriptsize
\centering
\begin{tabular}{c|p{50pt}<{\centering}| lllllll}
\hline
\renewcommand{\arraystretch}{1.0}
\textbf{Dataset} & 
\diagbox[width=53pt]{
  \textbf{Detector}
}{
  \textbf{Evaluator}\hspace{-4pt}
}
& \makecell[c]{\textbf{PW}} & \makecell[c]{\textbf{PA}} & \makecell[c]{\textbf{PA\%K}} & \bfseries \makecell[c]{\textbf{RP/RR}} & \bfseries \makecell[c]{\textbf{TaPR}} & \makecell[c]{\textbf{AM}} & \makecell[c]{\textbf{OIPR (Ours)}}  \\
\hline
\renewcommand{\arraystretch}{1.3}
\multirow{2}{*}{\makecell[c]{\textbf{MSL}}}
& $d_{fp}$ & 1.0/0.005/0.009$^{\circledcirc}$  & \textbf{1.0/1.0/1.0}$^{\circledast}$  & 1.0/0.005/0.009$^{\circledcirc}$  & \textbf{1.0/0.514/0.679}  & \textbf{1.0/0.507/0.673}  & \textbf{1.0/0.885/0.939}  & \textbf{1.0/0.386/0.557} \\
& $d_l$ & \textbf{1.0/0.46/0.63}$^{\circ}$  & 1.0/0.46/0.63$^{\circ}$  & \textbf{1.0/0.46/0.63}$^{\circ}$  & 1.0/0.111/0.2  & 1.0/0.111/0.2  & 1.0/0.111/0.2  & 1.0/0.328/0.494 \\
\hline
\multirow{2}{*}{\makecell[c]{\textbf{SMAP}}}
& $d_{fp}$ & 1.0/0.001/0.002$^{\circledcirc}$  & \textbf{1.0/1.0/1.0}$^{\circledast}$  & 1.0/0.001/0.002$^{\circledcirc}$  & \textbf{1.0/0.509/0.675}  & \textbf{1.0/0.505/0.671}  & \textbf{1.0/0.894/0.944}  & 0.994/0.381/0.551 \\
& $d_l$ & \textbf{1.0/0.704/0.826}$^{\circ}$  & 1.0/0.704/0.826$^{\circ}$  & \textbf{1.0/0.704/0.826}$^{\circ}$  & 1.0/0.179/0.304  & 1.0/0.179/0.304  & 1.0/0.179/0.304  & \textbf{1.0/0.507/0.673} \\
\hline
\multirow{2}{*}{\makecell[c]{\textbf{PSM}}}
& $d_{fp}$ & 1.0/0.003/0.006$^{\circledcirc}$  & \textbf{1.0/1.0/1.0}$^{\circledast}$  & 1.0/0.128/0.226$^{\circledcirc}$  & \textbf{1.0/0.697/0.821}  & \textbf{1.0/0.667/0.801}  & \textbf{1.0/0.892/0.943}  & 0.993/0.328/0.493 \\
& $d_l$ & \textbf{1.0/0.781/0.877}$^{\circ}$  & 1.0/0.781/0.877$^{\circ}$  & \textbf{1.0/0.781/0.877}$^{\circ}$  & 1.0/0.069/0.13  & 1.0/0.069/0.13  & 1.0/0.069/0.13  & \textbf{1.0/0.579/0.734} \\
\hline
\multirow{2}{*}{\makecell[c]{\textbf{SMD}}}
& $d_{fp}$ & 1.0/0.395/0.566$^{\circledcirc}$  & \textbf{1.0/1.0/1.0}$^{\circledast}$  & 1.0/0.395/0.566$^{\circledcirc}$  & \textbf{1.0/0.887/0.94}  & \textbf{1.0/0.857/0.923}  & \textbf{1.0/0.955/0.977}  & \textbf{0.993/0.91/0.95} \\
& $d_l$ & \textbf{1.0/0.572/0.728}$^{\circ}$  & 1.0/0.572/0.728$^{\circ}$  & \textbf{1.0/0.572/0.728}$^{\circ}$  & 1.0/0.203/0.338  & 1.0/0.216/0.355  & 1.0/0.203/0.338  & 0.95/0.26/0.408 \\
\hline
\multirow{2}{*}{\makecell[c]{\textbf{SWaT}}}
& $d_{fp}$ & 1.0/0.001/0.001$^{\circledcirc}$  & \textbf{1.0/1.0/1.0}$^{\circledast}$  & 1.0/0.001/0.001$^{\circledcirc}$  & \textbf{1.0/0.503/0.669}  & \textbf{1.001/0.501/0.668}  & \textbf{1.0/0.849/0.918}  & 0.994/0.361/0.529 \\
& $d_l$ & \textbf{1.0/0.657/0.793}$^{\circ}$  & 1.0/0.657/0.793$^{\circ}$  & \textbf{1.0/0.657/0.793}$^{\circ}$  & 1.0/0.029/0.056  & 1.0/0.029/0.056  & 1.0/0.029/0.056  & \textbf{1.0/0.455/0.626} \\
\hline
\multicolumn{9}{l}{Lack of beneficial characteristic: $^{\circledcirc}$lack of existence detection reward, $^{\circledast}$lack of overlapping proportion awareness.} \\
\multicolumn{9}{l}{Misleading characteristic: $^{\circ}$long anomaly misleading.} \\
\hline
\end{tabular}
\label{table_7}
\end{table*}

\begin{table*}[htbp]
\renewcommand{\arraystretch}{1.2}
\caption{Experimental results of the dispersed disturbance detector $d_{disp}$, the aggregated disturbance detector $d_{aggr}$, and the continuous disturbance detector $d_{cont}$. Evaluation metrics are presented in the \textbf{Precision/Recall/F1-score} format. \textbf{Bold} text indicates the highest f1-score and \underline{underlined} text represents the second-highest f1-score.}
\setlength{\tabcolsep}{2pt}
\scriptsize
\centering
\begin{tabular}{p{30pt}<{\centering} | p{50pt}<{\centering} | l llllll}
\hline
\renewcommand{\arraystretch}{1.0}
\textbf{Dataset} & 
\diagbox[width=53pt]{
  \textbf{Detector}
}{
  \textbf{Evaluator}\hspace{-3pt}
}
& \makecell[c]{\textbf{PW}} & \makecell[c]{\textbf{PA}} & \makecell[c]{\textbf{PA\%K}} & \bfseries \makecell[c]{\textbf{RP/RR}} & \bfseries \makecell[c]{\textbf{TaPR}} & \makecell[c]{\textbf{AM}} & \makecell[c]{\textbf{OIPR (Ours)}}  \\
\hline
\renewcommand{\arraystretch}{1.2}
\multirow{3}{*}{\makecell[c]{\textbf{MSL}}}
& $d_{disp}$ & \textbf{0.913/1.0/0.955}$^{\circ}$  & \textbf{0.913/1.0/0.955}$^{\circ}$  & \textbf{0.913/1.0/0.955}$^{\circ}$  & 0.047/1.0/0.091  & 0.133/1.0/0.234  & 0.914/1.0/0.955$^{\triangledown}$  & 0.218/0.941/0.354 \\
& $d_{aggr}$ & \textbf{0.913/1.0/0.955}  & \textbf{0.913/1.0/0.955}  & \textbf{0.913/1.0/0.955}  & 0.06/1.0/0.114$^{\circleddash}$  & 0.211/1.0/0.349$^{\circleddash}$  & \textbf{0.961/1.0/0.98}  & \textbf{0.803/0.991/0.887} \\
& $d_{cont}$ & 0.704/1.0/0.826  & 0.704/1.0/0.826  & 0.704/1.0/0.826  & \textbf{0.972/1.0/0.986}  & \textbf{0.988/1.0/0.994}  & 0.948/1.0/0.973  & \textbf{0.802/0.991/0.887} \\
\hline
\multirow{3}{*}{\makecell[c]{\textbf{SMAP}}}
& $d_{disp}$ & \textbf{0.928/1.0/0.962}$^{\circ}$  & \textbf{0.928/1.0/0.962}$^{\circ}$  & \textbf{0.928/1.0/0.962}$^{\circ}$  & 0.016/1.0/0.031  & 0.119/1.0/0.213  & 0.859/1.0/0.924$^{\triangledown}$  & 0.197/0.93/0.325 \\
& $d_{aggr}$ & \textbf{0.928/1.0/0.962}  & \textbf{0.928/1.0/0.962}  & \textbf{0.928/1.0/0.962}  & 0.019/1.0/0.038$^{\circleddash}$  & 0.079/1.0/0.146$^{\circleddash}$  & \textbf{0.98/1.0/0.99}  & \textbf{0.813/0.996/0.895} \\
& $d_{cont}$ & 0.723/1.0/0.839  & 0.723/1.0/0.839  & 0.723/1.0/0.839  & \textbf{0.985/1.0/0.992}  & \textbf{0.993/1.0/0.996}  & 0.978/1.0/0.989  & \textbf{0.813/0.996/0.895} \\
\hline
\multirow{3}{*}{\makecell[c]{\textbf{PSM}}}
& $d_{disp}$ & \textbf{0.965/1.0/0.982}$^{\circ}$  & \textbf{0.965/1.0/0.982}$^{\circ}$  & \textbf{0.965/1.0/0.982}$^{\circ}$  & 0.077/1.0/0.143  & 0.237/1.0/0.383  & 0.775/1.0/0.874$^{\triangledown}$  & 0.406/0.948/0.568 \\
& $d_{aggr}$ & 0.\textbf{965/1.0/0.982}  & \textbf{0.965/1.0/0.982}  & \textbf{0.965/1.0/0.982}  & 0.094/1.0/0.171$^{\circleddash}$  & 0.365/1.0/0.535$^{\circleddash}$  & \textbf{0.962/1.0/0.981}  & \textbf{0.909/0.994/0.95} \\
& $d_{cont}$ & 0.853/1.0/0.921  & 0.853/1.0/0.921  & 0.853/1.0/0.921  & \textbf{0.985/1.0/0.993}  & \textbf{0.995/1.0/0.997}  & 0.957/1.0/0.978  & \textbf{0.909/0.994/0.95} \\
\hline
\multirow{3}{*}{\makecell[c]{\textbf{SMD}}}
& $d_{disp}$ & \textbf{0.81/1.0/0.895}$^{\circ}$  & \textbf{0.81/1.0/0.895}$^{\circ}$  & \textbf{0.81/1.0/0.895}$^{\circ}$  & 0.632/1.0/0.774  & 0.658/1.0/0.793  & 0.91/1.0/0.953$^{\triangledown}$  & 0.687/0.981/0.808 \\
& $d_{aggr}$ & \textbf{0.81/1.0/0.895}  & \textbf{0.81/1.0/0.895}  & \textbf{0.81/1.0/0.895}  & 0.674/1.0/0.805$^{\circleddash}$  & 0.677/1.0/0.808$^{\circleddash}$  & \textbf{0.99/1.0/0.995}  & \textbf{0.83/0.998/0.907} \\
& $d_{cont}$ & 0.459/1.0/0.629  & 0.459/1.0/0.629  & 0.459/1.0/0.629  & \textbf{0.992/1.0/0.996}  & \textbf{0.996/1.0/0.998}  & \textbf{0.99/1.0/0.995}  & 0.809/0.998/0.894 \\
\hline
\multirow{3}{*}{\makecell[c]{\textbf{SWaT}}}
& $d_{disp}$ & \textbf{0.924/1.0/0.96}$^{\circ}$  & \textbf{0.924/1.0/0.96}$^{\circ}$  & \textbf{0.924/1.0/0.96}$^{\circ}$  & 0.008/1.0/0.016  & 0.083/1.0/0.154  & 0.909/1.0/0.952$^{\triangledown}$  & 0.172/0.951/0.292 \\
& $d_{aggr}$ & \textbf{0.924/1.0/0.96}  & \textbf{0.924/1.0/0.96}  & \textbf{0.924/1.0/0.96}  & 0.01/1.0/0.02$^{\circleddash}$  & 0.352/1.0/0.521$^{\circleddash}$  & \textbf{0.948/1.0/0.973}  & \textbf{0.851/0.993/0.916} \\
& $d_{cont}$ & 0.752/1.0/0.858  & 0.752/1.0/0.858  & 0.752/1.0/0.858  & \textbf{0.965/1.0/0.982}  & \textbf{0.989/1.0/0.994}  & 0.905/1.0/0.95  & \textbf{0.851/0.993/0.916} \\
\hline
\multicolumn{9}{l}{Lack of beneficial characteristic: $^{\circleddash}$lack of fragments merging. Misleading characteristics: $^{\circ}$long anomaly misleading, $^{\triangledown}$sparse anomaly misleading.} \\
\hline
\end{tabular}
\label{table_8}
\end{table*}

\subsubsection{Experimental Results on Real-world Datasets}
First, we obtain the TAD results of three advanced detectors, Autoformer \cite{Autoformer}, DLinear \cite{DLinear}, and Timesnet \cite{Timesnet}, on all real-world datasets. These results are evaluated using different evaluators and ranked by f1-score, as presented in Table \ref{table_6}. Inappropriate evaluators may affect the ranking of detectors due to either the lack of beneficial characteristics or the presence of misleading ones. In Table \ref{table_6}, we uniformly mark the relevant characteristics where inappropriate evaluators yield misleadingly overestimated rankings. Corresponding demonstrations are provided in the four subplots of Fig. \ref{fig_6}. Point-based evaluators (\textit{PW}, \textit{PA}, \textit{PA\%K}) are mainly affected by the characteristic of long anomaly misleading, thus overestimating the ranking of Timesnet on the MSL, SMAP, PSM, and SWaT datasets. In these experiments, Timesnet generates more FP points than the other two detectors, severely compromising its detection performance, as shown in Fig. \ref{fig_6a} and Fig. \ref{fig_6c}. Point-based evaluators fail to attach sufficient importance to this flaw and instead prioritize the detection of long anomaly events, thereby leading to an overestimated ranking of Timesnet. On the IOPS dataset, Autoformer misses more anomaly events than DLinear, as shown in Fig. \ref{fig_6b}. However, due to the lack of existence detection reward, the \textit{PW} and \textit{PA\%K} evaluators do not prominently reflect these missed detections in the recall metric, resulting in the overestimated ranking of Autoformer. Moreover, the lack of fragment merging characteristic of event-based evaluators (\textit{RP/RR}, \textit{TaPR}, and \textit{AM}) was evident on the SWaT dataset: these evaluators neglect that the fragmented and scattered FP points in Timesnet’s detection results have obscured the TP points (see Fig. \ref{fig_6c}), thus overestimating its ranking. Besides, the \textit{AM} evaluator considers the FP points relatively close to the ground truth anomaly events beneficial rather than detrimental to the f1-score (i.e., the characteristic of sparse anomaly misleading), which results in its ranking deviations across most datasets. Finally, by incorporating all beneficial characteristics and avoiding misleading ones, \textit{OIPR} attains the most rational evaluation ranking across all real-world datasets, without any of the four aforementioned misoverestimation phenomena.

In the second group of experiments, we compare two adversary detectors $d_l$ and $d_{fp}$ across the MSL, SMAP, PSM, SMD and SWaT datasets to analyze the impact of the long anomaly effect, as shown in Table \ref{table_7}. Two point-based evaluators, \textit{PW} and \textit{PA\%K}, are significantly affected  due to the lack of existence detection reward and the presence of long anomaly misleading, and they favor $d_l$ over $d_{fp}$ across all five datasets. Due to the lack of overlapping proportion awareness, \textit{PA} incorrectly regards $d_{fp}$ as ideal and consistently overestimates its f1-score as 1. The event-based evaluators (\textit{RP/RR}, \textit{TaPR}, and \textit{AM}) completely eliminate the impact of the long anomaly misleading, and prefer $d_{fp}$ to $d_l$ on all datasets. As for \textit{OIPR}, it mitigates rather than fully eliminates this impact, thereby deems $d_{fp}$ superior on MSL and SMD while favoring $d_l$ on SMAP, PSM, and SWaT. We argue this confirms \textit{OIPR} as the only evaluator sensitive to striking a balance between the detection of longer and more anomalies. The underlying rationale is that when all anomaly events are relatively short in duration, the total number of events becomes a critical factor, suggesting that a detector capable of identifying all anomalies suffices. In contrast, for anomalies with drastically prolonged durations (e.g., the SWaT dataset, where the longest anomaly lasts 10 hours and the shortest merely 100 seconds), assigning greater importance to the former is more reasonable than treating these two events as entirely equivalent. We discuss the threshold of long/short anomalies in Appendix C, and present the parameter sensitivity analysis in Appendix D.

\begin{table}[!t]
\renewcommand{\arraystretch}{1.2}
\caption{Additional statistics for the real-world datasets. $R_N$ represents the ratio of contaminated normal intervals.}
\setlength{\tabcolsep}{0pt}
\centering
\begin{tabular}{ p{120pt}<{\centering} | p{25pt}<{\centering} p{27pt}<{\centering} p{25pt}<{\centering} p{25pt}<{\centering} p{27pt}<{\centering}}
\hline
\renewcommand{\arraystretch}{1.0}
\diagbox[width=118pt]{
  \hspace{20pt}\textbf{Statistics}
}{
  \textbf{Dataset}\hspace{20pt}
}
& \textbf{MSL} & \textbf{SMAP} & \textbf{PSM} & \textbf{SMD} & \textbf{SWaT}\\
\hline
\renewcommand{\arraystretch}{1.2}
Size of dataset $N$                             & 73630 & 427518 & 87742 & 7084  & 449820\\
Number of anomaly events $N_a$                  & 36    & 67     & 72    & 118   & 35    \\
Average length of $gt$ events $\overline{L}_a$  & 216   & 817    & 338   & 3     & 1561  \\
\hline
Long anomaly threshold $L$ for $d_l$ & 540   & 2043   & 845   & 4     & 3903  \\
Long anomaly event radio $R_e(L)$    & 0.111 & 0.179  & 0.069 & 0.203 & 0.029 \\
Long anomaly point radio $R_p(L)$    & 0.46  & 0.704  & 0.781 & 0.572 & 0.657 \\
\hline
$R_N$ for $d_{disp}$    & 0.946 & 1.0    & 0.694 & 0.345 & 0.972 \\
$R_N$ for $d_{aggr}$    & 0.108 & 0.059  & 0.083 & 0.017 & 0.25  \\
$R_N$ for $d_{cont}$    & 0.108 & 0.059  & 0.083 & 0.017 & 0.25  \\
\hline
\end{tabular}
\label{table_9}
\end{table}

In the third group of experiments, we compare three adversary detectors $d_{disp}$, $d_{aggr}$, and $d_{cont}$ to assess the impact of the fragmentation effect, as shown in Table \ref{table_8}. To this end, we introduce a metric called the ratio of normal intervals containing FP points, denoted as $R_N$, which quantifies the proportion of normal intervals contaminated by FP points relative to the total number of normal intervals. A normal interval is defined as a non-anomalous interval between two adjacent anomaly events in the ground truth. The $R_N$ values and other statistical data for experiments are presented in Table \ref{table_9}. A higher $R_N$ means operators are more likely to face disturbances induced by false alarms during routine operations. Notably, $d_{disp}$ yields an extremely high $R_N$, as it causes a substantial number of contaminated normal intervals. However, all point-based evaluators (\textit{PW}, \textit{PA}, and \textit{PA\%K}) erroneously judge $d_{disp}$ as satisfactory due to the presence of long anomaly misleading. Additionally, $d_{aggr}$ has the same low $R_N$ as $d_{cont}$, indicating few contaminated normal intervals and that the two detectors deliver comparable, solid performance. Nevertheless, \textit{RP/RR} and \textit{TaPR} lack the beneficial characteristic of fragment merging, thereby misjudging $d_{aggr}$ as underperforming. Besides, the sparse anomaly misleading characteristic causes the AM evaluator to erroneously rate $d_{disp}$ favorably. Ultimately, only \textit{OIPR} successfully identifies $d_{disp}$ as a subpar detector and recognizes $d_{aggr}$ and $d_{cont}$ as high-performing ones, as it takes into account the actual distribution of discrete FP points.

\section{Discussion}
In this work, the proposed \textit{OIPR} is characterized as an ``area-based'' TAD evaluator. To investigate its correlations and distinctions from existing point-based and event-based evaluators, we introduce two specific custom configurations of \textit{OIPR}, which can be interpreted as point-based and event-based evaluators, respectively.

\textbf{Zero observation phase.} In this configuration, we set $l_{obs}=0$ in Algorithm \ref{al_1}, indicating that there is no observation phase associated with the anomaly points. As a result, the operator interest in each anomaly point does not extend to the subsequent time point. It is evident that, in such a scenario, the operator interest curve of the ground truth and that of the detection results can be calculated as:
\begin{equation}
    \bm{I}=\bm{y}, \, \hat{\bm{I}}=\hat{\bm{y}}.
\label{eq_10}
\end{equation}

Hence, \textit{OIPR} degenerates into a point-based evaluator, which is essentially equivalent to the classical \textit{PW} evaluator.

\textbf{Strict occurrence detection evaluator.} By setting $l_{dis}=0$, $b_{dur}=0$, and $l_{obs}=1$ in Algorithm \ref{al_1}, \textit{OIPR} can be converted into a distinctive event-based evaluator: here, a predicted anomaly event is classified as a TP event only when its initial point coincides with the initial point of a ground truth event. Conversely, any ground truth event whose initial point does not align with that of a prediction event is classified as an FN event, while any prediction event whose initial point does not correspond to that of a ground truth event is categorized as an FP event. Although this evaluator may appear excessively stringent, it effectively demonstrates that \textit{OIPR} can be transformed into an event-based evaluator through specific parameter configurations.

Through the specific configurations discussed above, it can be observed that \textit{OIPR} lies between the point-based and event-based evaluators. It employs the observation phase to mitigate the long anomaly effect and to merge potential fragmented events. Additionally, the use of area in calculating the evaluation metrics enables \textit{OIPR} to effectively bridge the gap between point-based and event-based perspectives, thereby enhancing its versatility and applicability.

\section{Conclusions}
Given the key role of TAD in data analysis, numerous studies have focused on improving anomaly detector performance to identify anomalous behaviors and potential system faults. When evaluating these detectors, selecting an appropriate evaluator is critical: it helps operators choose optimal detectors and avoids misleading researchers into suboptimal optimization. In this work, we developed a novel TAD evaluator, \textit{OIPR}, which is inspired by the interest of operators in monitoring KPIs and associated detectors. Compared with existing evaluators, \textit{OIPR} has fewer limitations, adapts to diverse scenarios, and allows for smooth transitions between point-based and event-based paradigms via custom configuration to balance the two perspectives. We also introduced a special scenario dataset, which is carefully designed to highlight the characteristics and limitations of different evaluators. The superiority of \textit{OIPR} is verified through experiments on the special scenario dataset alongside several real-world datasets.

\bibliography{main.bib}
\bibliographystyle{IEEEtran}

\vspace{-10mm}
\begin{IEEEbiography}[{\includegraphics[width=1in,height=1.25in, clip,keepaspectratio]{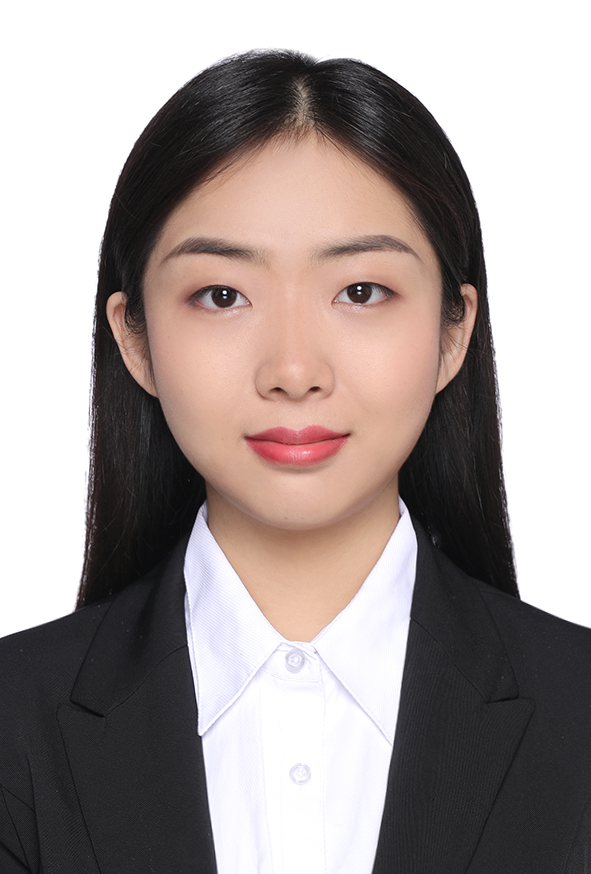}}]{Yuhan Jing} obtained her master's degree from Beijing University of Posts and Telecommunications, China, in 2020. She is currently a doctoral candidate at the State Key Laboratory of Networking and Switching Technology at the Beijing University of Posts and Telecommunications. Her research interests include AIOps, Time-series Analysis, Anomaly Detection, and Fault Localization.
\end{IEEEbiography}

\vspace{-10mm}
\begin{IEEEbiography}[{\includegraphics[width=1in,height=1.25in, clip,keepaspectratio]{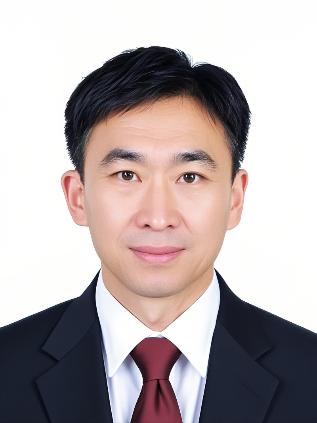}}]{Jingyu Wang} received his Ph.D. degree from the Beijing University of Posts and Telecommunications, Beijing, China, in 2008. He is currently a Tenured Professor with the State Key Laboratory of Networking and Switching Technology, Beijing University of Posts and Telecommunications. He is selected for the Yangtse River Scholar Award Program by the Ministry of Education. He has published more than 200 papers in such as the ToN, TMC, JSAC, NSDI, ASPLOS and so on. His research interests include broad aspects of Intelligent Networks, Edge/Cloud Computing, Machine Learning, Self-Driving Network, IoV/IoT, Knowledge-Defined Network and Intent-Driven Networking.
\end{IEEEbiography}

\vspace{-10mm}
\begin{IEEEbiography}[{\includegraphics[width=1in,height=1.25in, clip,keepaspectratio]{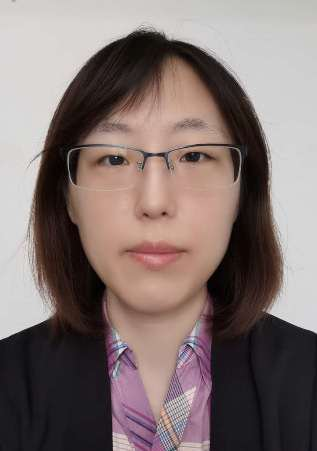}}]{Lei Zhang} received her master degree from Nanjing University of Posts and Telecommunications in 2004. She is the technical expert of department of Cloud Network Center at China Unicom, and the leader of Digital twin Project. Her research interest covers mobile network, Network Management, AIOps, Digital twin etc.
\end{IEEEbiography}

\vspace{-10mm}
\begin{IEEEbiography}[{\includegraphics[width=1in,height=1.25in,clip,keepaspectratio]{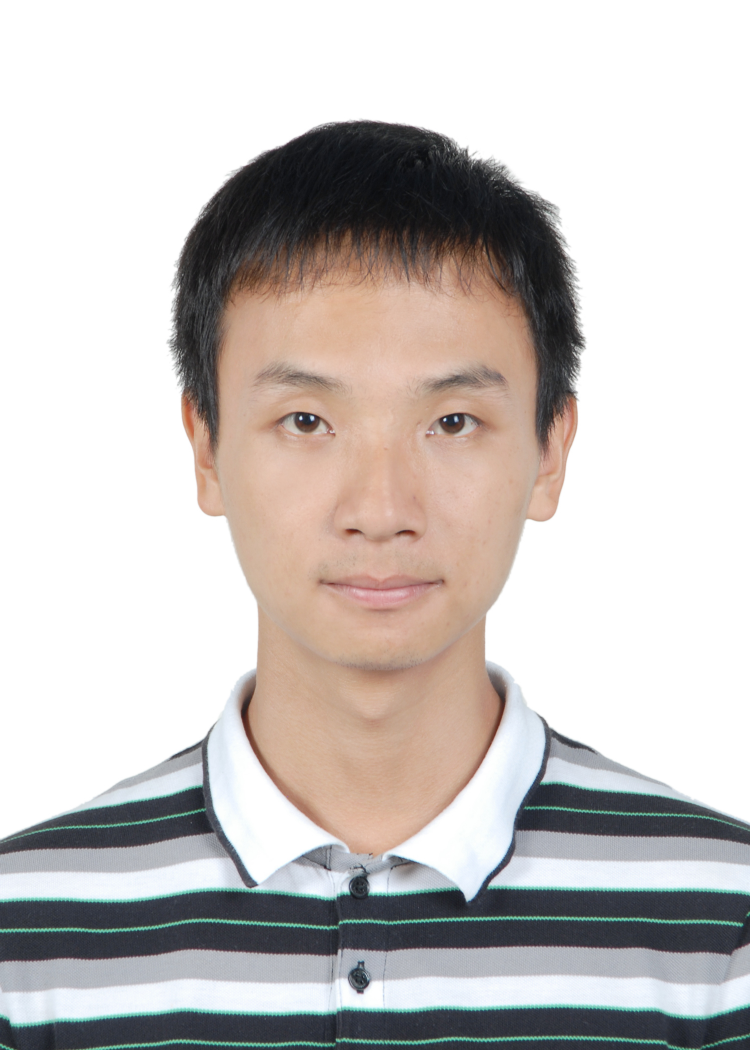}}]{Haifeng Sun} received the Ph.D. degree from the Beijing University of Posts and Telecommunications, Beijing, China, in 2017. He is currently an associate professor with the State Key Laboratory of Networking and Switching Technology, Beijing University of Posts and elecommunications. His research interests include broad aspects of AI, NLP, big data analysis, object detection, deep learning and pattern recognition.
\end{IEEEbiography}

\vspace{-10mm}
\begin{IEEEbiography}[{\includegraphics[width=1in,height=1.25in, clip,keepaspectratio]{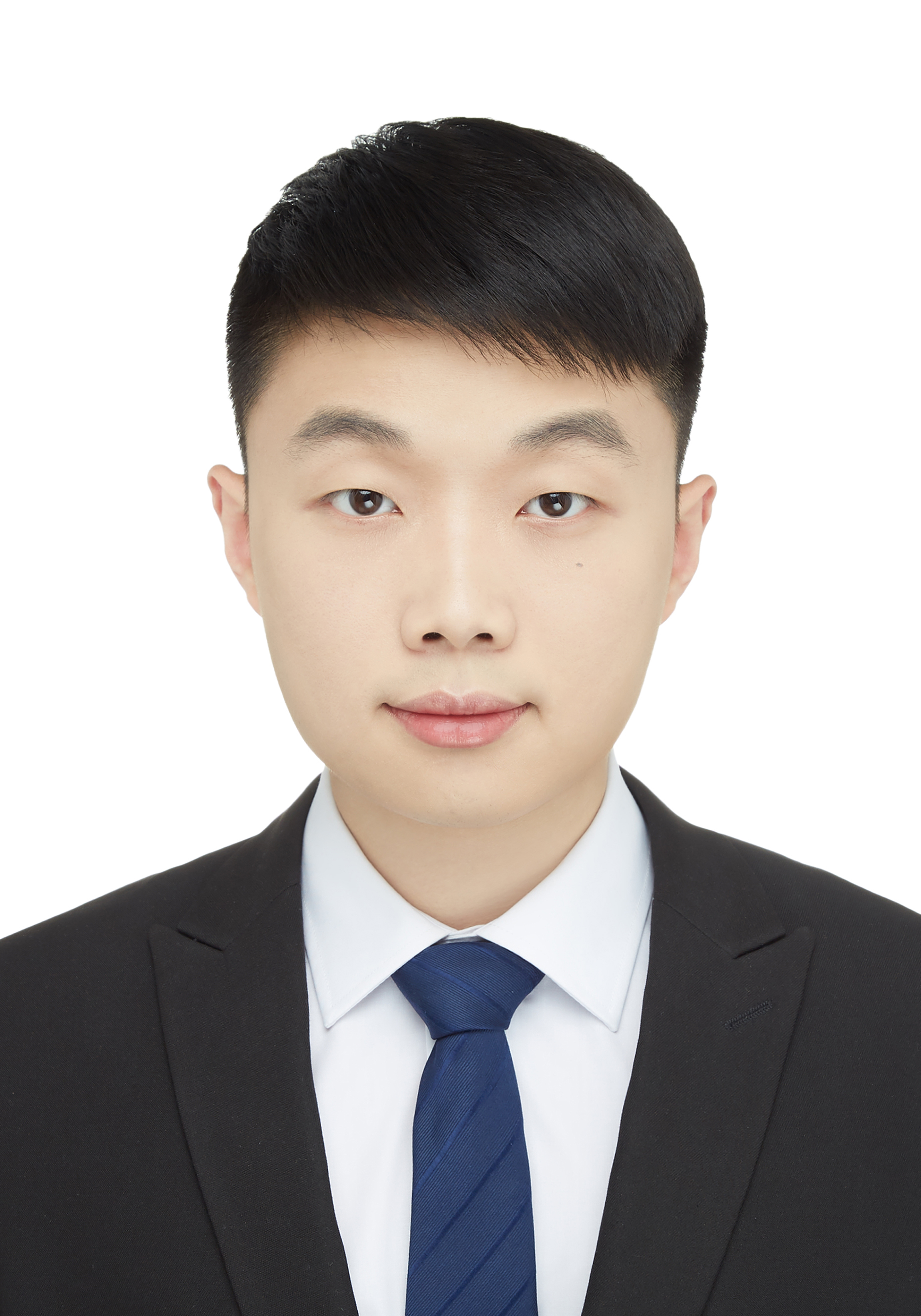}}]{Bo He} obtained his PhD degree from Beijing University of Posts and Telecommunications, China, in 2023. He is currently an associate researcher with the State Key Laboratory of Networking and Switching Technology, Beijing University of Posts and Telecommunications. From 2021 to 2022, he was a visiting PhD student at the University of Waterloo, Canada. His research interests include 5G/6G networks, multipath networks, collective communication, transmission control, and deep reinforcement learning.
\end{IEEEbiography}

\vspace{-10mm}
\begin{IEEEbiography}[{\includegraphics[width=1in,height=1.25in, clip,keepaspectratio]{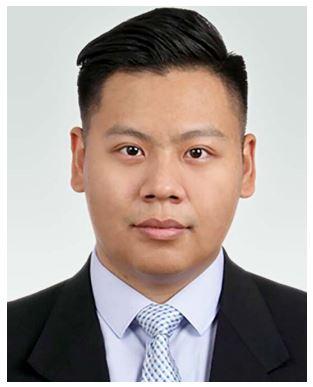}}]{Zirui Zhuang} received the B.S.and Ph.D. degrees from the Beijing University of Posts and Telecommunications in 2015 and 2020, respectively. He is currently a Post-Doctoral Researcher with the State Key Laboratory of Networking and Switching Technology, Beijing University of Posts and Telecommunications. In 2019, he visited the Department of Electrical and Computer Engineering, University of Houston. His research interests involve network routing and management for nextgeneration network infrastructures, using machine learning and artificial intelligence techniques, including deep learning, reinforcement learning, graph representation, multi-agent systems, and Lyapunovbased optimization.
\end{IEEEbiography}

\vspace{-10mm}
\begin{IEEEbiography}[{\includegraphics[width=1in,height=1.25in, clip,keepaspectratio]{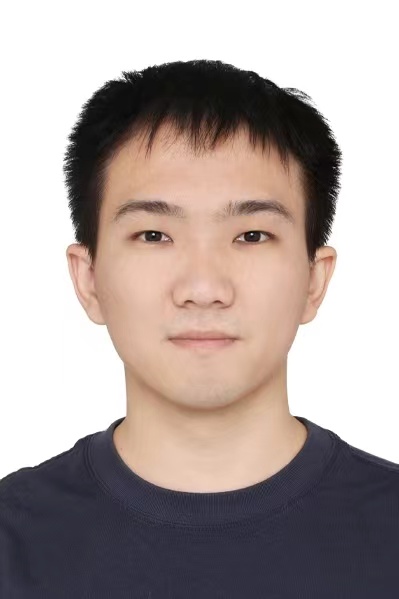}}]{Chengsen Wang} obtained his B.D. degree from Beijing University of Posts and Telecommunications in 2022. He is currently a doctoral candidate of State Key Laboratory of Networking and Switching Technology at Beijing University of Posts and Telecommunications. His main research interests include time-series analysis, anomaly detection, and multimodal learning.
\end{IEEEbiography}

\vspace{-10mm}
\begin{IEEEbiography}[{\includegraphics[width=1in,height=1.25in, clip,keepaspectratio]{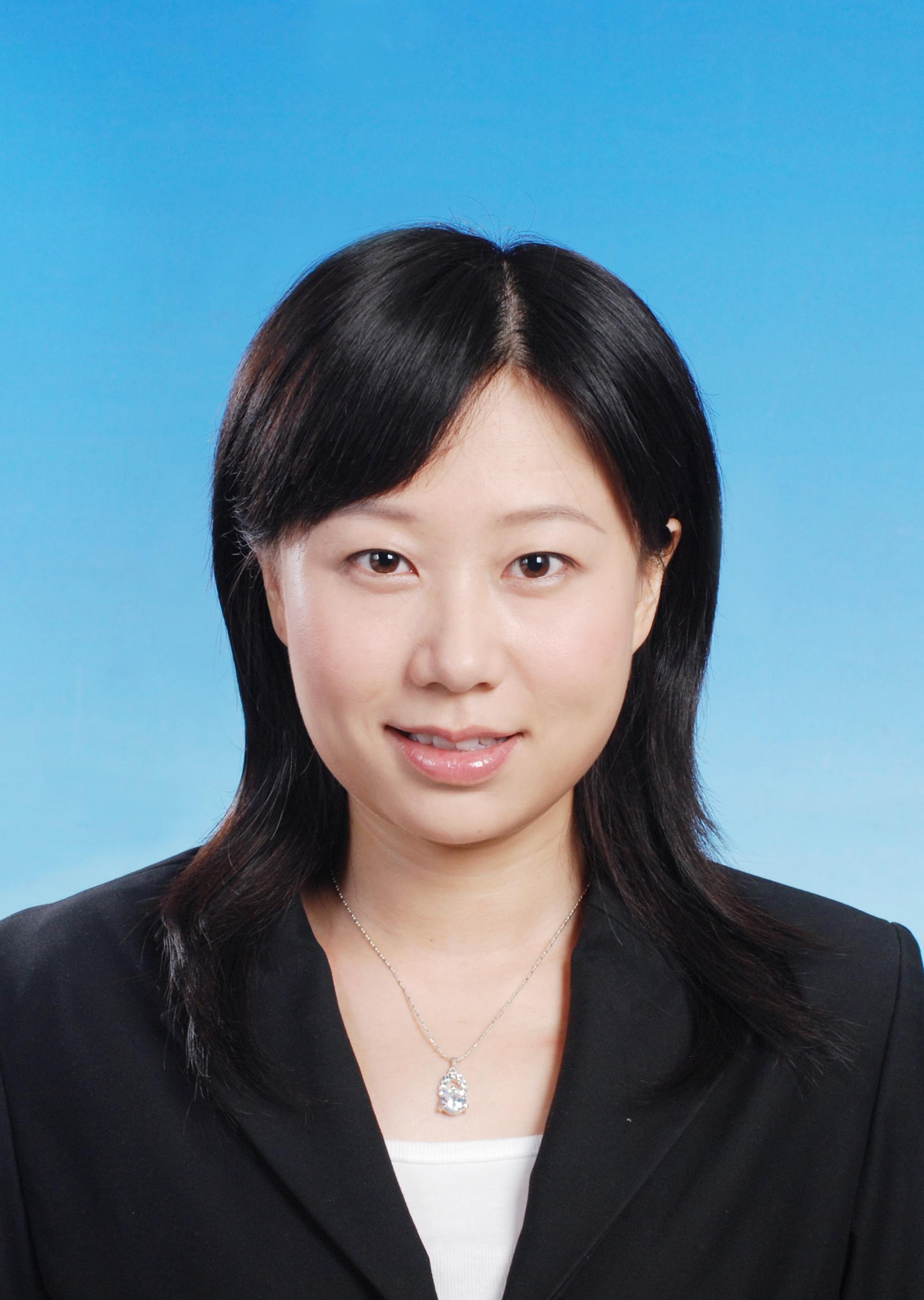}}]{Qi Qi} received the Ph.D. degree from the Beijing University of Posts and Telecommunications, Beijing, China, in 2010. She is currently a Professor with the State Key Laboratory of Networking and Switching Technology, Beijing University of Posts and Telecommunications. She has authored or co-authored more than 30 papers in the international journal and is the recipient of two National Natural Science Foundations of China. Her research interests include edge computing, cloud computing, the Internet of Things, ubiquitous services, deep learning, and deep reinforcement learning.
\end{IEEEbiography}

\vspace{-10mm}
\begin{IEEEbiography}[{\includegraphics[width=1in,height=1.25in,clip,keepaspectratio]{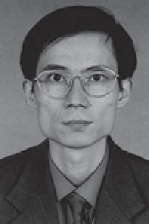}}]{Jianxin Liao} received the Ph.D. degree from the University of Electronics Science and Technology of China, Chengdu, China, in 1996. He is currently the Dean of the Network Intelligence Research Center and a Full Professor with the State Key Laboratory of Networking and Switching Technology, Beijing University of Posts and Telecommunications. He has authored or coauthored hundreds of research papers and several books. He has won several prizes in China for his research achievements, which include the Premiers Award of Distinguished Young Scientists from National Natural Science Foundation of China in 2005, and the specially invited Professor of the Yangtse River Scholar Award Program by the Ministry of Education in 2009. His main research interests include cloud computing, mobile intelligent network, service network intelligence, networking architectures and protocols, and multimedia communication.
\end{IEEEbiography}

\appendices
% \begin{appendices}
\section{Introduction and experimental results of the special scenario dataset}
% \appendix[Introduction and experimental results of the special scenario dataset]

\setcounter{table}{0}   %从0开始编号，显示出来表会A1开始编号
\setcounter{figure}{0}
% \setcounter{section}{0}
% \setcounter{equation}{0}
%定义编号格式，在数字序号前加字符“A"
\renewcommand{\thetable}{A\arabic{table}}
\renewcommand{\thefigure}{A\arabic{figure}}

\begin{figure*}[t]
	\centering
	\subfigure[Overlap proportion.] {\includegraphics[width=2.3in]{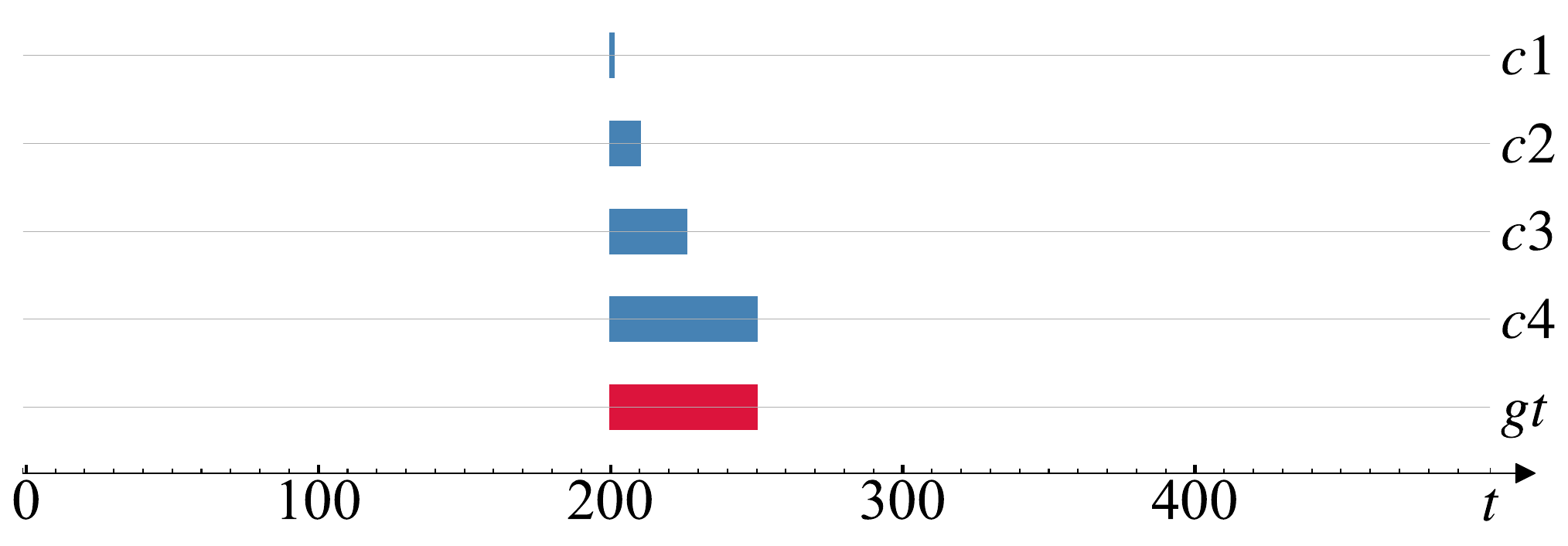}\label{fig_A1}}
	\subfigure[Fragmented TPs.] {\includegraphics[width=2.24in]{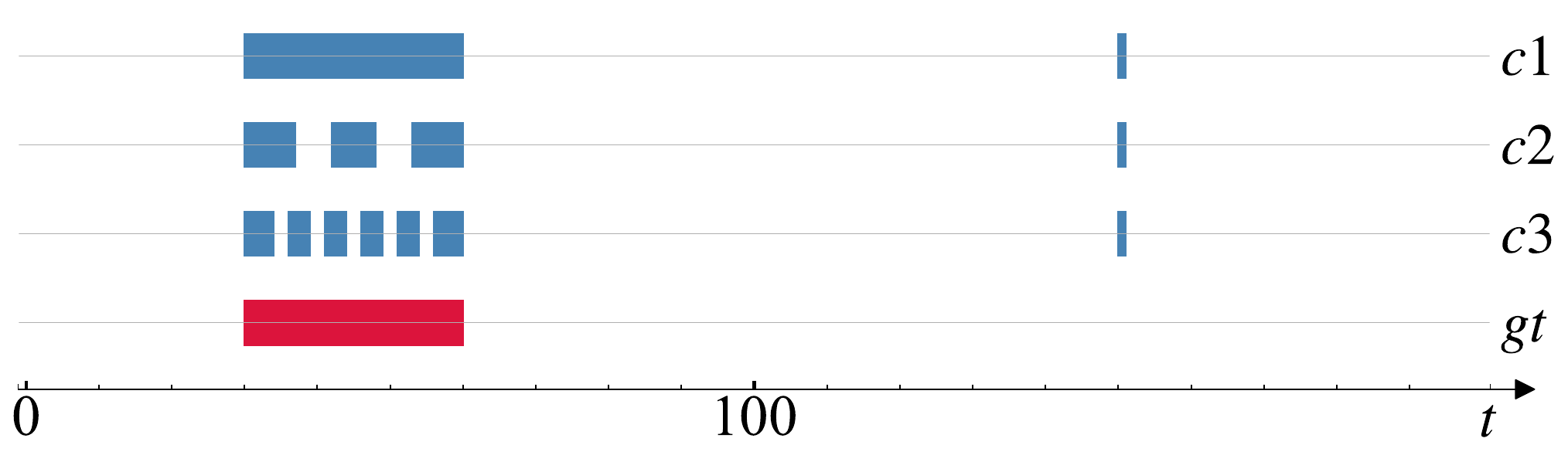}\label{fig_A2}}
	\subfigure[Fragmented FPs.] {\includegraphics[width=2.24in]{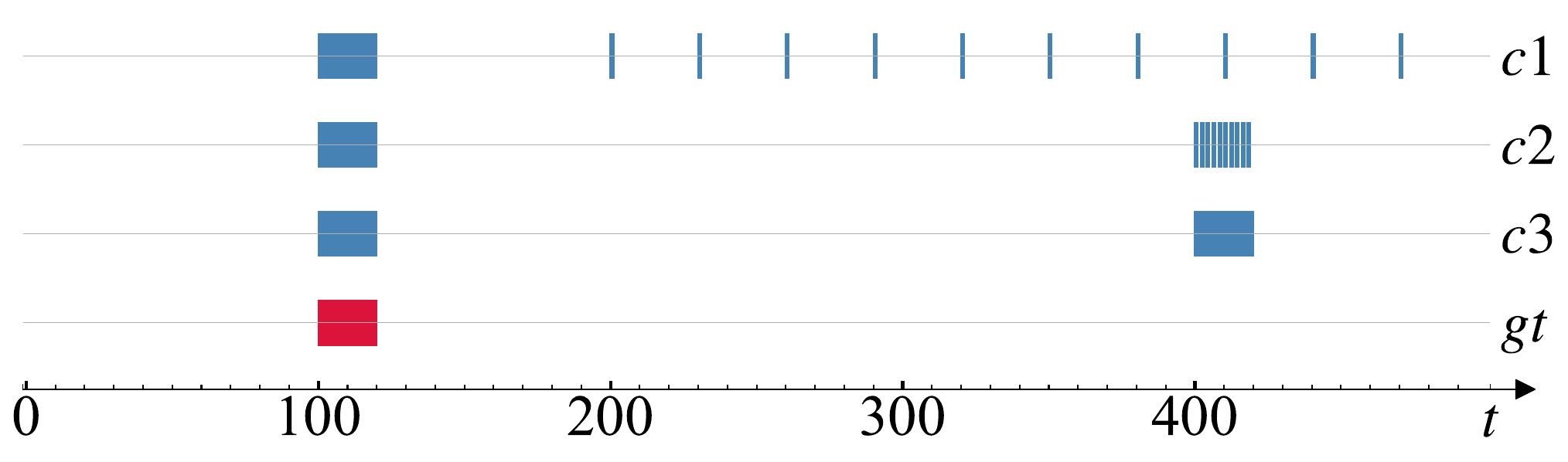}\label{fig_A3}}\\
        \vspace{-5pt} 
 
        \subfigure[Temporal shifting.] {\includegraphics[width=2.3in]{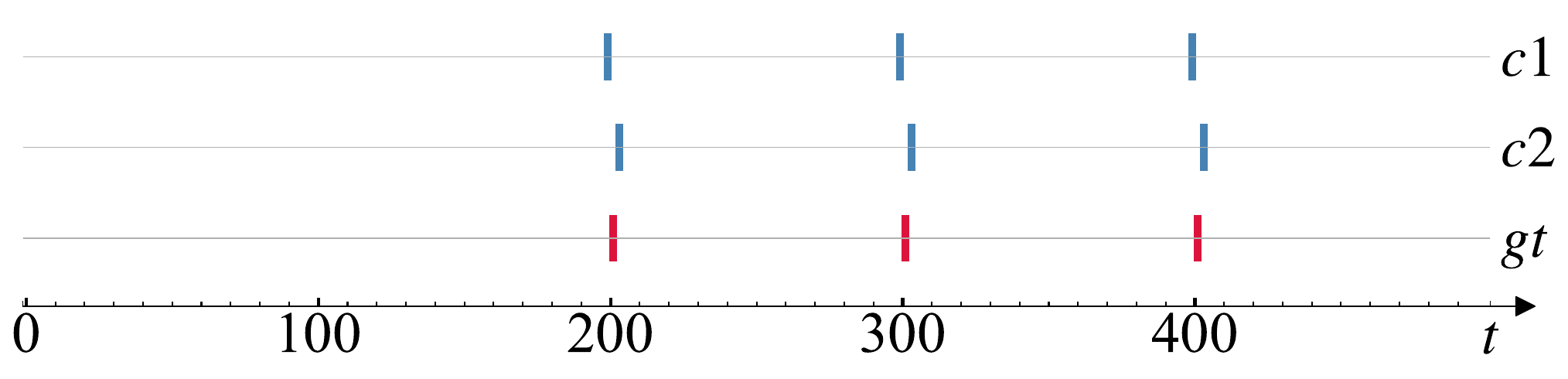}\label{fig_A4}}
        \subfigure[TP positions.] {\includegraphics[width=2.24in]{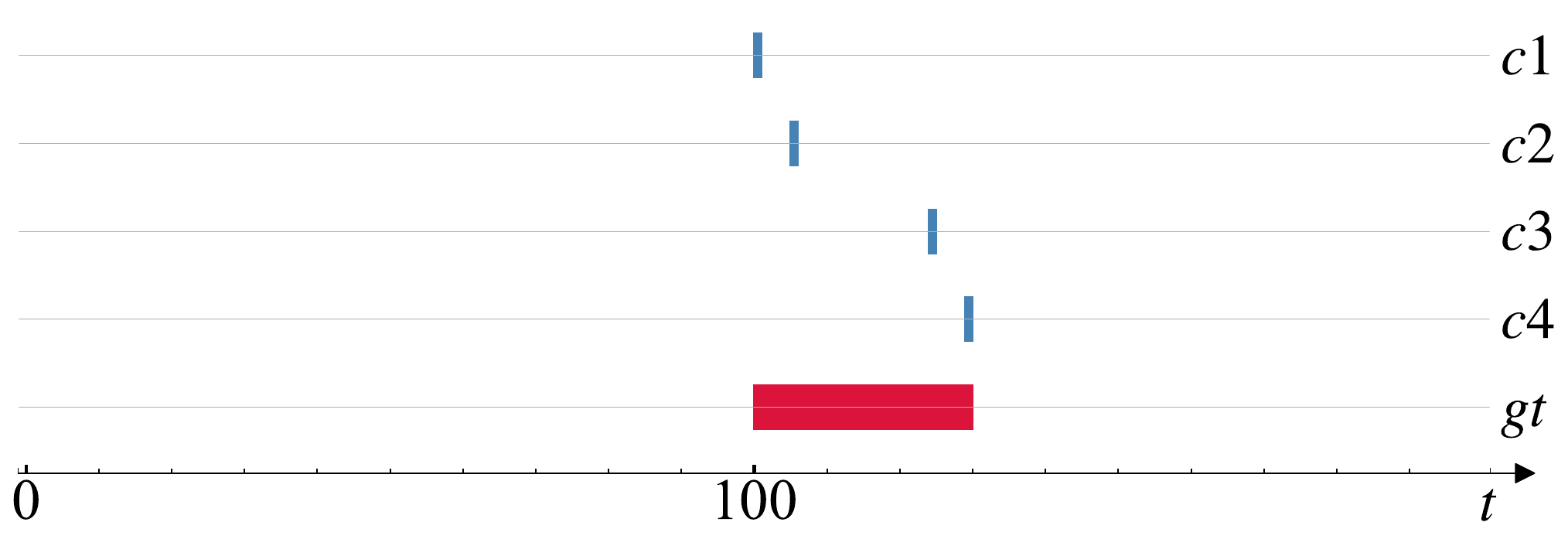}\label{fig_A5}}
        \subfigure[Long anomaly effect.] {\includegraphics[width=2.3in]{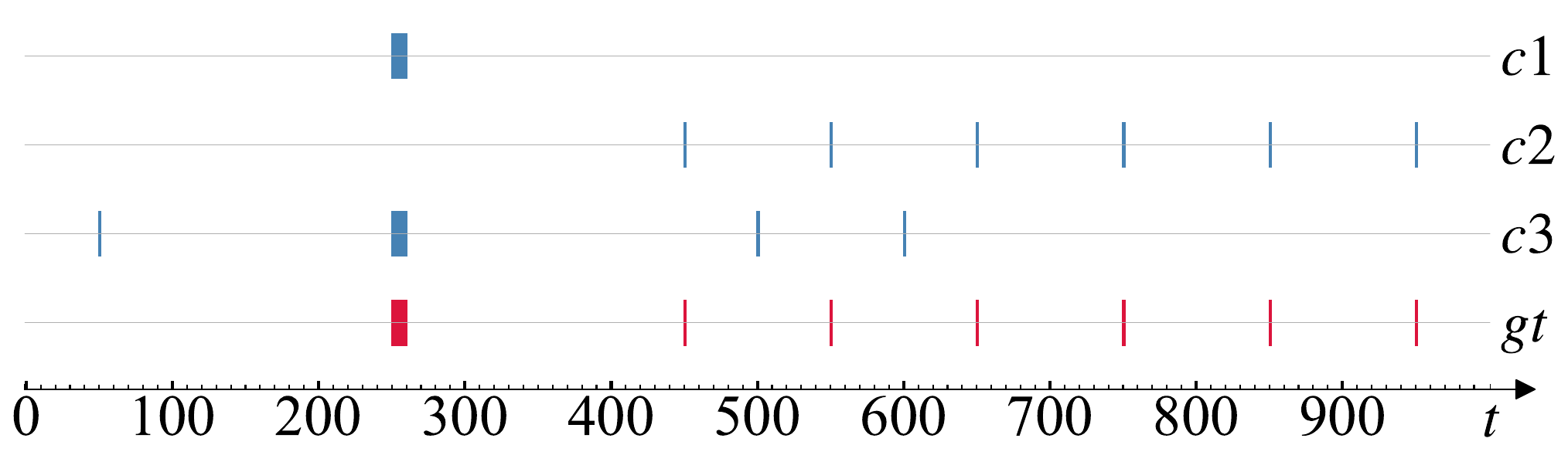}\label{fig_A6}}\\
        \vspace{-5pt} 

        \subfigure[Sparse anomalies.] {\includegraphics[width=2.3in]{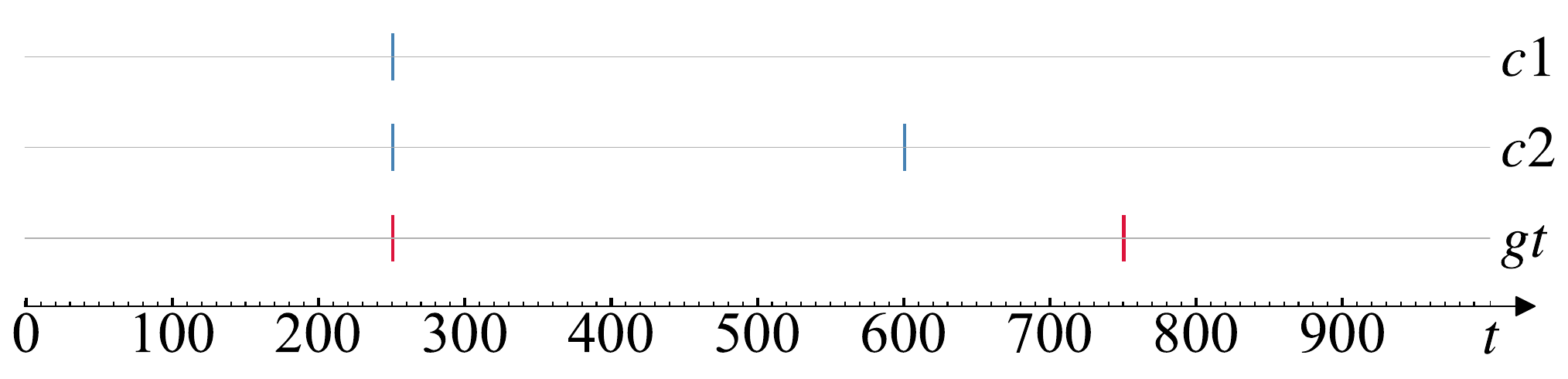}\label{fig_A7}}
        \subfigure[Constant detectors.] {\includegraphics[width=2.3in]{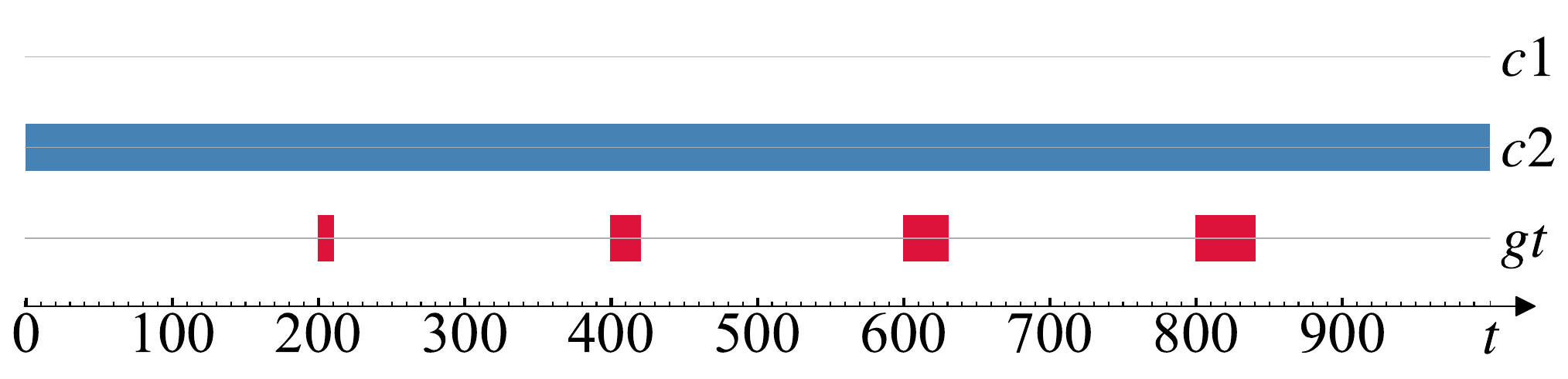}\label{fig_A8}}
        \subfigure[Random detector. (An example.)] {\includegraphics[width=2.3in]{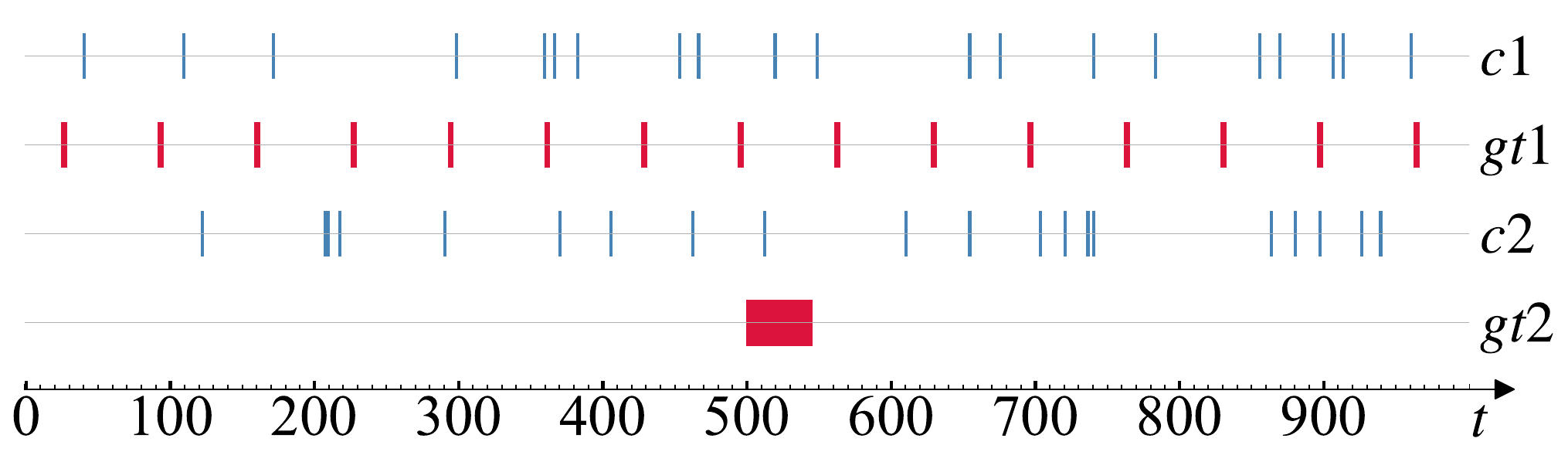}\label{fig_A9}}\\
    
	\caption{The visualization of the special scenario dataset. With the exception of the random detector scenario, all cases within a single scenario share the same ground truth (red bar) but yield different prediction results (blue bar).}
	\label{fig_A}
\end{figure*}

This appendix presents the introduction and experimental results of the special scenario dataset proposed in this work. It comprises nine distinct scenarios, each highlighting one or two evaluator characteristics. Visualizations of all special scenarios are presented in Fig. \ref{fig_A}, while the experimental results for different evaluators are summarized in Table \ref{table_A1}. The detailed descriptions of the special scenarios are as follows:

\textbf{Overlap proportion.} The ground truth comprises 50-point anomaly event; predicted anomaly proportions differ across cases \textit{c1}-\textit{c4}: the first point only, the initial 20\%, 52\%, and 100\% of the points, respectively. Evaluators that yield a higher f1-score than \textit{PW} for \textit{c1} exhibit the characteristic of existence detection reward. Additionally, those able to differentiate \textit{c2}-\textit{c4} based on f1-scores demonstrate the characteristic of overlapping proportion awareness.

\textbf{Fragmented TPs.} The ground truth consists of a 30-point anomaly event, detected in varying completeness across three cases, each containing an FP point. In \textit{c1}-\textit{c3}, the ground truth event is detected as 1 complete event, 3 fragmented events, and 6 fragmented events, respectively. Both \textit{c2} and \textit{c3} encompass a total of 20 TP points. Evaluators with the characteristic of fragmentation misleading in precision yield higher precision scores for \textit{c2} and \textit{c3} than for \textit{c1}.

\textbf{Fragmented FPs.} The ground truth consists of a 20-point anomaly event, which is completely detected. In cases \textit{c1}-\textit{c3}, varying sets of FP points are introduced into the prediction results: 10 dispersed FP points at intervals of 30, 10 aggregated FP points at intervals of 2, and 20 continuous FP points, respectively. Evaluators exhibiting the characteristic of fragments merging assign a significantly lower f1-score to \textit{c1} compared to \textit{c2} and \textit{c3}.

\textbf{Temporal shifting.} The ground truth encompasses 3 anomaly events, each consisting of 2 anomaly points. In \textit{c1}, a virtual early detector identifies anomalies 2 points prior to each ground truth event. In contrast, in \textit{c2}, a virtual delayed detector with a delay of 2 points is employed. Evaluators with the characteristic of addressing ambiguous labels yield f1-scores higher than 0 in both cases.

\textbf{TP positions.} 
The ground truth comprises a 30-point anomaly event. In cases \textit{c1}-\textit{c4}, the anomaly event is detected at the first, 6th, 25th and last points, respectively. Evaluators with the characteristic of early detection reward assign the highest f1-score to \textit{c1}, followed by \textit{c2}, \textit{c3}, and \textit{c4}.

\textbf{Long anomaly effect.} The ground truth has one long anomaly (10 points) and 6 short anomalies (1 point each). In cases \textit{c1}-\textit{c3}, the prediction results encompass the long anomaly, the 6 short anomalies, and the long anomaly accompanied by 3 FP points, respectively. Evaluators with the characteristic of long anomaly misleading exhibit a much higher f1-score for \textit{c1} compared to \textit{c2}, while the f1-score for \textit{c3} is slightly lower than that of \textit{c1}.

\textbf{Sparse anomalies.} The ground truth has two widely spaced anomaly events; one is correctly detected, the other missed, in both \textit{c1} and \textit{c2}. \textit{c2} has an extra FP point, \textit{c1} does not. The \textit{AM} evaluator mistakenly treats this FP point as beneficial for f1-score, due to the mapping of absolute distance to relative recall distance. Notably, this characteristic of sparse anomaly misleading is not observed in other evaluators.

\textbf{Constant detectors.} 
The ground truth consists of 4 anomaly events with lengths of 10, 20, 30, and 40, respectively. Two ineffective constant detectors are evaluated: the all\_0 detector (\textit{c1}), which outputs 0 at all time points, and the all\_1 detector (\textit{c2}), which outputs 1 consistently. Despite an f1-score of 0.6724 indicating the characteristic of overestimation for the all\_1 detector in \textit{AM}, this systematic overestimation is not considered to affect \textit{AM}’s performance ranking across detectors. The authors of \textit{AM} have stated that its baseline f1-score is 0.5, reflecting the expected value of the random detector, and detectors not significantly exceeding this are ineffective. \textit{TaPR} also shows the characteristic of overestimation for the all\_1 detector with an f1-score of 0.7138 due to its custom custom configuration. Users can mitigate this overestimation by lowering the detection score weight (minimum 0), but this adjustment sacrifice the beneficial characteristic of existence detection reward, highlighting an inherent trade-off between these two characteristics in \textit{TaPR}.

\textbf{Random detector.} The random detector generates anomalies with a probability of 0.02 at each timestamp over a 1000-point time-series. In \textit{c1}, the ground truth comprises 15 short anomaly events (3 points each). Conversely, \textit{c2} contains one long anomaly event (45 points). \textit{PA} shows significant overestimation for the random detector on \textit{c2}, with an average f1-score of 0.4428 across 100 experiments. For \textit{AM}, the overestimation for the random detector is predictable and systematic, similar to that seen in the all\_1 detector case. Due to the dense distribution of the 15 anomalies, \textit{OIPR}’s f1-score in \textit{c1} is higher than that in \textit{c2}. With the parameter setup of $l_{obs} = 20$, the observation phases of 15 short anomalies partially overlap. It is recommended to adjust $l_{obs}$ based on the context of the average anomaly length during practical applications, though we did not implement this adjustment to maintain the simplicity of the experiment.

The above special scenarios reveal critical boundary conditions that affect the evaluation outcomes. In practice, such special scenarios often occur in specific dataset slices. Since operators rarely have time to assess the detailed performance of each detector in all slices, they rely on general feedback from the evaluator to choose the best detector. As a result, the evaluator pitfalls can still be difficult to discern. To solve this, we propose this purpose-designed dataset that enables low-cost testing of evaluator characteristics. It helps operators select the most suitable evaluator while gaining insight into its limitations. For researchers, a well-defined dataset promotes experimentation with more effective evaluators.

\begin{table*}[!t]
\renewcommand{\arraystretch}{1.23}
\caption{Experimental results (P/R/F1) on the special scenario dataset. Beneficial characteristics of evaluators are marked with a star ($\star$) or a dagger ($\dagger$), while misleading characteristics are indicated with a diamond ($\diamond$).}
\setlength{\tabcolsep}{1pt}
% \fontsize{7pt}{9pt}\selectfont
\scriptsize
\centering
\begin{tabular}{p{35pt}<{\centering} p{14pt}<{\centering} p{70pt}<{\raggedright} p{70pt}<{\raggedright} p{70pt}<{\raggedright} p{70pt}<{\raggedright} p{70pt}<{\raggedright} p{70pt}<{\raggedright} p{70pt}<{\raggedright}}
\hline
\bfseries Scenario & \bfseries Case & \makecell[c]{\textbf{PW}} & \bfseries \makecell[c]{\textbf{PA}} & \makecell[c]{\textbf{PA\%K}} & \makecell[c]{\textbf{RP/RR}} & \makecell[c]{\textbf{TaPR}} & \bfseries \makecell[c]{\textbf{AM}} & \bfseries \makecell[c]{\textbf{OIPR (Ours)}} \\
\hline 
\multirow{4}{*}{\makecell[c]{\textbf{Overlap} \\ \textbf{proportion}}}
% & \textit{c1} & 1.0/0.02/0.039 & 1.0/1.0/1.0$^{\star}$ & 1.0/0.02/0.039 & 1.0/0.52/0.684$^{\star \! \dagger}$ & 1.0/0.51/0.675$^{\star \! \dagger}$ & 1.0/0.904/0.95$^{\star \! \dagger}$ & 1.0/0.217/0.356$^{\star \! \dagger}$ \\
% & \textit{c2} & 1.0/0.2/0.333$^{\dagger}$ & 1.0/1.0/1.0 & 1.0/0.2/0.333$^{\dagger}$ & 1.0/0.678/0.808$^{\star \! \dagger}$ & 1.0/0.6/0.75$^{\star \! \dagger}$ & 1.0/0.936/0.967$^{\star \! \dagger}$ & 1.0/0.361/0.53$^{\star \! \dagger}$ \\
% & \textit{c3} & 1.0/0.52/0.684$^{\dagger}$ & 1.0/1.0/1.0 & 1.0/1.0/1.0$^{\star}$ & 1.0/0.882/0.938$^{\star \! \dagger}$ & 1.0/0.76/0.864$^{\star \! \dagger}$ & 1.0/0.977/0.988$^{\star \! \dagger}$ & 1.0/0.617/0.763$^{\star \! \dagger}$ \\ 
% & \textit{c4} & 1.0/1.0/1.0$^{\dagger}$ & 1.0/1.0/1.0 & 1.0/1.0/1.0$^{\dagger}$ & 1.0/1.0/1.0$^{\star \! \dagger}$ & 1.0/1.0/1.0$^{\star \! \dagger}$ & 1.0/1.0/1.0$^{\star \! \dagger}$ & 1.0/1.0/1.0$^{\star \! \dagger}$ \\
& \textit{c1} & 1.0/0.02/0.0392 & 1.0/1.0/1.0$^{\star}$ & 1.0/0.02/0.0392 & 1.0/0.5196/0.6839$^{\star \! \dagger}$ & 1.0/0.51/0.6755$^{\star \! \dagger}$ & 1.0/0.904/0.9496$^{\star \! \dagger}$ & 1.0/0.2168/0.3564$^{\star \! \dagger}$ \\
& \textit{c2} & 1.0/0.2/0.3333$^{\dagger}$ & 1.0/1.0/1.0 & 1.0/0.2/0.3333$^{\dagger}$ & 1.0/0.6784/0.8084$^{\star \! \dagger}$ & 1.0/0.6/0.75$^{\star \! \dagger}$ & 1.0/0.936/0.9669$^{\star \! \dagger}$ & 1.0/0.3609/0.5304$^{\star \! \dagger}$ \\
& \textit{c3} & 1.0/0.52/0.6842$^{\dagger}$ & 1.0/1.0/1.0 & 1.0/1.0/1.0$^{\star}$ & 1.0/0.8824/0.9375$^{\star \! \dagger}$ & 1.0/0.76/0.8636$^{\star \! \dagger}$ & 1.0/0.977/0.9883$^{\star \! \dagger}$ & 1.0/0.6166/0.7628$^{\star \! \dagger}$ \\
& \textit{c4} & 1.0/1.0/1.0$^{\dagger}$ & 1.0/1.0/1.0 & 1.0/1.0/1.0$^{\dagger}$ & 1.0/1.0/1.0$^{\star \! \dagger}$ & 1.0/1.0/1.0$^{\star \! \dagger}$ & 1.0/1.0/1.0$^{\star \! \dagger}$ & 1.0/1.0/1.0$^{\star \! \dagger}$ \\
\hline 
\multicolumn{9}{l}{Evaluator characteristics: $^{\star}$existence detection reward, $^{\dagger}$overlapping proportion awareness.} \\
\hline
\multirow{3}{*}{\makecell[c]{\textbf{Fragmented} \\ \textbf{TPs}}}
% & \textit{c1} & 0.968/1.0/0.984   & 0.968/1.0/0.984 & 0.968/1.0/0.984 & 0.5/1.0/0.667                            & 0.5/1.0/0.667                 & 0.976/1.0/0.988   & 0.758/1.0/0.863 \\
% & \textit{c2} & 0.952/0.667/0.784 & 0.968/1.0/0.984 & 0.968/1.0/0.984 & 0.75/0.613/0.675$^{\diamond}$  & 0.75/0.833/0.789$^{\diamond}$ & 0.964/0.996/0.98  & 0.757/0.993/0.859 \\
& \textit{c1} & 0.9677/1.0/0.9836 & 0.9677/1.0/0.9836 & 0.9677/1.0/0.9836 & 0.5/1.0/0.6667 & 0.5/1.0/0.6667 & 0.9757/1.0/0.9877 & 0.7584/1.0/0.8626 \\
& \textit{c2} & 0.9524/0.6667/0.7843 & 0.9677/1.0/0.9836 & 0.9677/1.0/0.9836 & 0.75/0.6129/0.6746$^{\diamond}$ & 0.75/0.8333/0.7895$^{\diamond}$ & 0.9642/0.9958/0.9797 & 0.7571/0.993/0.8591 \\
& \textit{c3} & 0.9524/0.6667/0.7843 & 0.9677/1.0/0.9836 & 0.9677/1.0/0.9836 & 0.8571/0.5556/0.6742$^{\diamond}$ & 0.8571/0.8333/0.8451$^{\diamond}$ & 0.9642/0.9983/0.981 & 0.758/0.9982/0.8617 \\
\hline
\multicolumn{9}{l}{Evaluator characteristics:  $^{\diamond}$fragmentation misleading in precision.} \\

\hline
\multirow{3}{*}{\makecell[c]{\textbf{Fragmented} \\ \textbf{FPs}}}
% & \textit{c1} & 0.667/1.0/0.8 & 0.667/1.0/0.8 & 0.667/1.0/0.8 & 0.091/1.0/0.167 & 0.091/1.0/0.167 & 0.778/1.0/0.875 & 0.194/1.0/0.324$^{\star}$ \\
% & \textit{c2} & 0.667/1.0/0.8 & 0.667/1.0/0.8 & 0.667/1.0/0.8 & 0.091/1.0/0.167 & 0.091/1.0/0.167 & 0.727/1.0/0.842 & 0.508/1.0/0.674$^{\star}$ \\
% & \textit{c3} & 0.5/1.0/0.667 & 0.5/1.0/0.667 & 0.5/1.0/0.667 & 0.5/1.0/0.667   & 0.5/1.0/0.667   & 0.59/1.0/0.742  & 0.5/1.0/0.667$^{\star}$ \\
& \textit{c1} & 0.6667/1.0/0.8 & 0.6667/1.0/0.8 & 0.6667/1.0/0.8 & 0.0909/1.0/0.1667 & 0.0909/1.0/0.1667 & 0.7776/1.0/0.8749 & 0.1937/1.0/0.3245$^{\star}$ \\
& \textit{c2} & 0.6667/1.0/0.8 & 0.6667/1.0/0.8 & 0.6667/1.0/0.8 & 0.0909/1.0/0.1667 & 0.0909/1.0/0.1667 & 0.727/1.0/0.8419 & 0.5081/1.0/0.6739$^{\star}$ \\
& \textit{c3} & 0.5/1.0/0.6667 & 0.5/1.0/0.6667 & 0.5/1.0/0.6667 & 0.5/1.0/0.6667 & 0.5/1.0/0.6667 & 0.59/1.0/0.7421 & 0.5/1.0/0.6667$^{\star}$ \\
\hline 
\multicolumn{9}{l}{Evaluator characteristic: $^{\star}$fragments merging.} \\
\hline
\multirow{2}{*}{\makecell[c]{\textbf{Temporal} \\ \textbf{shifting}}}
% & \textit{c1} & 0.0/0.0/0.0 & 0.0/0.0/0.0 & 0.0/0.0/0.0 & 0.0/0.0/0.0 & 0.0/0.0/0.0              & 0.972/0.986/0.979$^{\star}$ & 0.729/0.729/0.729$^{\star}$ \\
% & \textit{c2} & 0.0/0.0/0.0 & 0.0/0.0/0.0 & 0.0/0.0/0.0 & 0.0/0.0/0.0 & 0.97/0.97/0.97$^{\star}$ & 0.972/0.986/0.979$^{\star}$ & 0.729/0.729/0.729$^{\star}$ \\
& \textit{c1} & 0.0/0.0/0.0 & 0.0/0.0/0.0 & 0.0/0.0/0.0 & 0.0/0.0/0.0 & 0.0/0.0/0.0 & 0.9724/0.9862/0.9793$^{\star}$ & 0.7285/0.7285/0.7285$^{\star}$ \\
& \textit{c2} & 0.0/0.0/0.0 & 0.0/0.0/0.0 & 0.0/0.0/0.0 & 0.0/0.0/0.0 & 0.9875/0.9875/0.9875$^{\star}$ & 0.9724/0.9862/0.9793$^{\star}$ & 0.7285/0.7285/0.7285$^{\star}$ \\
\hline
\multicolumn{9}{l}{Evaluator characteristic: $^{\star}$addressing ambiguous labels.} \\
\hline
\multirow{4}{*}{\makecell[c]{\textbf{TP} \\ \textbf{positions}}}
% & \textit{c1} & 1.0/0.033/0.065 & 1.0/1.0/1.0 & 1.0/0.033/0.065 & 1.0/0.532/0.695$^{\star}$ & 1.0/0.517/0.681 & 1.0/0.86/0.925 & 1.0/0.319/0.483$^{\star}$ \\
% & \textit{c2} & 1.0/0.033/0.065 & 1.0/1.0/1.0 & 1.0/0.033/0.065 & 1.0/0.527/0.69$^{\star}$ & 1.0/0.517/0.681 & 1.0/0.9/0.947 & 0.7859/0.2504/0.3798$^{\star}$ \\
% & \textit{c3} & 1.0/0.033/0.065 & 1.0/1.0/1.0 & 1.0/0.033/0.065 & 1.0/0.506/0.672$^{\star}$ & 1.0/0.517/0.681 & 1.0/0.9/0.947 & 0.7853/0.2502/0.3795$^{\star}$ \\
% & \textit{c4} & 1.0/0.033/0.065 & 1.0/1.0/1.0 & 1.0/0.033/0.065 & 1.0/0.501/0.668$^{\star}$ & 1.0/0.517/0.681 & 1.0/0.86/0.925 & 0.779/0.248/0.376$^{\star}$ \\
& \textit{c1} & 1.0/0.0333/0.0645 & 1.0/1.0/1.0 & 1.0/0.0333/0.0645 & 1.0/0.5323/0.6947$^{\star}$ & 1.0/0.5167/0.6813 & 1.0/0.8598/0.9246 & 1.0/0.3186/0.4833$^{\star}$ \\
& \textit{c2} & 1.0/0.0333/0.0645 & 1.0/1.0/1.0 & 1.0/0.0333/0.0645 & 1.0/0.5269/0.6901$^{\star}$ & 1.0/0.5167/0.6813 & 1.0/0.8998/0.9473 & 0.7859/0.2504/0.3798$^{\star}$ \\
& \textit{c3} & 1.0/0.0333/0.0645 & 1.0/1.0/1.0 & 1.0/0.0333/0.0645 & 1.0/0.5065/0.6724$^{\star}$ & 1.0/0.5167/0.6813 & 1.0/0.8998/0.9473 & 0.7853/0.2502/0.3795$^{\star}$ \\
& \textit{c4} & 1.0/0.0333/0.0645 & 1.0/1.0/1.0 & 1.0/0.0333/0.0645 & 1.0/0.5011/0.6676$^{\star}$& 1.0/0.5167/0.6813 & 1.0/0.8598/0.9246 & 0.7789/0.2482/0.3764$^{\star}$ \\
\hline 
\multicolumn{9}{l}{Evaluator characteristic: $^{\star}$early detection reward.} \\
\hline
\multirow{3}{*}{ \makecell[c]{\textbf{Long} \\ \textbf{anomaly} \\ \textbf{effect}} }
% & \textit{c1} & 1.0/0.625/0.769$^{\diamond}$  & 1.0/0.625/0.769$^{\diamond}$  & 1.0/0.625/0.769$^{\diamond}$  & 1.0/0.143/0.25   & 1.0/0.143/0.25   & 1.0/0.143/0.25    & 1.0/0.217/0.357 \\
% & \textit{c2} & 1.0/0.375/0.545$^{\diamond}$  & 1.0/0.375/0.545$^{\diamond}$  & 1.0/0.375/0.545$^{\diamond}$  & 1.0/0.857/0.923  & 1.0/0.857/0.923  & 1.0/0.857/0.923   & 1.0/0.783/0.878 \\
% & \textit{c3} & 0.769/0.625/0.69$^{\diamond}$ & 0.769/0.625/0.69$^{\diamond}$ & 0.769/0.625/0.69$^{\diamond}$ & 0.25/0.143/0.182 & 0.25/0.143/0.182 & 0.312/0.192/0.238 & 0.357/0.217/0.27 \\
& \textit{c1} & 1.0/0.625/0.7692$^{\diamond}$ & 1.0/0.625/0.7692$^{\diamond}$ & 1.0/0.625/0.7692$^{\diamond}$ & 1.0/0.1429/0.25 & 1.0/0.1429/0.25 & 1.0/0.1429/0.25 & 1.0/0.2172/0.3569 \\
& \textit{c2} & 1.0/0.375/0.5455$^{\diamond}$ & 1.0/0.375/0.5455$^{\diamond}$ & 1.0/0.375/0.5455$^{\diamond}$ & 1.0/0.8571/0.9231 & 1.0/0.8571/0.9231 & 1.0/0.8571/0.9231 & 1.0/0.7828/0.8782 \\
& \textit{c3} & 0.7692/0.625/0.6897$^{\diamond}$ & 0.7692/0.625/0.6897$^{\diamond}$ & 0.7692/0.625/0.6897$^{\diamond}$ & 0.25/0.1429/0.1818 & 0.25/0.1429/0.1818 & 0.312/0.1922/0.2379 & 0.3569/0.2172/0.27 \\
\hline 
\multicolumn{9}{l}{Evaluator characteristic: $^{\diamond}$long anomaly misleading.} \\

\hline
\multirow{2}{*}{\makecell[c]{\textbf{Sparse} \\ \textbf{anomalies}}}
% & \textit{c1} & 1.0/0.5/0.667 & 1.0/0.5/0.667 & 1.0/0.5/0.667 & 1.0/0.5/0.667 & 1.0/0.5/0.667 & 1.0/0.5/0.667 & 1.0/0.5/0.667 \\
% & \textit{c2} & 0.5/0.5/0.5 & 0.5/0.5/0.5 & 0.5/0.5/0.5 & 0.5/0.5/0.5 & 0.5/0.5/0.5 & 0.7/0.701/0.7$^{\diamond}$ & 0.5/0.5/0.5 \\
& \textit{c1} & 1.0/0.5/0.6667 & 1.0/0.5/0.6667 & 1.0/0.5/0.6667 & 1.0/0.5/0.6667 & 1.0/0.5/0.6667 & 1.0/0.5/0.6667 & 1.0/0.5/0.6667 \\
& \textit{c1} & 0.5/0.5/0.5 & 0.5/0.5/0.5 & 0.5/0.5/0.5 & 0.5/0.5/0.5 & 0.5/0.5/0.5 & 0.6997/0.7007/0.7002$^{\diamond}$ & 0.5/0.5/0.5 \\
\hline 
\multicolumn{9}{l}{Evaluator characteristic: $^{\diamond}$sparse anomaly misleading.} \\

\hline
\multirow{2}{*}{\makecell[c]{\textbf{Constant} \\ \textbf{detector}}}
% & \textit{c1} & 0.0/0.0/0.0   & 0.0/0.0/0.0   & 0.0/0.0/0.0   & 0.0/0.0/0.0     & 0.0/0.0/0.0                  & nan/0.0/nan                  & 0.0/0.0/0.0 \\
% & \textit{c2} & 0.1/1.0/0.182 & 0.1/1.0/0.182 & 0.1/1.0/0.182 & 0.025/1.0/0.049 & 0.554/1.0/0.713$^{\diamond}$ & 0.506/1.0/0.672$^{\diamond}$ & 0.137/0.92/0.238 \\
& \textit{c1} & 0.0/0.0/0.0   & 0.0/0.0/0.0   & 0.0/0.0/0.0   & 0.0/0.0/0.0     & 0.0/0.0/0.0                  & nan/0.0/nan                  & 0.0/0.0/0.0 \\
& \textit{c2} & 0.1/1.0/0.1818 & 0.1/1.0/0.1818 & 0.1/1.0/0.1818 & 0.025/1.0/0.0488 & 0.555/1.0/0.7138$^{\diamond}$ & 0.5065/1.0/0.6724$^{\diamond}$ & 0.1366/0.9196/0.2378 \\
\hline 
\multicolumn{9}{l}{Evaluator characteristic: $^{\diamond}$overestimation for the all\_1 detector.} \\

\hline
\multirow{2}{*}{\makecell[c]{\textbf{Random} \\ \textbf{detector}}}
% & \textit{c1} & 0.046/0.019/0.027 & 0.115/0.057/0.076             & 0.047/0.019/0.027 & 0.046/0.039/0.042 & 0.087/0.077/0.08 & 0.5/0.45/0.47$^{\diamond}$     & 0.179/0.172/0.173 \\
% & \textit{c2} & 0.05/0.021/0.029  & 0.459/0.64/0.534$^{\diamond}$ & 0.05/0.021/0.029  & 0.05/0.327/0.085  & 0.054/0.341/0.09 & 0.497/0.956/0.652$^{\!\diamond}$ & 0.05/0.194/0.08 \\
& \textit{c1} & 0.0363/0.016/0.022  & 0.0915/0.0467/0.0613  & 0.0372/0.0164/0.0226  & 0.0358/0.0313/0.0327  & 0.0909/0.0838/0.086  & 0.488/0.4536/0.4668$^{\diamond}$  & 0.1707/0.168/0.1679 \\
& \textit{c2} & 0.0421/0.018/0.0249  & 0.3808/0.53/0.4428$^{\diamond}$  & 0.0421/0.018/0.0249  & 0.041/0.2713/0.0699  & 0.0445/0.3043/0.0759  & 0.5129/0.9513/0.6649$^{\diamond}$  & 0.043/0.1714/0.0687 \\
\hline
\multicolumn{9}{l}{Evaluator characteristic: $^{\diamond}$overestimation for the random detector. The evaluation of the random detector is based on the average of 100 experiments.} \\
\hline
\end{tabular}
\label{table_A1}
\end{table*}

\section{Comparison of Evaluation Logics Among Different Evaluators}
\setcounter{table}{0}   %从0开始编号，显示出来表会B1开始编号
\setcounter{figure}{0}
\renewcommand{\thetable}{B\arabic{table}}
\renewcommand{\thefigure}{B\arabic{figure}}

In this appendix, we provide the evaluation logics of different evaluators (point-based, event-based, and \textit{OIPR}), and analyze their computational complexities, respectively.

\begin{figure}[t]
    \begin{minipage}{\linewidth}
        \centering
        \includegraphics[width=3.5in]{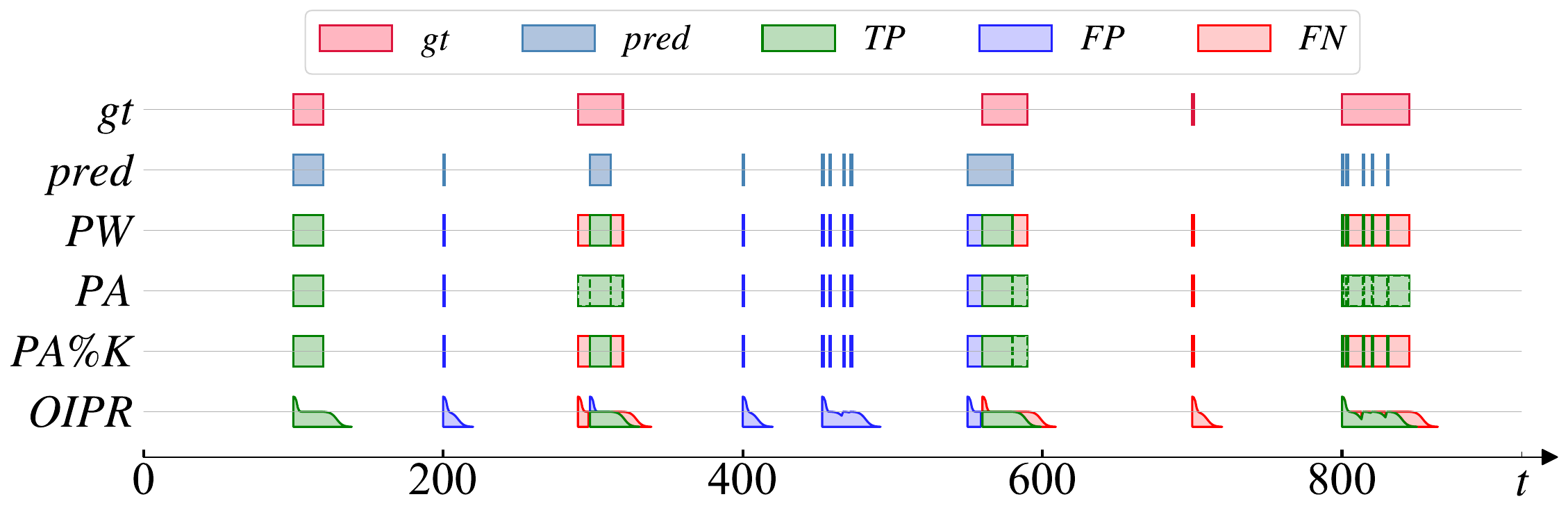}
        \caption{The evaluation logics of the point-based evaluators (\textit{PW}, \textit{PA} and \textit{PA\%K}) compared with \textit{OIPR}.}
        \label{fig_B1}
        
        \vspace{10pt}
        
        \includegraphics[width=3.5in]{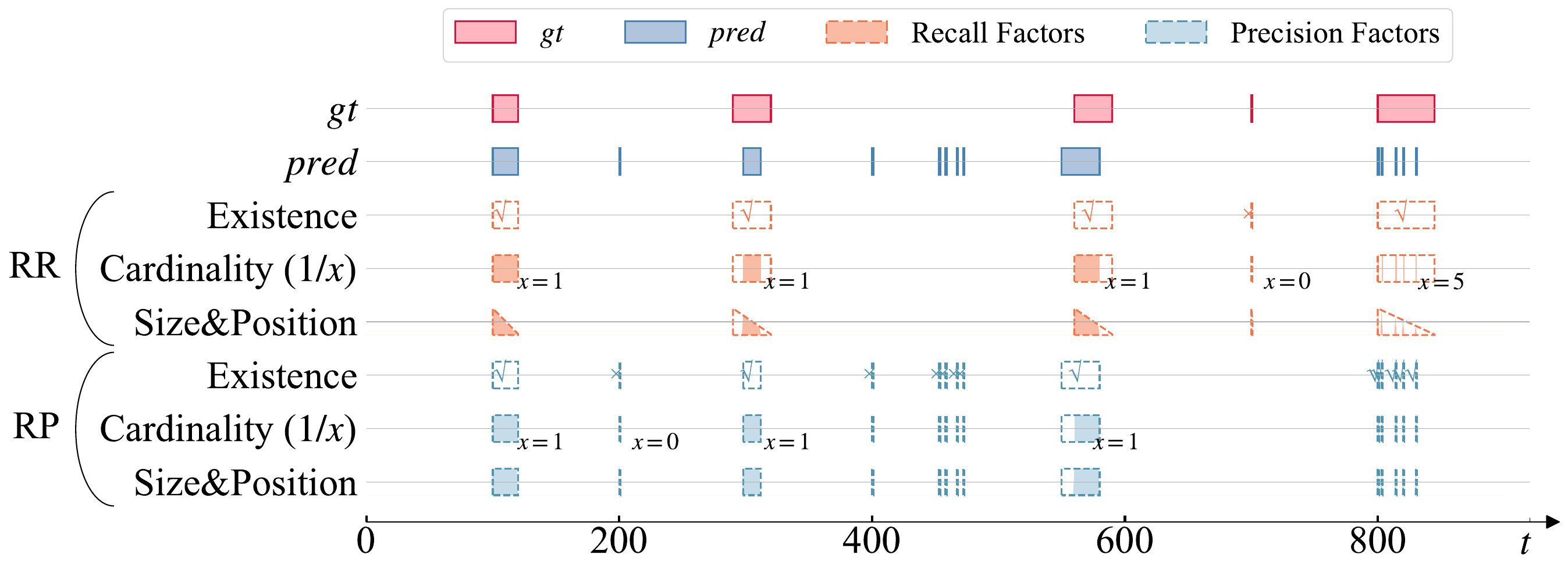}
        \caption{The evaluation logic of \textit{RP/RR}.}
        \label{fig_B2}
    \end{minipage}
\end{figure}

\begin{figure}[t]
    \begin{minipage}{\linewidth}
        \centering
        \includegraphics[width=3.5in]{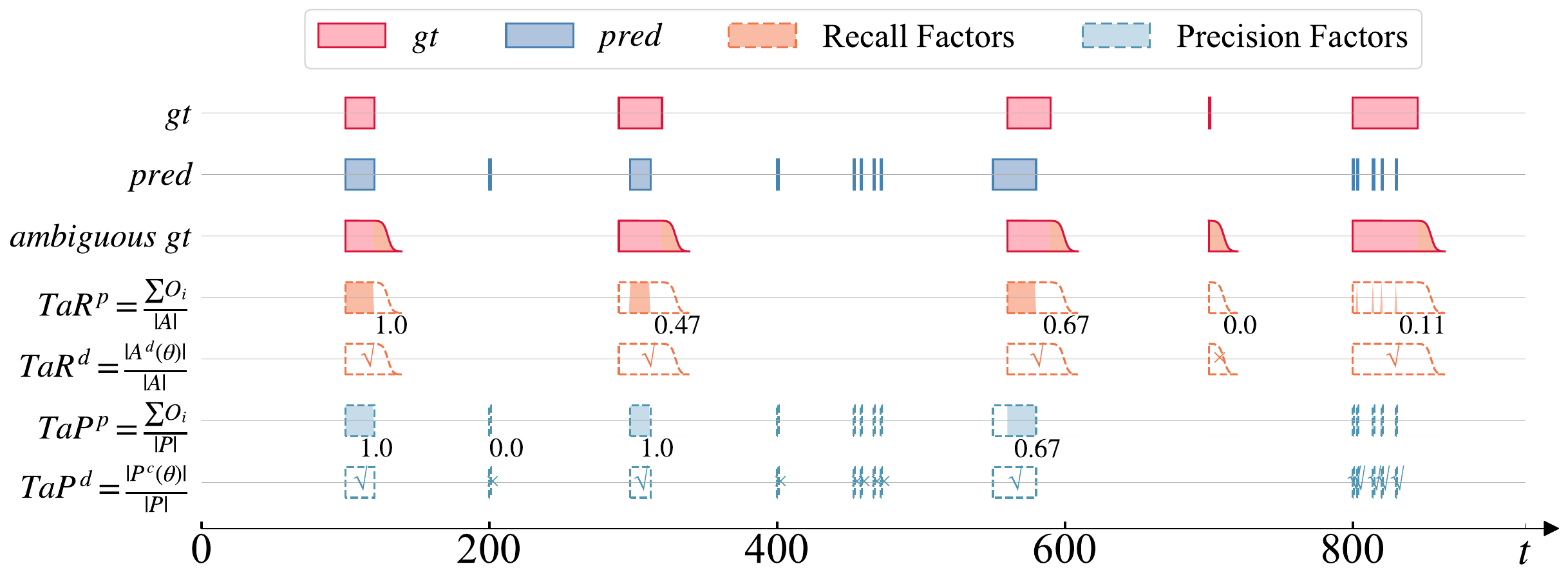}
        \caption{The evaluation logic of \textit{TaPR}.}
        \label{fig_B3}
        
        \vspace{10pt}
        
        \includegraphics[width=3.5in]{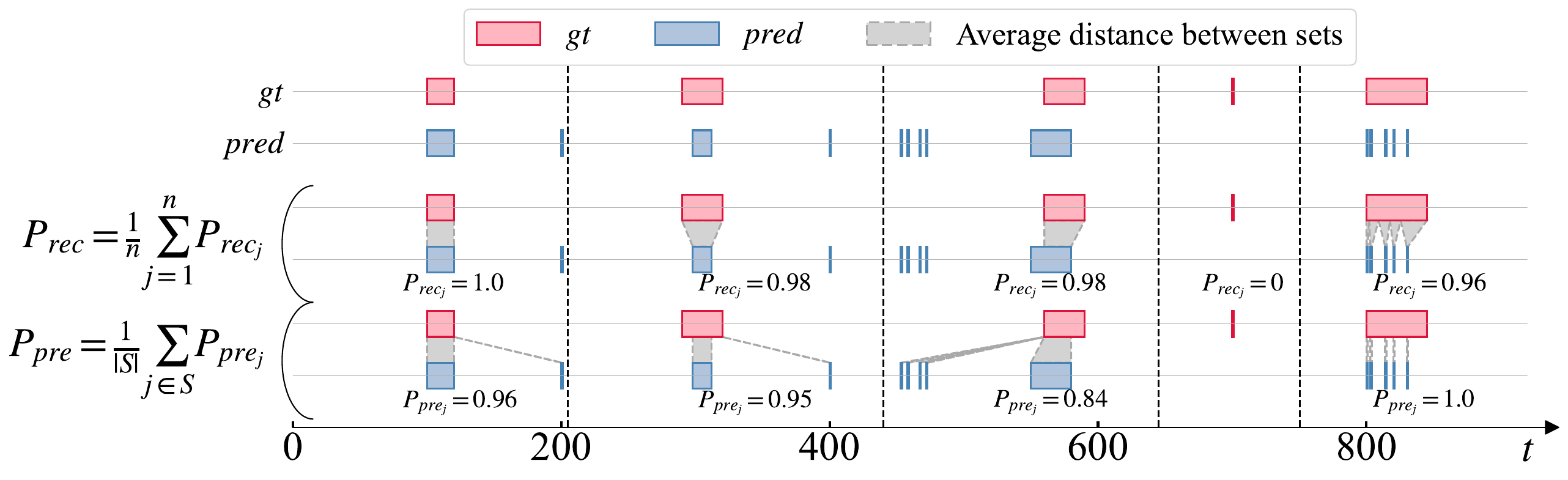}
        \caption{The evaluation logic of \textit{AM}.}
        \label{fig_B4}
        
        \vspace{10pt}
        
        \includegraphics[width=3.5in]{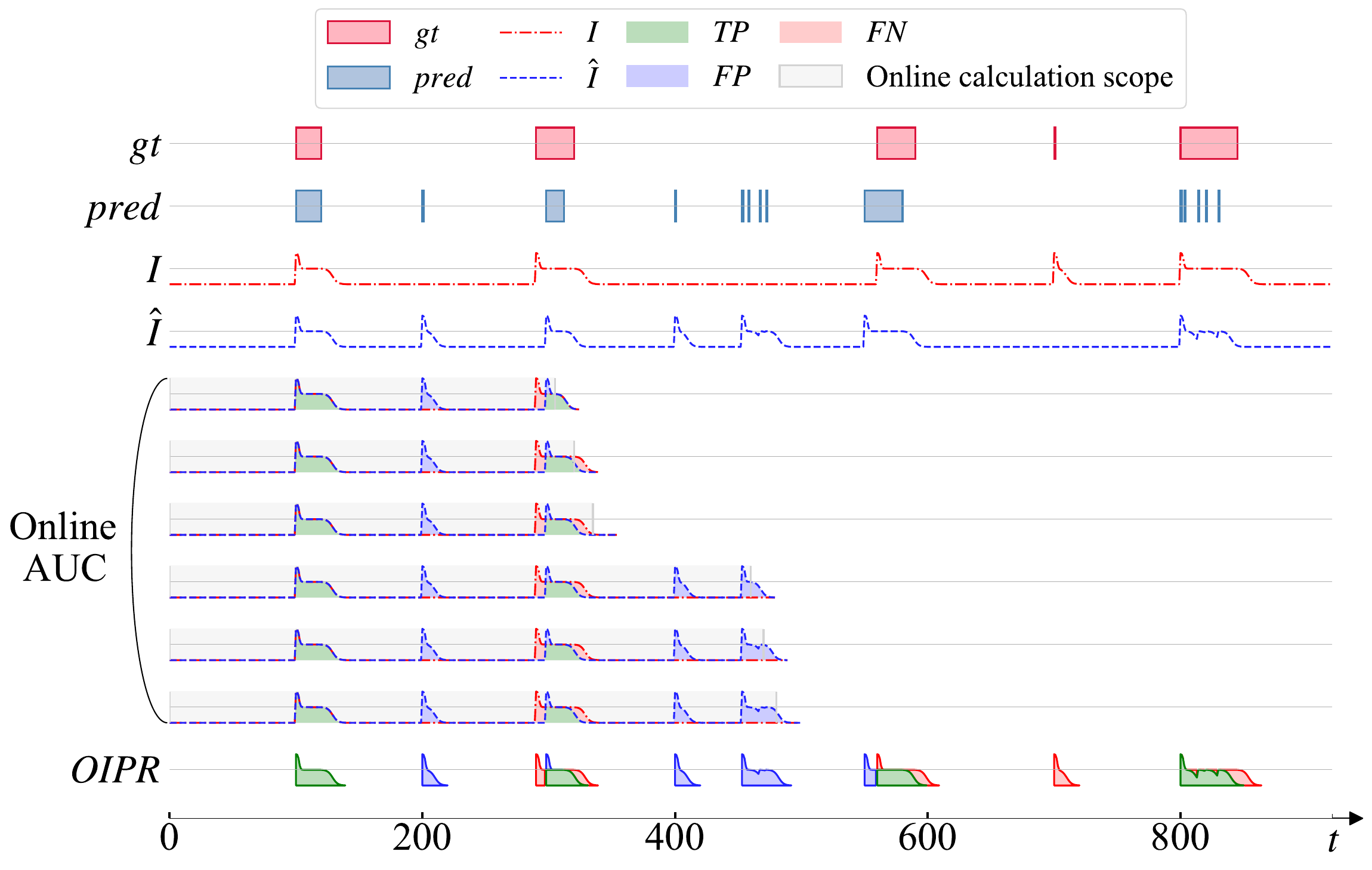}
        \caption{The evaluation logic of \textit{OIPR}, and it online  AUC calculation process.}
        \label{fig_B5}
    \end{minipage}
\end{figure}

Fig. \ref{fig_B1} illustrates the comparison of evaluation logics between point-based evaluators and \textit{OIPR}. First, we present the evaluation processes of three evaluators (\textit{PW}, \textit{PA}, and \textit{PA\%}K) in the form of coverage area. Since each point has the same ``weight'' in point-based evaluators, the quantification of corresponding TP, FP and FN in these evaluators appear as rectangles with constant height on the time axis. The differences between the other evaluators and the classical \textit{PW} evaluator are as follows:

\textbullet\ \textit{PA} adjusts the successful detection of any point in an \textit{gt} (ground truth) event to the successful detection of the event. Due to its point adjustment process, some points evaluated as FN (red areas) in \textit{PW} are evaluated as TP points (green areas) in \textit{PA}, as shown in row 4 of Fig. \ref{fig_B1}.

\textbullet\ \textit{PA\%K} also uses the point adjustment process; the difference is that it only adjusts events where more than $K\%$ of the points are detected. In row 5 of Fig. \ref{fig_B1}, fewer points evaluated as FN in \textit{PW} are evaluated as TP points in \textit{PA\%K}.

\textbullet\ \textit{OIPR} defines TP, FP, FN areas (row 6, Fig. \ref{fig_B1}) with three features: (i) a higher height at the initial (discovery phase) of \textit{gt} or \textit{pred} (prediction) events; (ii) a decaying observation phase trailing after the end of events; (iii) for fragmented prediction results, the trailing covers them to ``connect into a single piece'' with each other, forming a relatively continuous event instead of multiple split ones.

In terms of computational complexity, \textit{PW} evaluates each sample point-by-point with a complexity of $O(N)$. \textit{PA} and \textit{PA\%K} require forward reference up to the length of the current \textit{gt} event for point-by-point evaluation, with a complexity of $O(\overline{L}_a \cdot N)$. \textit{OIPR}'s online calculation (Algorithm 1 in the main text) needs forward extension by $l_{obs}$ for each point, resulting in a complexity of $O(l_{obs} \cdot N)$ . For general datasets, $\overline{L}_a \ll N$, and $l_{obs}$ is a constant. Thus, \textit{PA}, \textit{PA\%K}, and \textit{OIPR} are also regarded having $O(N)$ complexities.

Figs. \ref{fig_B2}–\ref{fig_B5} demonstrate the comparison of evaluation logics between event-based evaluators and \textit{OIPR}. Given the relatively complex internal logic of these evaluators, we provide only a brief overview of their key focuses:

\textbullet\ \textit{RP/RR} mainly considers three key factors: existence, cardinality, and size\&position, as shown in Fig. \ref{fig_B2}. For the calculation of recall (\textit{RR}) (marked in orange), it considers: whether each \textit{gt} event is detected (existence); how many \textit{pred} events are used to detect one \textit{gt} event (cardinality); the position of \textit{pred} events to detect the \textit{gt} event and the overlap ratio between them (size\&position). Finally, the \textit{RR} values of each \textit{gt} event are averaged at the event level. The calculation of precision (\textit{RP}) (marked in blue) is symmetric to that of \textit{RR}, except for the difference in position function.

\textbullet\ \textit{TaPR}'s evaluation logic is illustrated in Fig. \ref{fig_B3}. First, accounting for ambiguous labels, \textit{TaPR} calculates a trailing interval for each \textit{gt} event, as shown in row 3, with subsequent computations performed on \textit{pred} and \textit{ambiguous gt} events. For recall (\textit{TaR}), the factors considered include $TaR^p$ and $TaR^d$: $TaR^p$ accounts for the ratio of the overlapping area (where an \textit{ambiguous gt} event is detected by \textit{pred}) to the total area of the event, while $TaR^d$ considers whether the anomaly event is detected at a threshold level of $\theta$. Finally, the \textit{TaR} values of each \textit{ambiguous gt} event are averaged at the event level. The calculation of precision (\textit{TaP}) is symmetric to that of \textit{TaR}.

\textbullet\ \textit{AM}'s evaluation logic is demonstrated in Fig. \ref{fig_B4}. First, the time-series is divided into affiliation zones according to \textit{gt} events. Then, within each affiliation zone, recall is calculated using the average inter-set distance from \textit{gt} to \textit{pred} events, while precision uses that from \textit{pred} to \textit{gt}. Finally, both precision and recall are averaged at the \textit{gt} event level.

\textbullet\  The commonality of \textit{OIPR} and these event-based evaluators is that they all decompose into several factors. For \textit{OIPR} (Fig. \ref{fig_B5}), this is manifested by first calculating the operator interest curve $I$ for \textit{gt} and $\hat{I}$ for \textit{pred}, then computing their overlapping and non-overlapping areas, and finally obtaining \textit{OIPR}'s TP, FP and FN areas. The difference is that when calculating each factor, \textit{OIPR} uses online AUC instead of the current event as the calculation unit. The ``merging'' of multiple fragmented events is naturally achieved through the overlap of observation phases.

In terms of computational complexity, \textit{RP/RR} and \textit{TaPR} require forward reference to at most one event's length (\textit{gt}, \textit{ambiguous gt}, \textit{pred} events). Thus, \textit{RP/RR} has a complexity of $O(kN)$ ($k$: average length of \textit{gt} and \textit{pred} events); \textit{TaPR} has a complexity of $O(\hat{k}N)$ ($\hat{k}$: average length of \textit{ambiguous gt} and \textit{pred} events). Since $k, \hat{k} \ll N$, both are effectively $O(N)$. For \textit{AM}, it differs: forward reference until the next \textit{gt} event is needed to determine the affiliation zone, giving a complexity of $O(sN)$ ($s$: average interval between the start times of adjacent \textit{gt} events), with $s$ non-negligible relative to $N$. Thus, the final complexity of \textit{AM} is $O(sN)$.

\section{Discussion on Attenuation Function and Long/Short Anomaly Events}
\setcounter{table}{0}
\renewcommand{\thetable}{C\arabic{table}}

\setcounter{figure}{0}
\renewcommand{\thefigure}{C\arabic{figure}}

\setcounter{equation}{0}
\renewcommand{\theequation}{C.\arabic{equation}}

% 关键：分离“子图标题编号”和“正文引用编号”
\makeatletter
% 1. 让子图标题仅显示“(字母)”（如(a)）
\def\thesubfigure{(\alph{subfigure})}
% 2. 让正文引用时，在“(字母)”前自动拼接“C+主图数字”（形成C1(a)）
% \p@subfigure 控制引用时的前缀，此处手动拼接“C\arabic{figure}”
\def\p@subfigure{C\arabic{figure}}
\makeatother

\begin{figure}[t]
	\centering
    \subfigure[Operator interest curves under different attenuation functions for a complete anomaly event.] {\includegraphics[width=1.8in]{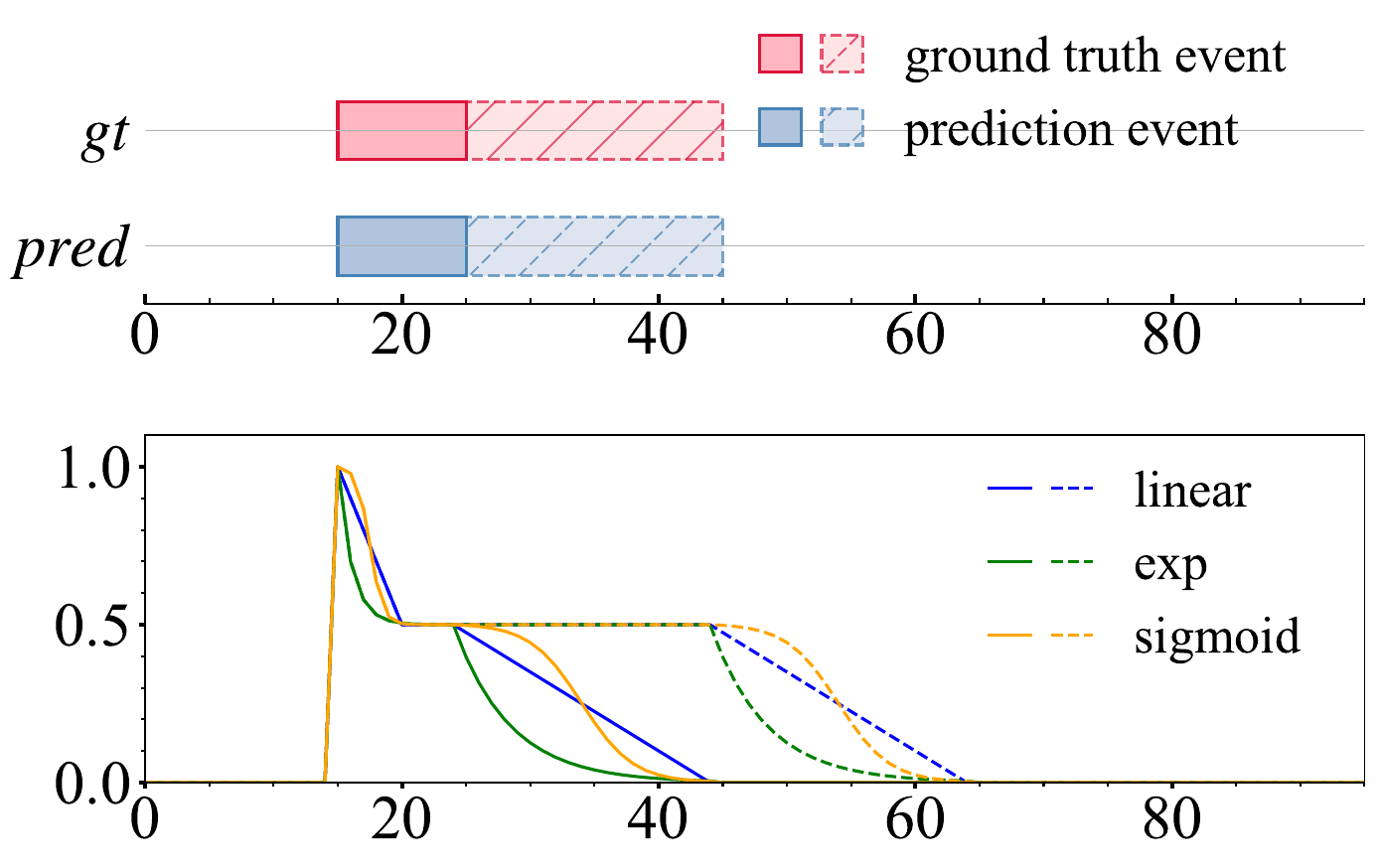}\label{fig_C1a}}
	\subfigure[Curves of TP variation with $x$ under different attenuation functions.] {\includegraphics[width=1.55in]{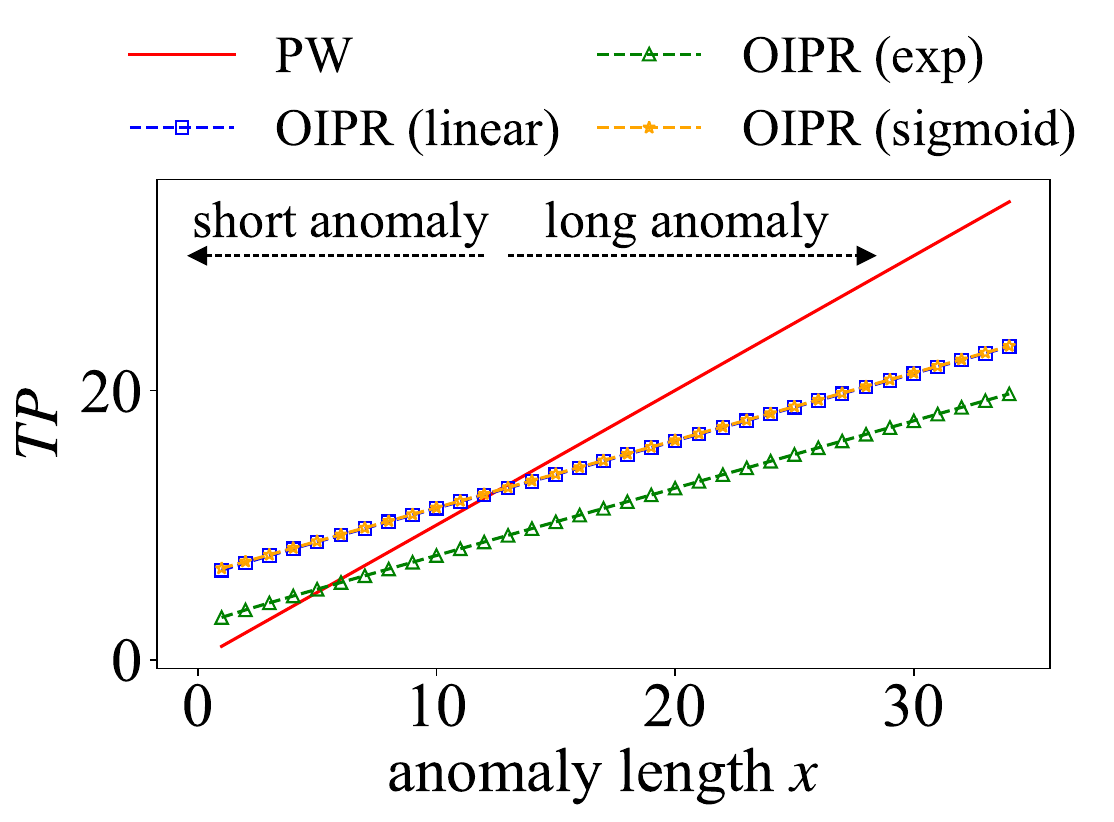}\label{fig_C1b}}
	\caption{Comparison of \textit{OIPR} implementations under linear, exponential, and sigmoid attenuation functions.}
	\label{fig_C1}
\end{figure}

This appendix discusses the definitions and experimental comparisons of different attenuation functions, as well as how \textit{OIPR} distinguishes long and short anomaly events.

\subsection{Long/Short Anomaly Events}\label{sec C.1}
The standard implementation of \textit{OIPR} employs sigmoid attenuation functions, as presented in (3) and (4) in the main text, for the discovery and observation phases. Herein, we further propose \textit{OIPR} implementations under linear and exponential attenuation schemes, for comparative analysis.

The detailed functions of $\omega$ and $\gamma$ under linear attenuation are defined as follows:
\begin{equation}
    \omega_{linear}(i) \! =  \! \left \{
    \begin{aligned}
       & 1, i = 0\\
       & 1 - (1 - b_{dur}) \times \frac{i}{l_{dis}}, 0 < i \leq l_{dis} \\
       & b_{dur}, i > l_{dis}
    \end{aligned}
    \right.
    \!,
\label{eqC.1}
\end{equation}
\begin{equation}
    \gamma_{linear}(i)=\left \{
    \begin{aligned}
       &1, i = 0\\
       &1 - \frac{i}{l_{obs}}, 0 < i \leq l_{obs}\\
       &0, i > l_{obs}
    \end{aligned}
    \right.
    .
\label{eqC.2}
\end{equation}

Similarly, the functions $\omega$ and $\gamma$ under exponential attenuation are defined as follows:
\begin{equation}
    \omega_{exp}(i) \! =  \! \left \{
    \begin{aligned}
       & 1, i = 0\\
       & b_{dur} + (1 - b_{dur}) \times e^{(-\text{log}\;100 \times \frac{i}{l_{dis}})}, i  \! >  \! 0
    \end{aligned}
    \right.
    \!,
\label{eqC.3}
\end{equation}
\begin{equation}
    \gamma_{exp}(i)=\left \{
    \begin{aligned}
       &1, i = 0\\
       &e^{(-\text{log}\;100 \times \frac{i}{l_{obs}})}, 0 < i \leq l_{obs}\\
       &0, i > l_{obs}
    \end{aligned}
    \right.
    .
\label{eqC.4}
\end{equation}

Fig. \ref{fig_C1a} presents a comparison of the operator interest curves for a complete anomaly event under linear, exponential, and sigmoid attenuation functions. We vary the anomaly length $x$ (as indicated by the dashed box) and calculate TP (equivalent to $\text{AUC}(I)$ here) as a function of $x$ under different attenuation functions (Fig. \ref{fig_C1b}). The TP curve of classical \textit{PW} is included as a reference. It exhibits a standard proportional relationship with $x$, starting from the origin with a slope of 1. The range where the TP value of \textit{OIPR} exceeds that of \textit{PW} is defined as ``short anomaly events'', meaning that \textit{OIPR} gives higher evaluation weight to short anomalies than \textit{PW} to reward their successful detection as independent events. This shows that \textit{OIPR} prioritizes the detection of anomaly events over anomaly points compared to \textit{PW}. Correspondingly, the range where the TP value of \textit{OIPR} is lower than that of \textit{PW} is defined as ``long anomaly events.''

Subsequently, we calculate the threshold distinguishing long and short anomaly events. For clarity, Table \ref{table_C1} shows how the threshold varies with $l_{dis}$ under a fixed configuration of $l_{obs}=4l_{dis}$. For linear and sigmoid attenuation functions, the threshold between long and short anomalies can be approximated as $2.5l_{dis}$. For the exponential attenuation function, the relationship between the threshold and $l_{dis}$ cannot be estimated using a simple linear relationship. This is one advantage of linear and sigmoid attenuation functions over exponential attenuation function. A more detailed proof follows:

For the linear attenuation function, TP can be expressed in a closed form by calculating the cumulative sum of the operator interest curve for a complete anomaly event $\Phi_{linear}$, and then substituted into $l_{obs}=4l_{dis}$, $b_{dur}=0.5$ and $x = l_{dis} + l_{dur}$ to obtain a simplified form:

\begin{equation}
\begin{aligned}
    & \sum_{i=0}^{l_{dis}+l_{dur}+l_{obs}-1}\Phi_{linear}(i)\\
    &= \frac{1+l_{dis} \cdot (1 + b_{dur}) }{2} +  b_{dur} \cdot (l_{dur} + \frac{l_{obs}}{2} - 1)\\
    &= \frac{5l_{dis} + 2x}{4}.
\end{aligned}
\label{eqC.5}
\end{equation}

Calculate the intersection point between \ref{eqC.5} and the reference line $TP=x$ (representing \textit{PW}), which gives $2.5l_{dis}$.

Similarly, TP under the exponential attenuation function can be expressed as follows, where $C=e^{(-\text{log}\;100)}$:
% \begin{equation}
\begin{align}
    & \sum_{i=0}^{l_{dis}+l_{dur}+l_{obs}-1}\Phi_{exp}(i) \\
    =& b_{dur} \cdot (l_{dis} \! + \! l_{dur} \! - \! 1) \! + \! (1 \! - \!  b_{dur}) \cdot C^\frac{1}{l_{dis}} \cdot \frac{1 - C^\frac{l_{dis} + l_{dur} - 1}{l_{dis}}}{1-C^{l_{dis}}} \notag \\
    & + [b_{dur} + (1 - b_{dur}) \cdot C^\frac{l_{dis}+l_{dur}}{l_{dis}}] \cdot C^\frac{1}{l_{obs}} \cdot \frac{1 - C}{1 - C^\frac{1}{l_{obs}}} + 1 \notag \\
    =& \frac{x+1}{2} + \frac{1 - C^\frac{x-1}{l_{dis}}}{2(1-C^\frac{1}{l_{dis}})} + \frac{(1 + C^\frac{x}{l_{dis}}) \cdot C^\frac{1}{4l_{dis}} \cdot (1 - C)}{2(1-C^\frac{1}{4l_{dis}})}. \notag
\end{align}
\label{eqC.6}
% \end{equation}

In this case, the intersection point does not yield a simple closed-form linear solution.

For the sigmoid attenuation function, there is no analytical closed-form expression for its cumulative sum. However, due to its shape symmetry (see Fig. \ref{fig_C1a}), the area under its curve can be reasonably approximated using the result derived from the linear attenuation function. Thus, in integer-level estimation, the $2.5l_{dis}$ intersection value can be used as the threshold between long and short anomalies, consistent with the results in Table \ref{table_C1}.

In summary, for the sigmoid implementation of \textit{OIPR} proposed in the main text, we can approximately define $2.5l_{dis}$ as the threshold between long and short anomaly events.

\begin{table}[!t]
	\renewcommand{\arraystretch}{1.3}
	\caption{The threshold between long/short anomalies varies with $l_{dis}$ under a fixed configuration of $l_{obs}=4l_{dis}$.}
	\label{table_C1}
	\centering
    \setlength{\tabcolsep}{5pt}
        % \begin{tabular}{c | p{13pt}<{\centering} p{15pt}<{\centering} p{15pt}<{\centering} p{15pt}<{\centering} p{15pt}<{\centering} p{15pt}<{\centering}}
        \begin{tabular}{c|cccccc}
        \hline
        \diagbox[width=73pt]{\textbf{Attenua. func.}}{$l_{dis}$, $l_{obs}$} & \textbf{1, 4} & \textbf{3, 12} & \textbf{5, 20} & \textbf{10, 40} & \textbf{15, 60} & \textbf{20, 80}\\     
	\hline
        \textbf{linear} & 2.5 & 7.5 & 12.5 & 25 & 37.5 & 50\\
        \textbf{exp} & 1.46 & 3.39 & 5.48 & 10.82 & 16.19 & 21.56\\
        \textbf{sigmoid} & 2.52 & 7.56 & 12.59 & 25.18 & 37.77 & 50.35\\
		\hline
	\end{tabular}
\end{table}

\subsection{Differences Among Attenuation Functions}
In Appendix \ref{sec C.1}, we present the first advantage of the sigmoid attenuation function over the exponential one: the ability to linearly estimate the threshold between long and short anomaly events.

Next, we provide the evaluation results using \textit{OIPR} implementations with three attenuation functions on several selected scenarios from the special scenario dataset, as shown in Table \ref{table_C2}. In the TP positions scenario, sigmoid and exponential attenuation functions show another advantage over the linear function: during the duration phase, their operator interest values are not entirely constant (though changing slightly), enabling differentiation of prediction points at different positions within the duration phase for the same \textit{gt} event. In contrast, linear attenuation function yields completely constant operator interest during the duration phase (row 3 of \ref{eqC.1}), causing it to lose the ability to distinguish the prediction positions within the duration phase. Notably, all \textit{OIPR} implementations with each of the three attenuation functions provide sufficient discriminative capability for prediction positions during the discovery phase and the observation phase.

Furthermore, the sigmoid attenuation function exhibits two additional characteristics:

First, at the transition position between the duration phase and the observation phase, the sigmoid attenuation function demonstrates smoother behavior compared to linear and exponential ones, as illustrated in Fig. \ref{fig_C1a}.

Second, the existence reward and the ambiguous label reward provided by the sigmoid and linear attenuation functions are higher than that of the exponential one (see Table \ref{table_C2}. Existence reward: Overlap proportion, c1; Ambiguous label reward: Temporal shifting, c1 and c2).

\begin{table}[!t]
\renewcommand{\arraystretch}{1.3}
\caption{Comparison of evaluation results (P/R/F1) of \textit{OIPR} with different attenuation functions in several special scenarios.}
\centering
\setlength{\tabcolsep}{1pt}
% \fontsize{7.25pt}{11pt}\selectfont
\scriptsize
\begin{tabular}{p{28pt}<{\raggedright} p{13pt}<{\raggedleft} lll}
\hline
\bfseries Scenario & \bfseries Case  & \makecell[c]{\textbf{OIPR (linear)}} & \bfseries \makecell[c]{\textbf{OIPR (exp)}} & \bfseries \makecell[c]{\textbf{OIPR (sigmoid)}} \\
\hline 
\multirow{4}{*}{\makecell[c]{Overlap \\ proportion}}
& \textit{c1} & 1.0/0.2128/0.3509 & 1.0/0.1133/0.2035 & 1.0/0.2168/0.3564 \\
& \textit{c2} & 1.0/0.36/0.5294 & 1.0/0.2791/0.4364 & 1.0/0.3609/0.5304 \\
& \textit{c3} & 1.0/0.616/0.7624 & 1.0/0.5675/0.724 & 1.0/0.6166/0.7628 \\ 
& \textit{c4} & 1.0/1.0/1.0 & 1.0/1.0/1.0 & 1.0/1.0/1.0 \\
\hline
\multirow{2}{*}{\makecell[c]{Temporal \\ shifting}}
& \textit{c1} & 0.7361/0.7361/0.7361 & 0.5412/0.5412/0.5412 & 0.7285/0.7285/0.7285 \\
& \textit{c2} & 0.7361/0.7361/0.7361 & 0.5412/0.5412/0.5412 & 0.7285/0.7285/0.7285 \\
\hline
\multirow{4}{*}{\makecell[c]{TP \\ positions}}
& \textit{c1} & 1.0/0.3129/0.4767 & 1.0/0.1771/0.301 & 1.0/0.3186/0.4833 \\
& \textit{c2} & 0.8241/0.2579/0.3928 & 0.8256/0.1462/0.2485 & 0.7859/0.2504/0.3798 \\
& \textit{c3} & \fontsize{7.3pt}{9pt} 0.8241/0.2579/0.3928 & 0.8233/0.1458/0.2478 & 0.7853/0.2502/0.3795 \\
& \textit{c4} & 0.7895/0.2471/0.3763 & 0.7673/0.1359/0.231 & 0.7789/0.2482/0.3764 \\
\hline
\multirow{3}{*}{ \makecell[c]{Fragmented \\ TPs} }
& \textit{c1} & 0.7616/1.0/0.8647 & 0.8495/1.0/0.9186 & 0.7584/1.0/0.8626 \\
& \textit{c2} & 0.7551/0.9647/0.8471 & 0.8303/0.867/0.8483 & 0.7571/0.993/0.8591 \\
& \textit{c3} & 0.7584/0.9821/0.8559 & 0.8384/0.9188/0.8767 & 0.758/0.9982/0.8617 \\
\hline
\multirow{3}{*}{ \makecell[c]{Fragmented \\ FPs} }
& \textit{c1} & 0.1964/1.0/0.3283 & 0.2885/1.0/0.4478 & 0.1937/1.0/0.3245 \\
& \textit{c2} & 0.5119/1.0/0.6772 & 0.5307/1.0/0.6934 & 0.5081/1.0/0.6739 \\
& \textit{c3} & 0.5/1.0/0.6667 & 0.5/1.0/0.6667 & 0.5/1.0/0.6667 \\
\hline
% \multirow{3}{*}{ \makecell[c]{Long \\ anomaly \\ effect} }
% & \textit{c1} & 1.0/0.22/0.361 & 1.0/0.291/0.451 & 1.0/0.217/0.357 \\
% & \textit{c2} & 1.0/0.78/0.876 & 1.0/0.709/0.83 & 1.0/0.783/0.878 \\
% & \textit{c3} & 0.361/0.22/0.273 & 0.451/0.291/0.354 & 0.357/0.217/0.27 \\
% \hline 
\end{tabular}
\label{table_C2}
\end{table}

% \clearpage
% \newpage

\section{Parameter Selection and Sensitivity Analysis}
\setcounter{table}{0}   %从0开始编号，显示出来表会B1开始编号
\setcounter{figure}{0}
\renewcommand{\thetable}{D\arabic{table}}
\renewcommand{\thefigure}{D\arabic{figure}}

\begin{figure*}[t]
	\centering
	\includegraphics[width=7in]{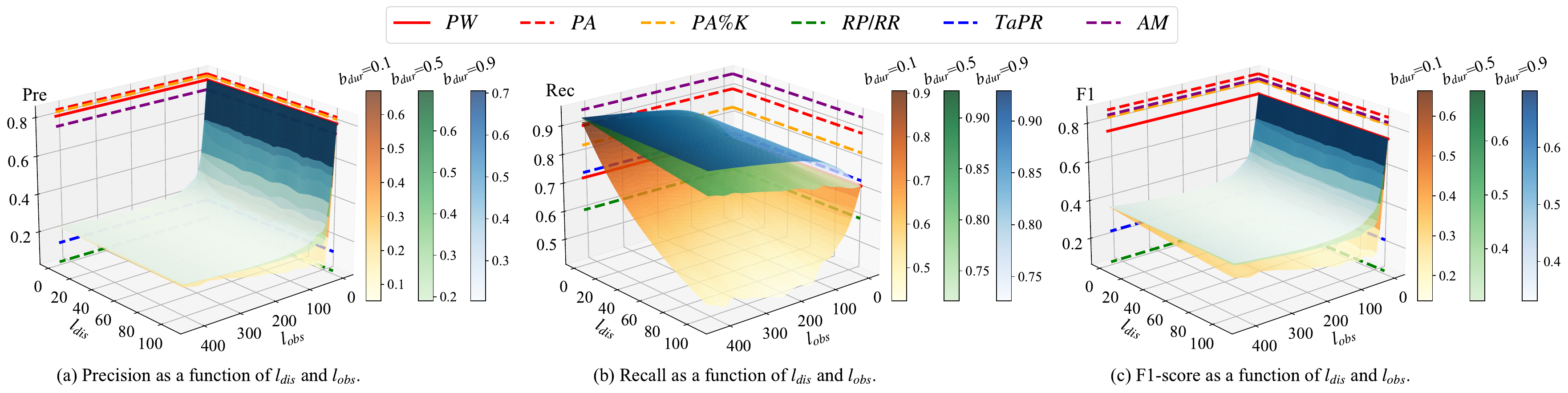}
	\caption{\textit{OIPR} evaluation metrics (P/R/F1) under different parameters ($l_{dis}$, $l_{obs}$, $b_{dur}$) with the detection results of Autoformer on MSL dataset.}
	\label{fig_D1}
\end{figure*}

\begin{figure*}[t]
	\centering
	\includegraphics[width=7in]{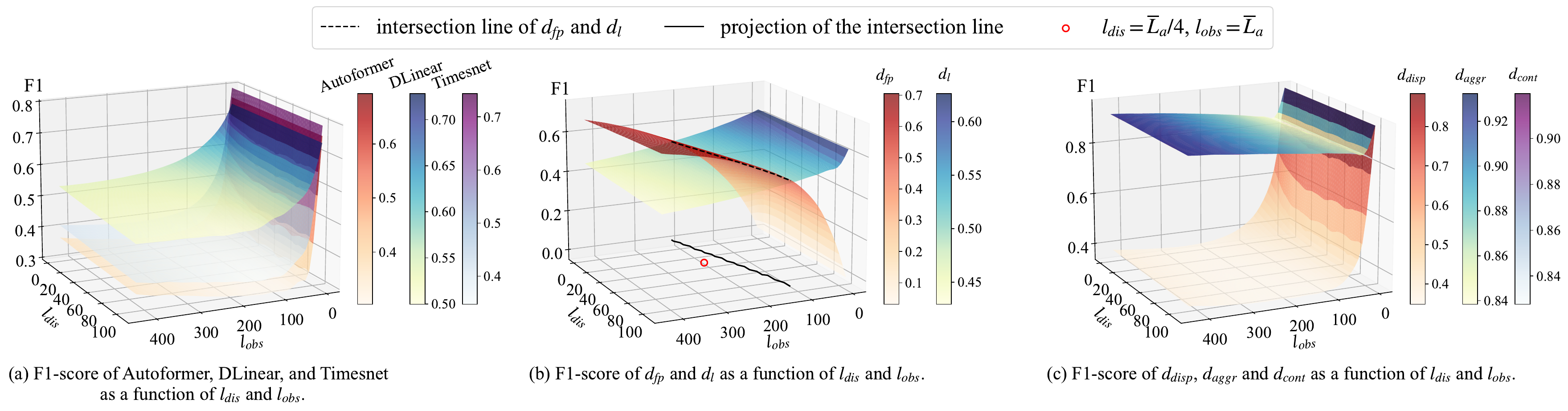}
	\caption{F1-scores varying with $l_{dis}$ and $l_{obs}$ across three baseline detectors and five adversary detectors on MSL dataset with $b_{dur}=0.5$.}
	\label{fig_D2}
\end{figure*}

In this appendix, we perform parameter sensitivity analysis experiments, and plot 3D surfaces under different parameter selections to demonstrate the impact of parameter combinations on \textit{OIPR}’s evaluation results.

Fig. \ref{fig_D1} presents the 3D surface plots of \textit{OIPR} evaluation metrics (P/R/F1) under different parameter selections ($l_{dis}$, $l_{obs}$, $b_{dur}$), using the prediction results of Autoformer detector on the MSL dataset. Additionally, the results of other evaluators (2D lines) are marked in the plots for comparison. It can be observed that \textit{OIPR} is equivalent to the classical \textit{PW} evaluator when $l_{obs} = 0$. An increase in $l_{obs}$ leads to a decrease in precision and an increase in recall when $b_{dur}$ = 0.5 and 0.9, ultimately resulting in a decrease in f1-score. A longer $l_{obs}$ causes each event to have a larger trailing area of the observation phase. However, due to the continuity of \textit{gt} labels and the discreteness of \textit{pred} results, there are significantly more \textit{pred} events than \textit{gt} events. This reduces the increase in TP smaller than that in FP, thus reducing the precision. Meanwhile, the increase in the observation phase area leads to part of the FN areas being ``merged into'' adjacent \textit{pred} events and judged as TP areas. In general, FN areas decrease, resulting in an increase in recall.

The situation differs when $b_{dur} = 0.1$: an excessively small $b_{dur}$ leads to a significant reduction in the areas occupied by the duration and observation phases, making the duration of event less important. At this point, \textit{OIPR} only focuses on whether the start of \textit{gt} events aligns with that of \textit{pred} events. A larger $l_{obs}$ merges numerous discrete \textit{pred} points into fewer \textit{pred} events, reducing the number of start points and thus causing a substantial decrease in TP area, which in turn lowers the recall. Obviously, an excessively small $b_{dur}$ is detrimental to the fair evaluation of \textit{OIPR}.

Compared to $l_{obs}$, an increase in $l_{dis}$ has a smaller impact on P/R/F1. This is because $l_{dis}$ mainly affects the overlapping parts of \textit{gt} and \textit{pred} events; for more events without overlap, an increase in $l_{dis}$ leads to proportional increases in TP, FP, and FN. Overall, since the start points of \textit{gt} and \textit{pred} events rarely align perfectly, increases in FP and FN areas caused by a larger $l_{dis}$ both exceed the increase in TP area, resulting in slight decreases in P/R/F1.

Fig. \ref{fig_D2} presents the f1-scores varying with $l_{obs}$ and $l_{dis}$, across three baseline detectors and five adversary detectors on the MSL dataset, with $b_{dur}=0.5$. Compared to the absolute values of the evaluation metrics, the ranking among different detectors is of greater significance. As shown in Fig. \ref{fig_D2}(a) and Fig. \ref{fig_D2}(c), the ranking of detectors by \textit{OIPR} remains consistent across a large parameter range; this is particularly true for the popular detectors in Fig. \ref{fig_D2}(a), confirming the stability of \textit{OIPR}. The parameters can thus be selected approximately within a large range without strict requirements.

In Fig. \ref{fig_D2}(b), when $l_{dis}$ and $l_{obs}$ are small, $d_l$ ranks higher than $d_{fp}$, indicating that \textit{OIPR} prefers detectors that detect anomalies with longer durations at this point. Conversely, when $l_{dis}$ and $l_{obs}$ are large, $d_{fp}$ ranks higher than $d_l$, suggesting that \textit{OIPR} prefers detectors that detect more anomaly events. Our selected parameters ($l_{dis} = \overline{L}_a/4$, $l_{obs} = \overline{L}_a$) lie near the boundary of these two scenarios, demonstrating that \textit{OIPR} has the sensitivity to strike a balance between the detection of longer and more anomalies, unlike point-based evaluators which are completely inclined toward longer anomalies, and event-based evaluators which are entirely biased toward more anomalies (as described in Section V-B).

% \clearpage
% \newpage

\end{document}